%% file: main.tex
\documentclass{article}

\PassOptionsToPackage{numbers, sort&compress}{natbib}

\usepackage{enumitem}
\usepackage[eandd,preprint]{neurips_2026}
\usepackage{graphicx}
\usepackage{multirow}
\usepackage{wrapfig}

\usepackage{etoc}
\usepackage[utf8]{inputenc} 
\usepackage[T1]{fontenc}    
\usepackage{hyperref}       
\usepackage{url}            
\usepackage{booktabs}       
\usepackage{amsfonts}       
\usepackage{nicefrac}       
\usepackage{microtype}      
\usepackage[table]{xcolor}  
\usepackage{amsmath}        
\usepackage{colortbl}   
\usepackage{algorithm}
\usepackage{algpseudocode}
\usepackage{caption}    
\usepackage{array}
\usepackage{amssymb}

\definecolor{StimulusOnset}{HTML}{52A754}
\definecolor{movementOnset}{HTML}{E16D3F}
\definecolor{feedbackOnset}{HTML}{A95097}
\definecolor{highlight}{HTML}{EBF5FB}

\hypersetup{
    colorlinks,
    linkcolor={blue!64!black},
    citecolor={blue!64!black},
    urlcolor={blue!64!black}
}

\newcommand{\tpms}[1]{\raisebox{0.0ex}{\scriptsize{\ensuremath{\,\pm\,}#1}}}

\newcommand{\brainwidebench}{\texttt{BrainWideBench}}
\newcommand{\tss}{\texttt{TSS}}
\newcommand{\tsu}{\texttt{TSU}}

\newcommand{\numsubjects}{139}

\newcommand{\numregions}{276}
\newcommand{\numprobes}{688}

\newcommand{\pretrainsubjects}{126}
\newcommand{\pretrainsessions}{423}
\newcommand{\pretrainregions}{274}

\newcommand{\evalsubjects}{13}
\newcommand{\evalsessions}{29}
\newcommand{\evalregions}{124}

\definecolor{pistachio}{HTML}{D7E7D0} 

\colorlet{bwbTSS}{pistachio!60} 
\colorlet{bwbTSU}{magenta!10} 
\colorlet{bwbTSUind}{magenta!10} 

\definecolor{trainColor}{HTML}{84E386}
\definecolor{testColor}{HTML}{ED9F9F}
\definecolor{valColor}{HTML}{FCE49A}
\newcommand{\splitbox}[2]{\colorbox{#1}{\texttt{\vphantom{A}#2}}}
\newcolumntype{+}{>{\global\let\currentrowstyle\relax}}
\newcolumntype{^}{>{\currentrowstyle}}

\definecolor{tablebg_avgrank}{rgb}{0.95,0.95,0.95}
\newcolumntype{R}{>{\centering\arraybackslash\columncolor{tablebg_avgrank}}c}

\title{BrainWideBench: Benchmarking large-scale pretraining and across-animal transfer in multi-region neural recordings}

\author{%
  \noindent\hspace*{-0.0664\linewidth}%
  \begin{minipage}{1.1\linewidth}
  \centering
  {\bf Alexandre Andre}$^{1,}$\thanks{Equal contribution. \quad $^\dagger$Shared senior authorship. \\
  \hspace*{1.8em}\normalfont Contact: \ttfamily\{aandre1, smahato, dyer1\}@upenn.edu, \{m.whiteway, lmp2107\}@columbia.edu}\ ,
  {\bf Shivashriganesh P. Mahato}$^{1,}$\footnotemark[1]\ , \\
  {\bf Vinam Arora}$^{1}$, 
  {\bf Keshav Balaji}$^{1}$, 
  {\bf Divyansha Lachi}$^{1}$, 
  {\bf Nanda H. Krishna}$^{2,3}$, 
  {\bf Jingyun Xiao}$^{1}$, \\
  {\bf Yizi Zhang}$^{4}$,
  {\bf Ximeng Mao}$^{2,3}$, 
  {\bf Wenrui Ma}$^{1}$, 
  {\bf Han Yu}$^{5}$, 
  {\bf International Brain Laboratory}, \\
  {\bf Daniel Birman}$^{6}$, 
  {\bf Niccolò Bonacchi}$^{7,8}$, 
  {\bf Gaelle A. Chapuis}$^{9}$, 
  {\bf Joana A. Catarino}$^{10}$, \\
  {\bf Felicia Davatolhagh}$^{11}$, 
  {\bf Mayo Faulkner}$^{12}$, 
  {\bf Laura Freitas-Silva}$^{13}$,
  {\bf Fei Hu}$^{14}$, \\
  {\bf Julia M. Huntenburg}$^{13}$, 
  {\bf Anup Khanal}$^{11}$, 
  {\bf Inês Laranjeira}$^{13}$, 
  {\bf Petrina Lau}$^{15}$, \\
  {\bf Guido T. Meijer}$^{16}$, 
  {\bf Nathaniel J. Miska}$^{12}$, 
  {\bf Jean-Paul Noel}$^{17}$, \\
  {\bf Alejandro Pan-Vazquez}$^{18}$, 
  {\bf Georg Raiser}$^{13}$, 
  {\bf Cyrille Rossant}$^{12}$, 
  {\bf Karolina Z. Socha}$^{11}$, \\
  {\bf Anne E. Urai}$^{19}$, 
  {\bf Miles J. Wells}$^{12}$, 
  {\bf Steven J. West}$^{12}$, 
  {\bf Olivier Winter}$^{13}$, \\
  {\bf Blake Richards}$^{20,2}$, 
  {\bf Guillaume Lajoie}$^{3,2}$,
  {\bf Cole Hurwitz}$^{21}$, 
  {\bf Mehdi Azabou}$^{5}$, \\
  {\bf Matthew R. Whiteway}$^{5,\dagger}$, 
  {\bf Liam Paninski}$^{5,\dagger}$,
  {\bf Eva L. Dyer}$^{1,\dagger}$
  \vspace{0.5em}
  \\
  {\normalfont\mdseries
  $^{1}$ University of Pennsylvania, 
  $^{2}$ Mila, 
  $^{3}$ Université de Montréal, 
  $^{4}$ Stanford University, \\
  $^{5}$ Columbia University, 
  $^{6}$ Allen Institute, 
  $^{7}$ William James Center for Research, \\
  $^{8}$ ISPA - Instituto Universitário,
  $^{9}$ University of Geneva, 
  $^{10}$ Karolinska Institutet, 
  $^{11}$ UCLA, \\
  $^{12}$ University College London, 
  $^{13}$ Champalimaud Foundation, 
  $^{14}$ Lingang Laboratory, \\
  $^{15}$ The Chinese University of Hong Kong, 
  $^{16}$ Donders Institute, 
  $^{17}$ University of Minnesota, \\
  $^{18}$ Princeton University, 
  $^{19}$ Leiden University, 
  $^{20}$ McGill University, 
  $^{21}$ IBM
  }
  \end{minipage}
}

\addtocontents{toc}{\setcounter{tocdepth}{-10}}

\begin{document}

\maketitle

\input{files/abstract}

\input{files/introduction_v2}

\input{files/dataset}

\input{files/evaluation}

\input{files/ts1}

\input{files/ts2}

\input{files/ts3}

\input{files/results}

\input{files/discussion}

\input{files/acknowledgements}

\bibliographystyle{ieeetr} 
\bibliography{paper}

\newpage
\input{appendix/appendix}

\newpage

\end{document}

%% file: files/abstract.tex
\begin{abstract}
Advances in large-scale neural recording have made it possible to collect data across many animals and distributed brain regions, raising the question of whether this scale can be exploited to learn general-purpose neural representations transferable across diverse downstream tasks. Yet, progress toward this goal has been limited by fragmented evaluation protocols and a narrow focus on individual task domains. Here, we present \brainwidebench, a benchmark for evaluating across-animal transfer on multi-region neural recordings, built on the International Brain Laboratory Brainwide Map dataset of neural and behavioral recordings spanning \numregions~brain regions from \numsubjects~mice performing a sensory-guided decision-making task. The benchmark is organized around three complementary task suites that evaluate whether learned representations support downstream decoding of behavior, can predict masked or future neural activity, and can recover biologically meaningful anatomical organization.
With this benchmark, we systematically evaluate pretraining methods across transfer settings, including finetuning on downstream objectives and zero-shot generalization to unseen animals. Our results confirm pretraining improves performance over matched single-session baselines, but we show current methods exhibit heterogeneity in transfer capabilities: gains depend strongly on the alignment between pretraining objectives and downstream tasks. No single approach performs uniformly well across all three suites, and most methods are designed to only address a subset of them. Together, these findings suggest that learning representations that jointly generalize across behavior, dynamics, and anatomy remains an open challenge. By providing a unified and reproducible evaluation suite, \brainwidebench~establishes a framework for measuring progress toward general-purpose models of the mouse brain. 
Code is available at \href{https://github.com/brainbench-org/ibl-bwb}{this link}.

\end{abstract}

%% file: files/introduction_v2.tex
\section{Introduction}

Over the past decade, advances in large-scale neural recording technologies have fundamentally reshaped systems neuroscience. High-density electrophysiology~\citep{jun2017fully, steinmetz2021neuropixels, ye2025ultra} now enables simultaneous recording of hundreds to thousands of individual neurons across multiple brain regions, producing brain-wide datasets up to single-cell, single-spike resolution across many animals performing complex behaviors~\citep{steinmetz2019distributed, Ottenheimer_2023, chen2024brain, khilkevich2024brain,  international2025brain, findling2025brain}. For the first time, it is possible to observe distributed neural computation across the brain at scale. These datasets promise insight into how perception, decision-making, memory, and action emerge from coordinated population dynamics spanning cortex, thalamus, hippocampus, and beyond. Yet the field remains constrained by analysis approaches that operate at the level of individual sessions or isolated regions, limiting our ability to study shared principles from the growing body of multi-animal recordings~\citep{stringer2024analysis}.

Foundation models offer a fundamentally different way to approach these datasets~\citep{dyer2025accepting,bommasani2021opportunities}. Rather than training separate models for each animal, region, or task, foundation models seek to learn shared representations across heterogeneous recordings  by extracting patterns that recur across individuals and brain areas~\citep{azabou2024unified, ye2024neural, azabou2025multisession, ryoo2025generalizable, zhang2024towards, zhangneural, wang2025foundation,vermani2024meta}. In this view, each recording is not an isolated experiment but a partial observation of a broader underlying computational system. Scaling across animals and anatomical coverage may allow models to distill invariant features of neural dynamics that are not visible within a single session or region. Similar scaling efforts in language and vision have revealed predictable trends in performance and emergent capabilities as data and model size increase~\citep{radford2021learning, hoffmann2022training}. Whether analogous principles hold for neural population activity, and whether large-scale training can uncover shared computational motifs across the brain, remains an open question.

Answering this question requires more than methodological advancements in large-scale modeling of neural data; it requires benchmarks that make scaling and progress measurable. In areas ranging from language modeling, mathematics, and coding to scientific machine learning domains such as biology or materials science, standardized evaluation suites have driven rapid progress by providing curated datasets, unified metrics, and reproducible splits that enable systematic comparison across methods~\citep{donoho201750,zellers2019hellaswag, wang2019superglue, hendrycks2020measuring, cobbe2021training, chen2021evaluating, notin2023proteingym,dunn2020matbench}. These benchmarks have made it possible to quantify scaling behavior, measure transfer, and detect emergent capabilities. 

Existing benchmarks for spike-resolution neural recordings either focus on a small number of animals with limited anatomical coverage or a single predictive task (e.g., neural encoding)~\citep{pei2021neural, karpowicz2024few, willeke2022sensorium}, limiting the ability to study generalization across animals or tasks and distributed computation at scale. 
To our knowledge, no current benchmark exists for \textit{testing across-subject transfer in multi-region neural recordings}, where the goal is to first build a model on pretrained representations and deploy it across \textit{multiple downstream task families} within a unified framework. As a result, the field lacks a standardized platform for measuring how well pretrained models transfer and whether improvements reflect genuine representation learning or task-specific optimization.

To address this gap, we introduce \brainwidebench, a benchmark spanning multiple downstream tasks, built on the International Brain Laboratory’s Brainwide Map dataset~\citep{international2025brain}. 
The dataset comprises over 600 hours of neuron-level, single-spike recordings spanning \numregions~anatomical regions, 
 from \numsubjects~mice performing a vision-guided decision-making task with simultaneous behavioral measurements.
\brainwidebench~defines three complementary  families of tasks that probe distinct properties. The first task suite asks whether {\em behaviorally-relevant variables can be decoded} from neural activity. The second task suite asks whether {\em neural activity can be predicted} from other neurons or past activity. The third task suite asks whether {\em brain regions can be directly inferred} from neural activity. We establish a standardized split for pretraining on many animals and transferring to a held-out set of animals on downstream tasks spanning the three families. To characterize the current landscape of neural population models, we evaluate a diverse set of baselines spanning transformer-based and state-space-based architectures.
This evaluation framework moves beyond evaluation on a single task at a time, toward a more comprehensive notion of representation quality that reflects generalization across animals, brain regions, and tasks.

By combining standardized splits, evaluation of animal-level transfer, and brain-wide anatomical coverage, \brainwidebench~establishes a platform for studying scaling laws, transfer properties, and curriculum learning for pretrained models. 
Along with the task suites, we provide comprehensive session, probe, and unit-level quality control (QC) metadata. Baselines are trained under common QC standards, and full metadata is provided to enable research into how data quality shapes pretraining and model robustness. 
We envision this benchmark as a step toward generalizable, brain-wide models that capture shared principles of neural computation across individuals.

\textbf{The main contributions of this work include:}

\begin{itemize}[leftmargin=6mm,itemsep=2mm,topsep=0mm]
\vspace{-2mm}

\item A new benchmark for evaluating how models pretrained on neural recordings \textbf{transfer across diverse downstream tasks}, organized around three complementary task suites spanning behavior and stimulus decoding, neural activity prediction, and anatomical identification. Together, these tasks evaluate whether learned representations capture transferable behavioral, dynamical, and anatomical structure across animals and recording sessions.

\vspace{-2mm}

\item \textbf{A curated dataset derived from the IBL Brainwide Map}, accompanied by standardized train/test splits and quality-control metadata at the session, probe, and unit level. This supports reproducible evaluation while enabling systematic study of how scale, recording variability, and data quality impact pretraining and transfer.

\vspace{-2mm}  

\item \textbf{A comprehensive empirical study} of supervised, and self-supervised models, spanning both single-session baselines and recent large-scale pretraining approaches. Through evaluations across behavioral, neural, and anatomical tasks, we identify key strengths and limitations of current approaches.

\end{itemize}

\begin{figure}[!t]
    \centering
    \includegraphics[width=0.92\linewidth]{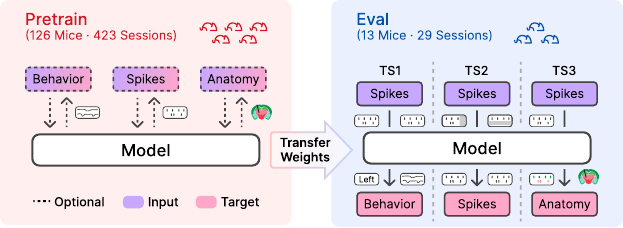}
    \caption{\footnotesize{{\bf Overview of the benchmark.} (Left) During pretraining, models have access to a large pretraining set with 126 mice that includes behavior, spike sorted neural recordings, and anatomical labels (e.g., brain region), to learn generalized representations. Models are pretrained and evaluated (Right) on a held-out set of 13 animals across three task suites: TS1 (decoding), TS2 (neural activity prediction), and TS3 (brain region classification). 
    }
}
\vspace{-2mm}
    \label{fig:benchmark_overview}
\end{figure}

%% file: files/dataset.tex
\section{Motivation}

\subsection{What are the goals of this benchmark?}

Large-scale modeling in neuroscience is rapidly expanding, with a growing number of works exploring pretraining and transfer for neural data~\cite{azabou2024unified,zhang2024towards,ye2024neural,willeke2022sensorium,ryoo2025generalizable}. These approaches have demonstrated promising gains across decoding, forecasting, and cross-session transfer, suggesting that neural activity contains substantial shared structure that can be leveraged through large-scale learning. However, evaluation remains highly fragmented: individual works often introduce new datasets, train/test splits, modalities, or downstream tasks, making it difficult to determine whether improvements reflect genuinely more general representations or simply differences in evaluation protocols.

At the same time, an important open question remains: \emph{what would a truly generalist neural foundation model look like?} Such a model should do more than solve a single decoding or forecasting problem. Instead, it should learn representations that support a diverse set of downstream objectives, including behavioral decoding, neural activity prediction, and anatomical identification, while generalizing across animals, recording sessions, and experimental conditions. 

The goal of \brainwidebench~is therefore two-fold: to provide a standardized and reproducible testbed for developing and comparing pretrained neural models, and to define a set of core downstream tasks that reflect fundamental properties of neural representations. To this end, \brainwidebench~is organized into three complementary task suites. The first evaluates \emph{behavioral decoding}, testing whether representations encode task-relevant variables that generalize across animals and sessions. The second measures \emph{neural activity prediction}, assessing whether models capture structured spatiotemporal population dynamics. The third evaluates \emph{neural identity prediction}, probing whether anatomical organization (brain region) emerges in pretrained representations and transfers without supervision. Together, these task suites test whether a single set of representations, pretrained across many mice and brain regions, transfers to held-out mice on multiple downstream tasks and under a standardized experimental setting.

\subsection{Why this dataset?}

\begin{figure}[!t]
    \centering
    \includegraphics[width=\linewidth]{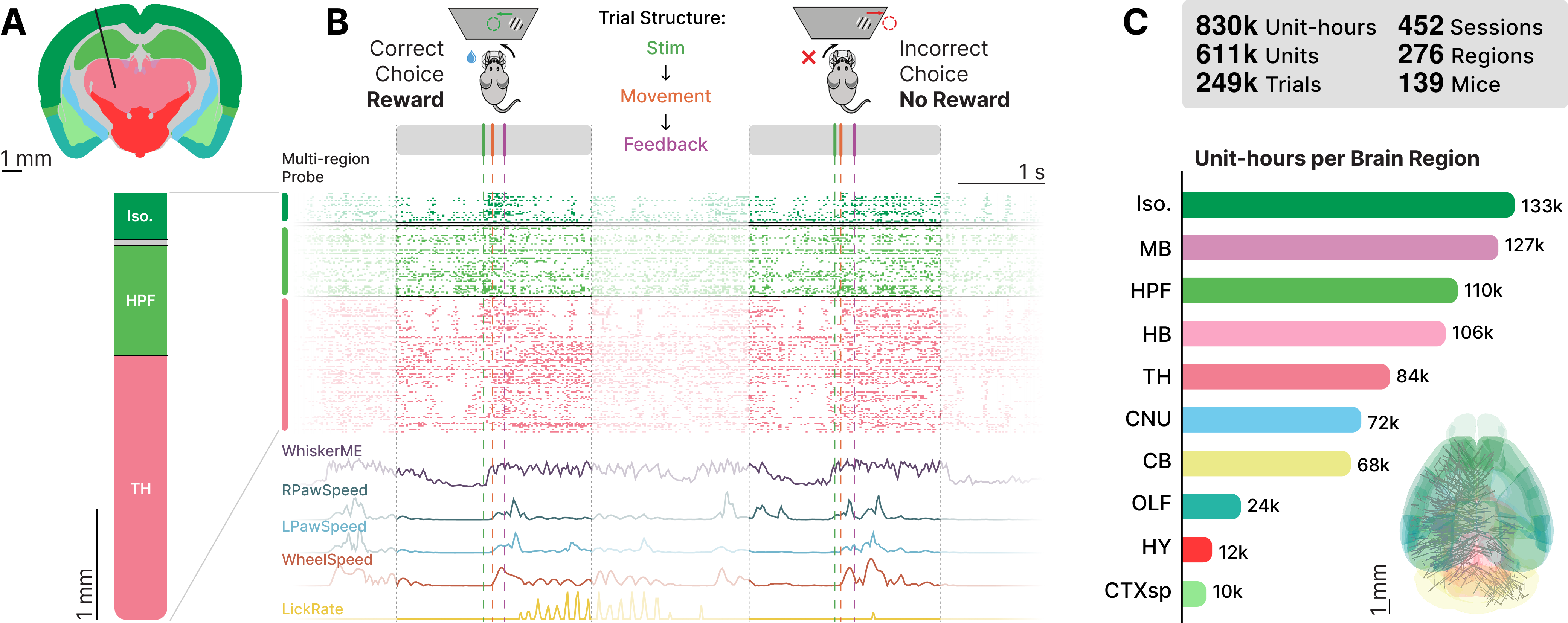}
    \caption{{\bf Overview of IBL Brainwide Map Dataset.} (A) Multi-region probes are inserted into the brain and mapped onto a common coordinate framework (CCF) atlas to map individual neural spike sorted units to different anatomical regions. (B) An example of neural and behavioral data captured simultaneously while the mouse tries to control a wheel to move a visual stimulus from either the left or right side of the screen to the center. We capture paws and wheel speed, whisker motion, and licking from the water spout. After a correct trial, the mouse is rewarded with water. (C) Data from individual probes and sessions are aggregated which in total amounts to over 800,000 neuron hours.
    \label{fig:dataset}}
\end{figure}

The International Brain Laboratory (IBL) Brainwide Map~\citep{international2025brain} provides a rich multi-region dataset that is ideal for benchmarking pretraining strategies for multi-animal, large-scale neural recordings. It is a large-scale, multi-laboratory neurophysiology resource that captures brain-wide neural activity along with detailed behavioral measurements during a standardized decision-making task (Figure~\ref{fig:dataset})~\citep{ibl_decision_task}. A key advantage of this dataset is its scale and consistency. Data were collected across a coordinated collaboration of laboratories using shared experimental protocols, hardware, and preprocessing pipelines, enabling reproducible comparisons across animals, sessions, and brain regions~\citep{abbott2017international, international2023modular,ibl_decision_task,international2025reproducibility}. This level of standardization is critical for evaluating generalization, as it reduces confounds introduced by dataset-specific variability.

The behavioral task involves visual decision-making, where mice respond to Gabor stimuli presented at varying contrasts and spatial locations. Structured manipulations of stimulus visibility and priors on spatial placement induce variability in perception, decision-making, and belief states, providing a rich substrate for probing behaviorally relevant representations. Neural activity is recorded using Neuropixels probes~\citep{jun2017fully}, yielding high-density spiking data from hundreds of neurons simultaneously. Across the dataset, recordings span \numprobes~probe insertions in \numsubjects~mice, covering \numregions~anatomical regions and yielding over 600,000 units if no filters are applied (see Table~\ref{tab:filter_comparison} for unit counts under various filtering criteria). This brain-wide coverage enables evaluation of representations across diverse functional circuits, including sensory, motor, and cognitive systems.

In addition to neural recordings, the dataset includes synchronized behavioral signals such as trial structure, stimulus parameters, wheel movements, and video-based pose estimation~\cite{biderman2024lightning}. Standardized preprocessing pipelines, including spike sorting~\citep{ibl2022spikesorting} and video analysis~\citep{ibl2022video}, align these modalities into a unified format.
The dataset is accessible through the Open Neurophysiology Environment (ONE) API~\citep{international2023modular}, which provides a consistent interface for querying data across sessions, probes, and regions\footnote{Data is processed into the standardized benchmark format using \href{https://torch-brain.readthedocs.io/en/latest/ }{\texttt{torch\_brain}} and released for download \href{https://brain-wide-bench.s3.amazonaws.com/index.html}{here}. We also release the processing pipeline in our codebase for reproducibility.}. This enables scalable and reproducible benchmarking workflows. Together, the scale, standardization, multimodality, and brain-wide coverage of the IBL dataset make it particularly well suited for our benchmark.

%% file: files/evaluation.tex
\section{BrainWideBench Task Suites}
\label{sec:task_suite}

In the following sections, we detail the task suites that constitute the \brainwidebench, outlining motivations and specific design choices around them, as well as representative baselines. 
We characterize pretrained baselines in terms of their training objective: \textit{task-suite-supervised} (\tss) models pretrain directly on the downstream tasks, and \textit{task-suite-unsupervised} (\tsu) models utilize a pretraining objective disjoint from the downstream tasks. To reduce complexity and standardize evaluation, models are tested on a single-session, single-task basis for each task suite. At the same time, users are free to explore different fine-tuning strategies, few-shot adaptation, or fully zero-shot transfer during evaluation. This design supports systematic study of scaling behavior, across-animal transfer, and representation learning in large-scale pretraining for neural data.

\subsection{Across-animal transfer as a test of generalization}

A central challenge in modeling neural data is learning representations that transfer across animals. Each recording provides only a partial and noisy view of a shared latent system underlying behavior and neural computation~\cite{cunningham2014dimensionality,churchland2012neural}. Given the broad anatomical coverage of the Brainwide Map~\cite{international2025brain}, successful transfer further requires models to capture neural structure and dynamics distributed across the brain. To evaluate this capability, \brainwidebench~uses a strict animal-level split between pretraining and evaluation datasets (Appendix~\ref{app:data_splits}), illustrated schematically in Figure~\ref{fig:benchmark_overview}.

This split structure enables evaluation of three complementary capabilities: (1) large-scale representation learning during pretraining, (2) adaptation to new animals with limited supervision, and (3) zero-shot generalization without task-specific fine-tuning. The pretraining corpus consists of recordings from \pretrainsubjects~animals spanning \pretrainsessions~sessions and \pretrainregions~brain regions, with access to all available supervision, including behavioral labels and anatomical annotations. Evaluation is performed on disjoint recordings from \evalsubjects~held-out animals spanning \evalsessions~sessions and \evalregions~regions, ensuring that models must generalize to unseen individuals rather than relying on subject-specific structure.

%% file: files/ts1.tex
\subsection{Task Suite 1: Behavior Prediction}

\paragraph{TS1: Motivation.}
A central goal in systems neuroscience is to understand how neural activity gives rise to behavior \cite{glaser2020machine,paninski2007statistical}. Behavior decoding provides a way to probe this relationship by evaluating how well models can extract behaviorally relevant information from neural population activity. 
From a neuroscience perspective, successful decoding enables the identification of behaviorally relevant signals distributed across populations, offering insights into how information is represented and transformed across circuits \cite{ibl_decision_task,international2025brain}. From an applications perspective, behavior decoding underpins brain-computer interfaces, neural prosthetics, and assistive technologies, where accurate inference of intent or state from neural activity is critical for real-time interaction \cite{pandarinath2018latent,pandarinath2017high,karpowicz2024few}. In the context of large-scale pretraining, decoding tasks test whether pretrained representations generalize across animals, sessions, and tasks while preserving behaviorally meaningful structure \cite{azabou2024unified,ye2025generalistndt3,mao2026unlabelled}.

\vspace{-3mm}
\paragraph{TS1: Tasks.}
This task suite includes eight behavioral decoding tasks derived from the Brainwide Map decision-making task~\citep{ibl_decision_task}. For each decoding task, the target is a 1 second window sampled around key trial events (i.e., stimulus onset, movement onset, feedback time) to ensure that there is both meaningful signal to decode the variable of interest and that we extract windows that are not confounded with other effects of movement (see Figure~\ref{fig:task_intervals_ts1}). In particular, we consider five {\em frame-level regression} tasks capturing continuous motor outputs: licking rate, whisker pad motion energy, wheel speed, left paw speed, and right paw speed (where wheel speed denotes the magnitude of angular velocity and paw speed the magnitude of 2D velocity); and three {\em sequence-level classification} tasks: stimulus contrast, choice, and reward. All of the continuous target variables are resampled at 50 Hz (details in Appendix~\ref{app:data_preprocessing}) and defined at each time point within the target window, producing a sequence-to-sequence prediction problem. See more details in Appendix~\ref{app:ts1}.

\vspace{-3mm}
\paragraph{TS1: Evaluation Setup.}
At evaluation, each session (from an animal unseen during training) is split across time such that the first 40\% of trials fall in the train segment, the next 20\% are in validation, and the remaining 40\% are in test. This temporal split serves two purposes: it eliminates autocorrelation leakage between segments, and it reflects a realistic deployment scenario in which a small portion of data is used to calibrate the model before generating predictions on the remainder. This setting is particularly challenging due to the long duration of recording sessions ($\sim$1.4h average over evaluation sessions), hence models must learn to deal with nonstationarities and shifting probes which can affect the quality of spike sorting. Final metrics are computed on the standardized test split. The train and validation splits are provided as a reference protocol, and we adopt supervised finetuning on all available data for our baseline models (see Appendix~\ref{app:hyperparameters} for finetuning details of individual models). However, the benchmark is designed to support alternative evaluation regimes, including few-shot and zero-shot transfer, enabling future work to explore different adaptation strategies. We measure performance using $R^2$ for regression targets, balanced accuracy for classification targets, and $D^2$ under a Poisson model for Licking Rate (rate variable).


\vspace{-3mm}
\paragraph{TS1: Baselines.}
We evaluate several pretrained models on TS1. We consider models to fall under the \tss~setting if they pretrain on any of the decoding tasks, whereas \tsu~models utilize other forms of supervision or self-supervision during pretraining. In our framework, pretraining methods should produce a \textit{single model} that can be applied downstream on all tasks in the suite. Hence, \tss~models are naturally multi-task or otherwise assume cross-task transfer is achievable within the decoding targets; for this setting, we use POYO+~\cite{azabou2025multisession} and the multi-task version of POSSM~\cite{ryoo2025generalizable}. For \tsu~models, we use NDT-Stitch~\cite{ye2024neural, zhang2024towards} and MtM~\cite{zhang2024towards} which both perform masked neural prediction. Note that MtM also uses anatomical information in its masking scheme. In addition, we benchmark several representative architectures on a single-session, single-task basis with extensive hyperparameter tuning. These single-session baselines establish a strong reference point for the pretrained models on individual decoding tasks, and provide a basis for comparing architectures in a controlled setting. For these baselines, we use Linear, MLP, CNN~\cite{lea2016temporal}, GRU~\citep{cho2014learningphraserepresentationsusing}, CEBRA~\citep{schneider2023cebra}, POYO~\citep{azabou2024unified}, and a supervised version of NDT \citep{ye2021representation}. See more details in Appendix~\ref{app:model_overview}.

%% file: files/ts2.tex
\subsection{Task Suite 2: Neural Activity Prediction}

\paragraph{TS2: Motivation.}
Neural systems operate through dynamic interactions across neurons, where patterns of activity propagate, transform, and ultimately give rise to behavior. 
Modeling these dynamics requires predicting how different neurons or brain regions shape each other's responses, making neural activity prediction a fundamental tool for capturing the temporal structure of population-level computation \cite{pandarinath2018latent,pei2021neural}. 
By learning to predict neural activity, models can reveal the dependencies and interactions that govern circuit function, providing a direct lens on how information flows through the brain \cite{gokcen2022disentangling, xia2025inpaintingneuralpictureinferring}. Neural activity prediction thus serves as a testbed for \emph{in silico} investigation of circuit mechanisms, enabling the construction of surrogate dynamical systems and brain-scale simulators \cite{wang2025foundation, haspel2023time} to probe hypotheses about circuit function. 
At the same time, generative modeling capabilities afforded by such prediction tasks support practical advances, including data augmentation to improve downstream analyses \cite{kapoor2024latentdiffusionneuralspiking, mccart2024diffusionbasedgenerationneuralactivity}, enhanced interpretability through the discovery of latent structure in neural dynamics (e.g.,~Latent Variable Modeling~\cite{pei2021neural,hurwitz2021building}), and real-time, closed-loop interfaces that adapt to evolving neural states \cite{minai2025omisoadaptiveoptimizationstatedependent}.

\vspace{-3mm}
\paragraph{TS2: Tasks.} This task suite includes two neural activity prediction tasks designed to probe complementary aspects of neural representation along both spatial and temporal dimensions.
First, to assess a model's ability to extract features of population-level structure and inter-neuron relationships, we design a \emph{co-smoothing} task \cite{pei2021neural} where a random subset of neurons are masked within the context window, and the model must reconstruct their activity from the remaining observed neurons (Figure~\ref{fig:benchmark_overview}). At test time, the masked neurons form a fixed held-out set, ensuring consistent evaluation across samples.
Next, to probe how well a model can extract fine-grained temporal structure from the population, we design a \emph{forecasting} task with the goal of predicting future neural activity from past observations. This task assesses whether learned representations capture temporal dynamics and support extrapolation beyond the observed context. At test time, the last $200$ms of each sample window are masked and set as the target for models to reconstruct. See more details in Appendix~\ref{app:ts2}.


\vspace{-2mm}
\paragraph{TS2: Evaluation Setup.}
We find the causal train/validation/test splits of TS1 to be especially difficult for neural activity prediction in TS2, limiting our ability to differentiate models. A possible explanation is that non-stationarity in neural recordings makes neuron identity unreliable across long timescales, so predicting the activity of a fixed, indexed set of neurons becomes ill-defined (see Appendix~\ref{app:additional_results_ts2}). Hence, in TS2, we adopt a split structure consisting of interleaved 5-minute temporal blocks, with 40\% of the session in train, 20\% in validation, and 40\% in test. This block structure reduces autocorrelation leakage that would arise from fully randomized splits while ensuring that each subset is large enough to contain representative samples of neural dynamics (see Appendix~\ref{app:ts2} for more details). In both tasks, we evaluate performance using the fraction of deviance explained ($D^2$) under a Poisson observation model and bits-per-spike (bps) \citep{pei2021neural}. $D^2$ can also be termed Deviance Fraction Explained (DFE) as in \citep{xia2025inpaintingneuralpictureinferring}. See more details on these metrics in Appendix~\ref{app:metrics}.


\vspace{-2mm}
\paragraph{TS2: Baselines.}
In TS2, MtM~\cite{zhang2024towards} serves as a representative \tss~model since it explicitly trains on masked neural prediction objectives related to co-smoothing and forecasting. We also evaluate NDT-Stitch~\cite{ye2024neural,zhang2024towards}, which pretrains using temporal masked prediction but in a non-causal setting, making it fall under \tsu~even with respect to forecasting. In addition, we include single-session baselines using a standard autoencoder (AE)~\cite{doi:10.1126/science.1127647}, LFADS~\cite{pandarinath2018latent}, and NDT~\cite{ye2021representation}. Following the Neural Latents Benchmark~\cite{pei2021neural}, these models are treated as latent variable models~\cite{hurwitz2021building}: models are fit to training data to learn population representations, which are then evaluated on co-smoothing and forecasting tasks. While TS2 also supports direct optimization on the downstream tasks themselves, we focus on representation quality as the primary evaluation objective. See Appendix~\ref{app:model_overview} for additional details. We also test simple statistical baselines that use training-set neural activity statistics, providing reference floors for how much co-smoothing and forecasting performance can be explained without learning a representation, and include these results in Appendix~\ref{app:ts2_statistical}.

%% file: files/ts3.tex
\subsection{Task Suite 3: Neuron Identity Prediction}

\paragraph{TS3: Motivation.}

A neuron's identity, including its anatomical location and cell type, strongly shapes its activity patterns across behaviors and cognitive states~\cite{steinmetz2019distributed}. Experimentally, identifying these properties typically relies on probe localization through histology and expert annotation~\cite{liu2021accurate}. However, these procedures are expensive, cannot be performed in vivo, and are subject to inter-rater variability~\cite{carey2023deepslice}, motivating the need for automated approaches that can infer neuronal identity directly from neural activity.

From a computational perspective, solving this task  requires learning representations that capture biologically meaningful latent structure from neural signals alone. Prior work has shown that anatomical region and cell type can, to some extent, be decoded from activity patterns~\cite{arora2025know,yu2025in,schneider2023transcriptomic}. However, the Brainwide Map dataset~\cite{international2025brain} presents a substantially more challenging setting due to its broad anatomical coverage, large inter-animal variability, and diversity of recording conditions.

Motivated by these challenges, TS3 evaluates whether models recover intrinsic and biologically grounded structure from neural activity. Because brain regions differ in their connectivity, functional roles, cell type composition, and population statistics, they induce consistent signatures in neural dynamics that persist across animals and experimental conditions~\cite{tolossa2025neurons,schneider2025robust}. We therefore evaluate brain region identification in a zero-shot transfer setting to test two complementary goals: whether learned representations uncover anatomical structure without explicit supervision, and whether they support practical in vivo localization in previously unseen animals.



\vspace{-2mm}
\paragraph{TS3: Tasks.}
Given neural activity from individual units, the task is to predict the associated anatomical region label. We align all probes to the Allen Common Coordinate Framework (CCF)~\cite{wang2020allen} and use Cosmos-level annotations from the IBL anatomical atlas~\cite{international2025brain}, resulting in a $10$-class classification problem spanning coarse but functionally meaningful regions such as \textit{Isocortex}, \textit{Hippocampus}, and \textit{Cerebellum}. 
We evaluate both the {\em single-unit} setting, where neurons are classified independently, and the {\em multi-unit} setting, where predictions are informed by neighboring units on the same probe. 
All evaluations are performed on held-out animals in a zero-shot setting, requiring generalization across recording sessions, probe placements, and inter-animal variability without using region labels from evaluation sessions during training. See Appendix~\ref{app:ts3} for details.

\vspace{-2mm}
\paragraph{TS3: Evaluation Setup.}
We consider two across-animal evaluation settings designed to probe different aspects of generalization. The first is the \emph{Transductive Zero-Shot} setting, where models are adapted on held-out animals using some form of supervision external to the brain region labels (e.g.,~behavioral labels as in POYO+), allowing them to adjust representations to the new domain before evaluation. This setting reflects realistic scenarios where some data from a new subject is available for adaptation. The second is the \emph{Inductive Zero-Shot} setting, where pretrained models are evaluated on entirely unseen animals and probe configurations without any adaptation, providing a stringent test of whether learned representations capture invariant features of neural identity. Performance is measured using macro-averaged F1 score to account for class imbalance.

\vspace{-2mm}
\paragraph{TS3: Baselines.}
We evaluate methods for learning unit-level representations, including NEMO~\cite{yu2025in} and NuCLR~\cite{arora2025know}, by pretraining them using their respective training objectives. Since these models are capable of generating embeddings for a new neural population via an inference step, they are representative baselines for the Inductive setting. In addition, we provide a simple baseline for the inductive setting using ISI features directly as embeddings. We also use LOLCAT from~\cite{tolossa2025neurons, schneider2023transcriptomic} which is directly trained on the brain region classification task, without any pretraining. Since LOLCAT extracts features directly using the brain region classification task for supervision, it represents a \tss~baseline for TS3, whereas NEMO and NuCLR fall under \tsu. Finally, we evaluate unit-level embeddings from models originally pretrained for TS1 (POYO+~\cite{azabou2025multisession}, POSSM~\cite{ryoo2025generalizable}) and TS2 (NDT-Stitch~\cite{ye2024neural,zhang2024towards} and MtM~\cite{zhang2024towards}) as representative models for the Transductive setting. These models are finetuned on eval sessions using their pretraining objectives, as they require gradients to generate embeddings on the new neural population. See more details in Appendix~\ref{app:model_overview}.

%% file: files/results.tex
\section{Benchmark Results}

\subsection{TS1 Results}
In Table~\ref{tab:ts1_main}, we compare behavior decoding performance across pretrained models on both frame-level and sequence-level tasks. Unsurprisingly, \tss~models pretrained directly on the decoding tasks (green) generally outperform \tsu~models pretrained on a different objective (magenta). However, across-task transfer remains surprisingly effective on frame-level tasks such as whisker and right paw, and sequence-level tasks such as reward and choice decoding. This suggests that pretrained representations capture latent neural structure that generalizes beyond their original supervision.

We also compare with single-session models to benchmark performance against representative architectures, though these require extensive tuning. Notably, POYO and NDT are both improved by their pretrained counterparts: POYO improves in average rank over tasks from $4.80 \rightarrow 2.53$ with POYO+, and NDT improves from $6.80 \rightarrow 3.58$ with NDT-Stitch (ranking procedure detailed in Appendix~\ref{app:rank_procedure}). Among the single-session baselines, recurrent and convolutional models achieve strong performance across most tasks, particularly for high-signal variables such as licking and whisker motion. Nonetheless, POYO+ achieves the highest average rank among our baselines, and POSSM the second highest. Together, these results indicate that pretraining results in representations that transfer well across animals and decoding objectives.

\input{tables/table_task1}

\input{tables/table_task2}

\input{tables/table_task3}

\subsection{TS2 Results}
In Table~\ref{tab:table2}, we evaluate neural activity prediction using both co-smoothing and forecasting tasks. Across all metrics, pretrained models outperform the single-session baselines, demonstrating that large-scale pretraining improves the ability to capture structured neural dynamics. In particular, NDT-Stitch achieves the strongest forecasting performance, while MtM performs best on co-smoothing, suggesting that different pretraining objectives emphasize complementary aspects of neural dynamics. Interestingly, NDT-Stitch is effective on forecasting despite being pretrained using a non-causal masked prediction objective, suggesting that representations learned through temporal reconstruction transfer well even with shift in the task distribution. Conversely, MtM performs particularly well on co-smoothing, likely due to the fact that for most training steps it employs a spatial masking strategy, hence a closer alignment to the downstream objective.


\subsection{TS3 Results}
In Table~\ref{tab:ts3_main}, we evaluate brain region identification across held-out animals in both transductive and inductive settings. Across all evaluation regimes, methods specifically designed to learn neuron-level representations substantially outperform models that optimize for a different objective, such as behavioral decoding or neural prediction. In particular, NuCLR achieves the strongest performance across both single-unit and multi-unit settings, reaching a macro-F1 of $0.654$ in the multi-unit linear probe setting. NEMO also performs strongly with a peak macro-F1 of $0.605$ with a multi-unit MLP probe. Both of these models focused on extracting invariant features of neuron identity. On the other hand, transductive methods such as POYO+, POSSM, NDT-Stitch, and MtM achieve substantially lower performance, suggesting that representations optimized for other objectives do not automatically organize around anatomical structure. Nevertheless, NDT-Stitch and MtM outperform POYO+ and POSSM, indicating that masked neural prediction may preserve more region-specific structure than behaviorally-aligned decoding objectives.

We also observe consistent improvements from single-unit to multi-unit evaluation, suggesting that neighboring neurons contain complementary information that improves localization accuracy. Interestingly, LOLCAT underperforms compared to some \tsu~baselines despite optimizing its latents for the downstream task. Overall, these results demonstrate that anatomical structure can emerge from neural activity representations in a zero-shot setting, while also highlighting the importance of pretraining objectives that explicitly encourage stable neuron-level representations across animals.

%% file: tables/table_task1.tex
\begin{table}[t]
\centering
\small
\caption{\footnotesize{\bf Task Suite 1: Behavior decoding performance.} Single-session models are trained and tested on trials from the same session. Pretrained models are finetuned on the new sessions. Sequence-level tasks are reported as Balanced Accuracy (\%); frame-level task performance is reported as $R^2$, except for Licks, which is an event stream, where we report $D^2$ instead. All metrics are reported as the average over evaluation sessions ($\tpms{\text{SEM}}$ over 5 finetuning seeds). 
Rankings incorporate statistical significance, see Appendix~\ref{app:rank_procedure} for more details. 
Green shading denotes a \tss~baseline, while magenta denotes \tsu. No shading denotes a single session baseline tuned and tested on the same session.}
\vspace{2mm}
\resizebox{\columnwidth}{!}{
\begin{tabular}{l|l|ccccc|ccc|R}

\toprule

& & \multicolumn{5}{c|}{{\bf Frame-level Tasks}} & \multicolumn{3}{c|}{{\bf Sequence-level Tasks}} & \multicolumn{1}{R}{{Avg}} \\
& Method
& Licks ($D^2$)
& Whisker ($R^2$)
& Wheel ($R^2$)
& RPaw ($R^2$)
& LPaw ($R^2$)
& Reward (Acc)
& Choice (Acc)
& Contrast (Acc)
& Rank \\

\midrule

\multirow{7}{*}{\rotatebox[origin=c]{90}{\textbf{Single-Session}}}
& Linear 
& $0.151 \tpms{0.000}$
& $0.175 \tpms{0.000}$
& $0.219 \tpms{0.002}$
& $0.091 \tpms{0.000}$
& $0.084 \tpms{0.001}$
& $0.813 \tpms{0.004}$
& $0.614 \tpms{0.004}$
& $0.219 \tpms{0.001}$
& 7.73\\

& MLP 
& $0.348 \tpms{0.001}$
& $0.268 \tpms{0.002}$
& $0.329 \tpms{0.001}$
& $0.157 \tpms{0.001}$
& $0.135 \tpms{0.001}$
& $0.877 \tpms{0.006}$
& $0.636 \tpms{0.004}$
& $0.224 \tpms{0.002}$
& 5.46\\

& GRU 
& $0.510 \tpms{0.002}$
& $0.358 \tpms{0.002}$
& $0.300 \tpms{0.006}$
& $0.214 \tpms{0.002}$
& $0.176 \tpms{0.002}$
& $0.885 \tpms{0.005}$
& $0.644 \tpms{0.007}$
& $0.214 \tpms{0.001}$
& 3.96\\

& CNN 
& $0.549 \tpms{0.008}$
& $0.389 \tpms{0.004}$
& $0.330 \tpms{0.004}$
& $0.225 \tpms{0.004}$
& $0.194 \tpms{0.002}$
& $0.887 \tpms{0.004}$
& $0.621 \tpms{0.002}$
& $0.205 \tpms{0.002}$
& 3.74\\

& CEBRA 
& $0.327 \tpms{0.003}$
& $0.280 \tpms{0.003}$
& $0.200 \tpms{0.002}$
& $0.164 \tpms{0.004}$
& $0.121 \tpms{0.002}$
& $0.780 \tpms{0.008}$
& $0.521 \tpms{0.002}$
& $0.198 \tpms{0.001}$
& 7.44\\

& NDT 
& $0.314 \tpms{0.001}$
& $0.303 \tpms{0.002}$
& $0.212 \tpms{0.003}$
& $0.163 \tpms{0.002}$
& $0.127 \tpms{0.001}$
& $0.871 \tpms{0.005}$
& $0.616 \tpms{0.003}$
& $0.208 \tpms{0.003}$
& 6.80\\

& POYO 
& $0.559 \tpms{0.002}$
& $0.291 \tpms{0.004}$
& $0.300 \tpms{0.003}$
& $0.195 \tpms{0.002}$
& $0.136 \tpms{0.002}$
& $0.882 \tpms{0.002}$
& $0.630 \tpms{0.005}$
& $0.224 \tpms{0.002}$
& 4.80\\

\midrule

\multirow{4}{*}{\rotatebox[origin=c]{90}{\textbf{Pretrain}}}
& POYO+ 
& \cellcolor{bwbTSS} $0.678 \tpms{0.003}$
& \cellcolor{bwbTSS} $0.380 \tpms{0.003}$
& \cellcolor{bwbTSS} $0.362 \tpms{0.001}$
& \cellcolor{bwbTSS} $0.249 \tpms{0.002}$
& \cellcolor{bwbTSS} $0.200 \tpms{0.000}$
& \cellcolor{bwbTSS} $0.898 \tpms{0.003}$
& \cellcolor{bwbTSS} $0.689 \tpms{0.003}$
& \cellcolor{bwbTSS} $0.267 \tpms{0.001}$
& 2.53\\

& POSSM 
& \cellcolor{bwbTSS} $0.661 \tpms{0.002}$
& \cellcolor{bwbTSS} $0.351 \tpms{0.003}$
& \cellcolor{bwbTSS} $0.355 \tpms{0.002}$
& \cellcolor{bwbTSS} $0.237 \tpms{0.001}$
& \cellcolor{bwbTSS} $0.175 \tpms{0.002}$
& \cellcolor{bwbTSS} $0.887 \tpms{0.004}$
& \cellcolor{bwbTSS} $0.662 \tpms{0.005}$
& \cellcolor{bwbTSS} $0.220 \tpms{0.002}$
& 3.18\\

& NDT-Stitch 
& \cellcolor{bwbTSU} $0.525 \tpms{0.002}$
& \cellcolor{bwbTSU} $0.384 \tpms{0.003}$
& \cellcolor{bwbTSU} $0.310 \tpms{0.002}$
& \cellcolor{bwbTSU} $0.219 \tpms{0.005}$
& \cellcolor{bwbTSU} $0.172 \tpms{0.003}$
& \cellcolor{bwbTSU} $0.863 \tpms{0.006}$
& \cellcolor{bwbTSU} $0.674 \tpms{0.001}$
& \cellcolor{bwbTSU} $0.229 \tpms{0.001}$
& 3.58\\

& MtM 
& \cellcolor{bwbTSU} $0.469 \tpms{0.004}$
& \cellcolor{bwbTSU} $0.362 \tpms{0.004}$
& \cellcolor{bwbTSU} $0.323 \tpms{0.006}$
& \cellcolor{bwbTSU} $0.224 \tpms{0.003}$
& \cellcolor{bwbTSU} $0.167 \tpms{0.003}$
& \cellcolor{bwbTSU} $0.874 \tpms{0.003}$
& \cellcolor{bwbTSU} $0.646 \tpms{0.004}$
& \cellcolor{bwbTSU} $0.213 \tpms{0.002}$
& 4.05\\

\bottomrule

\end{tabular}
}
\label{tab:ts1_main}
 \vspace{-2mm}

\end{table}

%% file: tables/table_task2.tex
\begin{table}[t!]
\centering
\small
\caption{\footnotesize{{\bf Task Suite 2: Co-smoothing and forecasting performance.} Model performance for both tasks is reported in terms of $D^2$ and bps. All metrics are reported as the average over evaluation sessions ($\tpms{\text{SEM}}$ over 5 finetuning seeds). 
Rankings incorporate statistical significance, see Appendix~\ref{app:rank_procedure} for more details. 
Green shading denotes a \tss~baseline, while magenta denotes \tsu. No shading denotes a single session baseline tuned and tested on the same session.}}
\vspace{2mm}
\resizebox{0.8\columnwidth}{!}{
\begin{tabular}{+c|^l|^c^c|^c^c|^R}
\toprule
& \textbf{Method} 
& \multicolumn{2}{c|}{\bf Co-smoothing}
& \multicolumn{2}{c|}{\bf Forecasting}
& \multicolumn{1}{R}{Avg} \\
& 
& $D^2$ & bps 
& $D^2$ & bps
& Rank \\\midrule

\multirow{3}{*}{\rotatebox[origin=c]{90}{\textbf{SS}}}






& Autoencoder 
& $0.088 \tpms{0.000}$ 
& $0.200 \tpms{0.001}$ 
& $0.037 \tpms{0.001}$ 
& $0.065 \tpms{0.001}$ 
& 4.81 \\

& LFADS 
& $0.176 \tpms{0.000}$ 
& $0.421 \tpms{0.001}$ 
& $0.144 \tpms{0.000}$ 
& $0.325 \tpms{0.001}$ 
& 2.34 \\

& NDT 
& $0.132 \tpms{0.001}$ 
& $0.304 \tpms{0.003}$ 
& $0.158 \tpms{0.000}$ 
& $0.343 \tpms{0.001}$ 
& 2.79 \\

\midrule
\multirow{2}{*}{\rotatebox[origin=c]{90}{\textbf{Pre}}}

& NDT-Stitch
& \cellcolor{bwbTSU}$0.157 \tpms{0.001}$ 
& \cellcolor{bwbTSU}$0.369 \tpms{0.002}$ 
& \cellcolor{bwbTSU}$0.177 \tpms{0.000}$ 
& \cellcolor{bwbTSU}$0.384 \tpms{0.000}$ 
& 1.64 \\

& MtM
& \cellcolor{bwbTSS}$0.191 \tpms{0.001}$ 
& \cellcolor{bwbTSS}$0.459 \tpms{0.002}$ 
& \cellcolor{bwbTSS}$0.131 \tpms{0.001}$ 
& \cellcolor{bwbTSS}$0.288 \tpms{0.001}$ 
& 2.57 \\

\bottomrule

\end{tabular}
}
\vspace{-2mm}
\label{tab:table2}
\end{table}


%% file: tables/table_task3.tex
\begin{table}[t!]
\centering
\small
\caption{\footnotesize{{\bf Task Suite 3: Brain region prediction  performance.} Results are reported as the macro-average F1 score ($\tpms{\text{SEM}}$ over 5 seeds).
For learned embeddings, seeds parameterize embedding generation, i.e.,~through target-session calibration (finetuning) for transductive models, and through pretraining or supervised training for inductive models. Methods are evaluated at both Single Unit  and Multi-Unit levels across Linear and MLP probes. All models are evaluated in zero-shot transfer with respect to the region labels after pretraining. Green shading denotes a \tss~baseline, while magenta denotes \tsu.
We omit shading for the ISI baseline since its embeddings are not trained, and omit SEM on its linear probe since it is fully deterministic.
}}
\vspace{2mm}
\resizebox{0.82\columnwidth}{!}{
\begin{tabular}{l  l c c c c}
\toprule
& & \multicolumn{2}{c}{\textbf{Single Unit} (F1)} & \multicolumn{2}{c}{\textbf{Multi-Unit} (F1)} \\
\cmidrule(lr){3-4} \cmidrule(lr){5-6}
\textbf{Regime} & \textbf{Method} & \textbf{Linear Probe} & \textbf{MLP Probe} & \textbf{Linear Probe} & \textbf{MLP Probe} \\
\midrule
\multirow{4}{*}{{\bf Transductive}}
& POYO+
& \cellcolor{bwbTSU}$0.099\tpms{0.003}$ 
& \cellcolor{bwbTSU}$0.120\tpms{0.005}$ 
& \cellcolor{bwbTSU}$0.072\tpms{0.001}$ 
& \cellcolor{bwbTSU}$0.132\tpms{0.008}$ 
\\

& POSSM
& \cellcolor{bwbTSU}$0.103\tpms{0.002}$ 
& \cellcolor{bwbTSU}$0.123\tpms{0.003}$ 
& \cellcolor{bwbTSU}$0.104\tpms{0.004}$ 
& \cellcolor{bwbTSU}$0.133\tpms{0.006}$ 
\\

& NDT-Stitch
& \cellcolor{bwbTSU}$0.136\tpms{0.007}$ 
& \cellcolor{bwbTSU}$0.167\tpms{0.007}$ 
& \cellcolor{bwbTSU}$0.155\tpms{0.011}$ 
& \cellcolor{bwbTSU}$0.187\tpms{0.008}$ 
\\

& MtM
& \cellcolor{bwbTSU}$0.137\tpms{0.004}$ 
& \cellcolor{bwbTSU}$0.140\tpms{0.006}$ 
& \cellcolor{bwbTSU}$0.159\tpms{0.009}$ 
& \cellcolor{bwbTSU}$0.158\tpms{0.007}$ 
\\

\midrule

\multirow{4}{*}{{\bf Inductive}}

& ISI Baseline
& $0.263 $
& $0.380\tpms{0.002}$
& $0.355 $
& $0.519\tpms{0.006}$
\\

& LOLCAT
& \cellcolor{bwbTSS}$0.359\tpms{0.003}$ 
& \cellcolor{bwbTSS} -- 
& \cellcolor{bwbTSS}$0.473\tpms{0.004}$ 
& \cellcolor{bwbTSS} -- 
\\

& NEMO
& \cellcolor{bwbTSUind}$0.447\tpms{0.003}$ 
& \cellcolor{bwbTSUind}$0.471\tpms{0.002}$ 
& \cellcolor{bwbTSUind}$0.579\tpms{0.007}$ 
& \cellcolor{bwbTSUind}$0.605\tpms{0.004}$ 
\\

& NuCLR
& \cellcolor{bwbTSUind}$0.616\tpms{0.006}$ 
& \cellcolor{bwbTSUind}$0.626\tpms{0.005}$ 
& \cellcolor{bwbTSUind}$0.654\tpms{0.010}$ 
& \cellcolor{bwbTSUind}$0.645\tpms{0.003}$ 
\\

\bottomrule

\end{tabular}
}
\vspace{-1mm}
\label{tab:ts3_main}
\end{table}

%% file: files/discussion.tex
\section{Discussion}



This benchmark evaluates model generalization across three complementary axes: behavior, dynamics, and anatomy, providing a unified view of how large-scale models pretrained on neural data transfer across animals and downstream tasks. Across these axes, we observe a consistent and encouraging trend: pretrained models can act as strong generalist representations that rapidly adapt to new animals and recording conditions. At the same time, performance differences across task suites reveal important distinctions in what current pretraining objectives capture most effectively and where gaps still remain.

In TS1, pretrained behavioral models such as POYO+ and POSSM substantially outperform self-supervised approaches on downstream decoding tasks, suggesting that behaviorally-aligned objectives remain highly beneficial for extracting task-relevant neural structure. In TS2, pretraining consistently improves co-smoothing and forecasting performance relative to single-session latent variable baselines, demonstrating the value of large-scale data for modeling neural population dynamics. In TS3, we find
that zero-shot brain region decoding across animals is possible, with self-supervised methods such as NuCLR and NEMO achieving strong performance without adaptation to evaluation animals. Together, these results suggest that pretrained models can uncover structure at scale that generalizes across individuals and probe locations, while also highlighting important differences in what current pretraining objectives capture most effectively. In particular, the gap between supervised and self-supervised approaches in TS1 highlights opportunities for SSL methods to better align their representations with behavior, while the strong performance of neuron-level SSL methods in TS3 suggests that preserving stable unit identity may be important for across-animal transfer.

More broadly, our results suggest that generalist neural models may help amortize the substantial cost of training and tuning models for individual sessions and tasks. Achieving strong single-session performance often requires extensive hyperparameter optimization, whereas pretrained models can be adapted to new animals and tasks using a largely standardized fine-tuning setup (see Appendix~\ref{app:computing_resources} for a breakdown of compute requirements per model, where we see single-session tuning is substantially more compute-intensive). Across our experiments, pretrained models were finetuned using minimal task-specific tuning, yet achieved competitive or superior performance across all settings. This suggests that large-scale pretraining not only improves transfer, but may also reduce the engineering burden required to deploy neural models across diverse datasets and experimental conditions. Still, important gaps remain in the ability of current models to transfer across new sessions and neural populations. For example, pretrained behavioral decoding and neural dynamics models still require target-session calibration to learn parameters specific to the new recording population, such as unit and session embeddings or session-specific read-in/read-out stitchers. Developing models for these tasks that require minimal or zero-shot calibration is a natural direction for future work that \brainwidebench~is well suited to evaluate.

These conclusions should be interpreted in light of several limitations of the current benchmark. While \brainwidebench~tests generalization across animals, brain regions, and downstream tasks, evaluation is limited to a single species (mouse), recording modality (Neuropixels electrophysiology), and behavioral paradigm (visually-guided decision-making). Furthermore, the held-out set was chosen to keep evaluation feasible while preserving broad anatomical coverage, but the result is a modest cohort of 13 animals. Thus, rankings are more reliable in aggregate than in fine-grained comparison: broad groupings of models hold up consistently across resampled splits, but rankings within a group are more sensitive to which animals are held out and should not be read as statistically distinguishable pointwise comparisons. Similarly, the categorical labels used throughout the benchmark (TSS versus TSU, full finetuning versus few-shot versus zero-shot, transductive versus inductive) organize model comparisons but do not fully capture the design choices and inductive biases built into each. Bringing disparate model families onto common ground for evaluation is precisely the benchmark's aim, but these categories remain an imperfect proxy; model performance is therefore best read in aggregate across suites and with attention to each model's underlying design choices, rather than as isolated, pairwise rankings. Within these constraints, the benchmark is best used to diagnose where current approaches generalize and where they fall short.

\brainwidebench~provides a framework for systematically measuring progress toward general-purpose models of neural activity. By absorbing the substantial cost of baseline training, dataset curation and standardization into a single, fully documented evaluation suite, \brainwidebench~allows researchers to focus on modeling and scientific questions, accelerating progress toward models that are genuinely useful for neuroscience.

%% file: files/acknowledgements.tex
\section*{Acknowledgments and Disclosure of Funding}

We thank Joel Ye for insightful discussions on NDT and NDT2, and Reid Laughton, Zihao Chen, Noah Foster, Max Mercado, Le Thuy Duong Nguyen, Alon Saguy, Skylar Tian, and Ji Xia for early testing and validation of the benchmark. We would also like to thank Spyridon Mylonas and Lena Mei from the NSF AI Institute for Artificial and Natural Intelligence (ARNI) for their support throughout this project.
This work was supported by funds provided by the National Science Foundation and by DoD OUSD (R\&E) under Cooperative Agreement DBI-2229929 (The NSF AI Institute for Artificial and Natural Intelligence). This work was also supported by the Fonds de recherche du Québec (Secteur NT, 2009130), the National Science Foundation (NSF CAREER Award RI:2146072 and NSF 1707398), the Gatsby Charitable Foundation (GAT3708), the Simons Foundation (543023), the Wellcome Trust (216324), the National Institutes of Health (U19NS123716, R00NS128075, and 1R50NS145433), the Sloan Research Fellowship, the Leopoldina Fellowship, the Natural Sciences and Engineering Research Council of Canada (NSERC Discovery Grant RGPIN-2020-05105), the CIFAR Learning in Machines and Brains Program, the Canada CIFAR AI Research Chair program, the Canada Research Chair in Neural Computations and Interfacing, and Zuckerman Institute Team Science.

This work would not be possible without the computing resources provided by the Penn Advanced Research Computing Center (PARCC) at the University of Pennsylvania. Thank you to Jaime Combariza and Ken Chaney for their support. This work used resources available through the National Research Platform (NRP) \cite{10.1145/3708035.3736060} at the University of California, San Diego. NRP has been developed, and is supported in part, by funding from National Science Foundation, from awards 1730158, 1540112, 1541349, 1826967, 2112167, 2100237, and 2120019, as well as additional funding from community partners. 


%% file: appendix/appendix.tex
\addtocontents{toc}{\setcounter{tocdepth}{1}}

\appendix
\newpage
\setcounter{figure}{0}
\setcounter{table}{0}
\renewcommand{\thefigure}{A\arabic{figure}}
\renewcommand{\thetable}{A\arabic{table}}

\begin{center}
    {\LARGE \textbf{Supplementary Material}}\\[0.5em]
    \rule{0.55\linewidth}{0.4pt}\\[0.7em]
    {\large \brainwidebench: 
    Benchmarking large-scale pretraining and  \\[0.15em] 
    across-animal transfer in multi-region neural recordings.}
\end{center}
\vspace{1.5em}

\tableofcontents
\newpage


\label{sec:appendix}

\input{appendix/data_qc}

\newpage

\input{appendix/data_processing}

\newpage

\input{appendix/data_splits}

\newpage

\input{appendix/task_suites}

\clearpage
\newpage

\input{appendix/metrics}

\newpage

\input{appendix/model_overview}

\newpage

\input{appendix/hyperparameters}

\newpage

\input{appendix/computing}

\newpage

\input{appendix/additional_results}

\newpage

\input{appendix/related_work}

\newpage

\input{appendix/contribution_statement}

%% file: appendix/data_qc.tex
\section{Data Quality Control}
\label{app:data_qc}

Large-scale neurophysiology datasets of the scale and complexity of the IBL Brainwide Map (BWM) inevitably contain data of variable quality across sessions and modalities. Recording artifacts, hardware failures, signal extraction issues (e.g., spike sorting and pose estimation), and imperfect behavioral engagement can all degrade data quality and vary in severity across sessions. Rather than applying a single binary data quality filter to each session, we characterize quality along multiple independent axes: session-level behavior; neural activity, including raw electrophysiology, spike sorting, probe alignment and individual unit quality; video-based behavior signals; and trial-level behavior. We make all quality metrics available as part of the benchmark release, enabling downstream users to apply more or less stringent filters as appropriate for their use case. We describe each quality metric below, then detail how these metrics are combined to construct the pretraining and eval splits in Appendix~\ref{app:data_splits}.

\subsection{Session-level Quality Control} 
Sessions were included in the BWM data release~\cite{bwmdata2022figshare} if the mouse performed at least 250 trials, with performance of at least 90\% correct on 100\%-contrast trials for both left and right stimulus blocks. Sessions were additionally required to pass a collection of hardware diagnostic tests, whose precise definitions are available at \url{https://int-brain-lab.github.io/iblenv/_autosummary/ibllib.qc.task_metrics.html}. These criteria constitute a baseline behavioral inclusion criterion applied uniformly across the full dataset prior to any further quality assessment, and therefore all sessions in the benchmark dataset pass these controls.

\subsection{Neural Quality Control}

\paragraph{Probe Alignment (\texttt{qc\_neural\_alignment}).}

The alignment quality was determined by running a brain-region decoder algorithm (not yet published), which assigns a brain region label to each channel of a probe insertion according to electrophysiology signatures (e.g., power in the Local Field Potential). By comparing the output of this decoder and the brain region label originally annotated by experimenters, the severity of the discrepancy was labeled as \texttt{HIGH}, \texttt{LOW} or \texttt{NONE}. \texttt{HIGH} severity alignments are marked as \texttt{FAIL}; \texttt{LOW} severity alignments are marked as \texttt{WARNING}; and alignments with \texttt{NONE} severity are marked \texttt{PASS}.

Alignment quality was not used to determine the pretraining set, but it did shape the evaluation set:
a subject was eligible for evaluation only if each of its sessions contained at least one probe with both \texttt{qc\_overall} and \texttt{qc\_neural\_alignment} marked \texttt{PASS}, and evaluation sessions retaining no such probe were discarded. 
Because insertions are extracted per session rather than individually, the released evaluation build still contains 4 of 45 insertions whose alignment is marked \texttt{FAIL}. 
These are excluded at scoring time: Task Suite~3 (brain region identification), where accurate anatomical labeling of units is essential, scores only units on probes whose alignment is marked \texttt{PASS}. 
This excludes both \texttt{FAIL} and \texttt{WARNING} alignments, the latter affecting only pretraining sessions, since no evaluation probe carries a \texttt{WARNING} alignment.

\paragraph{Raw Electrophysiology (\texttt{qc\_neural\_raw}).} 
Probes were manually marked as \texttt{PASS} / \texttt{WARNING} / \texttt{FAIL} depending on whether images presenting short snippets of raw destriped action potential (AP) data presented noise or artifacts, as done previously for the Brainwide Map data release~\citep{ibl2022spikesorting}.

\paragraph{Spike Sorting (\texttt{qc\_neural\_spikesorting}).} 
Probes were manually marked as \texttt{PASS} / \texttt{WARNING} / \texttt{FAIL} depending on whether the spike raster presented discontinuity, jitter, horizontal band and loss of spikes, as done previously for the Brainwide Map data release~\citep{ibl2022spikesorting}. Insertions presenting such issues in a very mild manner, or not at all, were marked as \texttt{PASS}.

\paragraph{Individual Units.} Units produced by the spike sorting pipeline~\citep{ibl2022spikesorting} were assessed for quality using three criteria established in prior work~\citep{international2025reproducibility}: (i) median spike waveform amplitude greater than 50 $\mu$V; (ii) noise cut-off below 5 (a.u.); and (iii) absence of refractory period violations. Units passing all three criteria are designated as ``good'' units. Unit-level quality is assessed for every session regardless of the session-level raw electrophysiology or spike sorting outcome. 
All units, ``good'' and otherwise, are released as part of the benchmark. Good-unit filtering is applied during evaluation of Task Suites 2 and 3 (neural activity prediction and brain region identification), where unit quality directly affects the validity of the neural signal. Task Suite 1 (behavioral decoding) is evaluated on all units, as decoder trained on population activity can in principle learn to down-weight uninformative units. Furthermore, all units with a firing rate less than 1 Hz are excluded from evaluation in Task Suites 2 and 3. See Table~\ref{tab:filter_comparison} for unit counts with varying filters applied.

\paragraph{Overall Neural QC (\texttt{qc\_neural}).} A probe-level neural QC label \texttt{qc\_neural} is computed as the minimum (most severe) label across \texttt{qc\_neural\_raw} and \texttt{qc\_neural\_spikesorting}. Probe alignment quality is tracked separately and used only for Task Suite 3 exclusions as described above.

\begin{table}[ht]
    \centering
    \caption{Neural Unit Attrition Across Filtering Stages}
    \label{tab:filter_comparison}
    \small 
    \begin{tabular}{l ccc r}
        \toprule
        \textbf{Filtering Stage} & {\textbf{\# Sessions}} & {\textbf{Units}} & {\textbf{Unit-hours}} & \textbf{\% Retained} \\
        \midrule
        Original Brain Wide Map        & 459 & 621,733    & {---}     & {---} \\
        Unfiltered \brainwidebench     & 452 & 610,701    & 830,324   & 100\% \\
        \quad + Probe QC               & 452 & 601,124    & 816,371   & 98.3\% \\
        \quad + Firing Rate            & 452 & 446,311    & 606,728   & 73.1\% \\
        \quad + Unit QC                & 452 & 67,160     & 90,390    & 11.0\% \\
        \bottomrule
    \end{tabular}
\end{table}
\noindent Note: Unfiltered \brainwidebench~corresponds to the same data as the Original Brain Wide Map, except for the exclusion of 7 sessions that failed QC on the evaluation animals.

\subsection{Behavioral Quality Control}
Sessions were recorded from three video cameras: two side cameras (``left'' and ``right'') that capture orthogonal views of the face and upper body of the mouse, and a ``body'' camera that captures the back of the mouse. We only utilize the ``left'' video in the benchmark, acquired at 60 frames per second. The IBL video processing pipeline~\citep{ibl2022video} produces two major outputs relevant to this benchmark: pose estimates from Lightning Pose for the left and right paws and the tongue~\citep{biderman2024lightning}, and an estimate of the motion of the whisker pad computed as the mean absolute pixel difference between successive frames in a region of interest (ROI) around the whisker pad. Three video-derived behavioral signals are subject to additional manual quality assessment, each assigned a \texttt{PASS}, \texttt{WARNING}, or \texttt{FAIL} label.

\begin{figure}[!t]
    \centering
  \includegraphics[width=\linewidth]{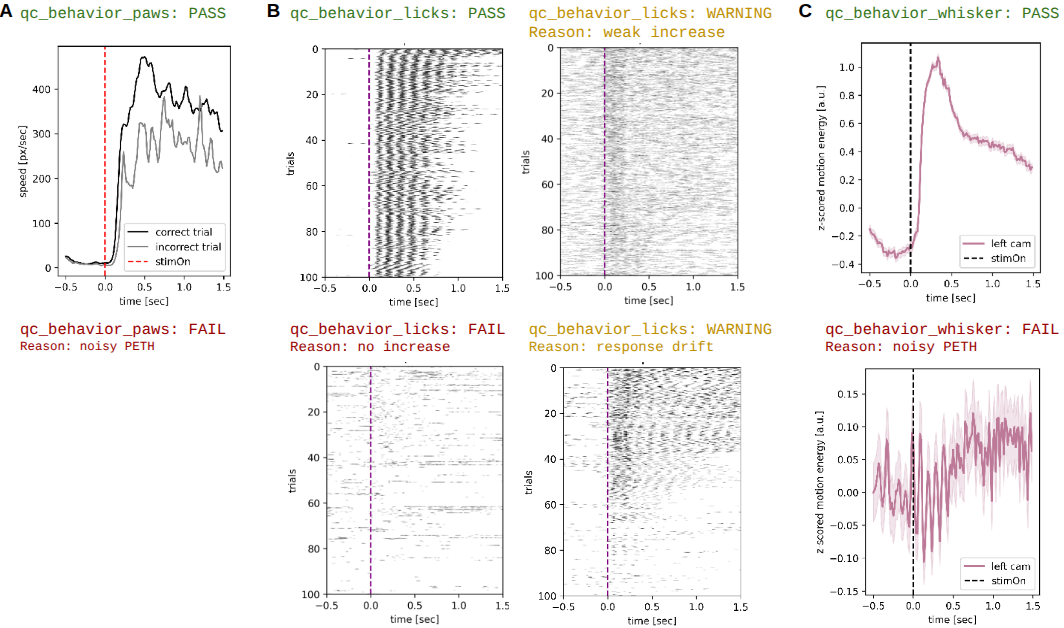}
    \caption{{\bf Behavior QC.} 
    \textbf{A}: \textit{Top}: Example session with \texttt{qc\_behavior\_paws=PASS}, showing paw trajectories averaged separately for correct and incorrect trials, aligned to stimulus onset. No sessions had \texttt{WARNING} or \texttt{FAIL} labels for paws.
    \textbf{B}: 
    \textit{Top left}: Example session with \texttt{qc\_behavior\_licks=PASS}, showing lick times for correct trials aligned to feedback onset. 
    \textit{Bottom left}: Example session with \texttt{qc\_behavior\_licks=FAIL}, where there is no discernible increase in lick rate during the reward period, likely due to poor tongue tracking. 
    \textit{Right}: Two example sessions with \texttt{qc\_behavior\_licks=WARNING}, showing weak but present lick rate increase (\textit{Top}) and non-stationary lick rate pattern over the course of the session (\textit{Bottom}).
    \textbf{C}:
    \textit{Top}: Example session with \texttt{qc\_behavior\_motion\_energy=PASS}, showing whisker pad motion energy averaged over correct trials, aligned to stimulus onset.
    \textit{Bottom}: Example session with \texttt{qc\_behavior\_motion\_energy=FAIL}, where the signal is dominated by noise. No sessions were marked \texttt{WARNING}.
    } \label{fig:behavior_qc}
\end{figure}

\paragraph{Paw Poses (\texttt{qc\_behavior\_paws}).} Paw pose estimation is assigned \texttt{FAIL} if obvious timestamp errors are present, as evidenced by a noisy peri-event time histogram (PETH), and \texttt{PASS} otherwise (Fig.~\ref{fig:behavior_qc}A).

\paragraph{Lick Rate (\texttt{qc\_behavior\_licks}).} Lick rate is computed from the tongue pose estimates~\citep{ibl2022video}. Lick rate is assigned \texttt{FAIL} if timestamp issues are apparent in other Lightning Pose traces for either camera, if no discernible increase in lick rate is observed at reward time, or if one ROI around the mouth is incorrect without a compensating strong oscillatory signal during reward consumption. \texttt{WARNING} is assigned if an increase around reward time is present but weak, if both ROIs are acceptable but the oscillatory pattern is moderate, or if the oscillatory pattern changes substantially across the session in a manner that could reflect either a real behavioral change or a tracking artifact. \texttt{PASS} requires strong oscillatory licking signal with both cameras yielding acceptable ROIs  (Fig.~\ref{fig:behavior_qc}B).

\paragraph{Whisker Pad Motion Energy (\texttt{qc\_behavior\_motion\_energy}).} Whisker pad motion energy is assigned \texttt{FAIL} if the ROI is not properly centered on the whisker pad or if timestamp errors produce a noisy PETH. \texttt{WARNING} is assigned if the stimulus-aligned PETH is present but weak. \texttt{PASS} is assigned otherwise (Fig.~\ref{fig:behavior_qc}C).

\paragraph{Overall Behavioral QC ( \texttt{qc\_behavior}).} A session-level behavioral QC label \texttt{qc\_behavior} is computed as the minimum (most severe) label across \texttt{qc\_behavior\_paws}, \texttt{qc\_behavior\_licks} and \texttt{qc\_behavior\_motion\_energy}.

\subsection{Trial-level Quality Control}
Within individual sessions, several filters were applied to remove trials with missing data or unusual behavior.
Trials were excluded if one of the
following trial events could not be detected: 
\texttt{stimOn\_times},
\texttt{probabilityLeft},
\texttt{firstMovement\_times},
\texttt{choice}, 
\texttt{feedback\_times}, 
and
\texttt{feedbackType}.
Trials were further excluded if the time between stimulus onset and the first movement of the wheel (a measure of reaction time) was longer than 10.0 s, which eliminates trials where the animal may not be actively engaged in the task.

%% file: appendix/data_processing.tex
\section{Data Processing}
\label{app:data_preprocessing}

Raw data were accessed via the International Brain Laboratory's Brainwide Map data release \citep{bwmdata2022figshare} using the ONE API under the MIT license (\url{https://docs.internationalbrainlab.org/notebooks_external/2025_data_release_brainwidemap.html}). All processed outputs are serialized using the \texttt{brainsets} format, a structured HDF5-based schema designed for neural population data that supports arbitrary temporal resolution and heterogeneous modalities within a unified session object (\url{https://brainsets.readthedocs.io}). Each session is represented as a \texttt{Data} object whose attributes expose the processed neural and behavioral streams described below; for instance, the unified neural population activity is accessible as \texttt{data.spikes}.

\paragraph{Neural Activity (\texttt{spikes}).}
Spiking activity and cluster metadata were extracted for all available Neuropixels probe insertions within a session. To account for multiple probes, probe-local cluster IDs were offset to generate globally unique unit indices across the entire session. Anatomical coordinates for each unit were mapped to the Beryl brain region atlas. Finally, spike timestamps, amplitudes, depths, and unique unit indices from all probes were concatenated and temporally sorted into a unified, irregular time series representing the session-wide population activity. Note, the irregular time series object enables the data to be stored unbinned, allowing each benchmark user to bin the neural activity to their desired bin size. The unified neural population is exposed as \texttt{data.spikes}.

\paragraph{Continuous Behavioral Processing and Alignment Methodology.}
All continuous behavioral traces were extracted, aligned to a common temporal grid, and optionally normalized (see below). Behavioral time series were aligned to a uniform 50~Hz sampling rate, accommodating missing frames or gaps using a 10~ms gap tolerance during alignment (see below).

\paragraph{Signal Resampling and Anti-Aliasing Filtering.}
To prevent aliasing artifacts during the downsampling and resampling of high-frequency continuous data (e.g., 1~kHz wheel kinematics or $>$60~Hz video tracking), an anti-aliasing low-pass filter was applied prior to rate reduction. Specifically, signals were processed using an 8th-order zero-phase filter (Butterworth filters provide better qualitative final results compared to Chebyshev) to attenuate high-frequency noise without introducing phase shifts into the behavioral time series. To maintain a conservative safety margin, the filter's cutoff frequency was strictly set to 80\% of the target Nyquist frequency. This ensured high-fidelity signal representation when decimating the data onto the final 50~Hz temporal grid.

\paragraph{Wheel Kinematics (\texttt{wheel\_speed}).} Wheel position was monitored with dedicated rotary encoder hardware.
Wheel angular velocity traces were extracted from position traces and aligned to the common timebase. To preserve high-frequency fidelity during alignment, the wheel data was first regularized to a 1~kHz sampling grid, strictly phase-aligned to the start and end timestamps of the video data. The 1~kHz trace was then sequentially decimated (downsampled) by factors of 10 and 2 to reach the target 50~Hz rate. Finally, wheel speed was derived by taking the absolute value of the downsampled velocity trace, yielding a non-negative scalar in physical (angular) units. The processed wheel signals are exposed as \texttt{data.wheel}.

\paragraph{Pose Dynamics (\texttt{left\_paw\_speed} and \texttt{right\_paw\_speed}).}
Pose tracking data, generated via Lightning Pose~\cite{biderman2024lightning}, was extracted for the left and right paws from the left camera video feed, which captures the face and upper trunk of the mouse. Because the pose dataset does not inherently contain timestamps, it inherited the timestamp array from the corresponding left camera video.

\textit{Keypoint Kinematics.} For each keypoint $k \in \{\text{left paw}, \text{right paw}\}$ with 2D position $\mathbf{p}_k(t) = (x_k(t), y_k(t))$, the instantaneous 2D velocity was computed by central-difference differentiation along the time axis, $\mathbf{v}_k(t) = d\mathbf{p}_k/dt$, and the scalar speed was obtained as its Euclidean magnitude, $s_k(t) = \lVert \mathbf{v}_k(t) \rVert_2$. The paw kinematics and their per-frame confidence flags are exposed as \texttt{data.paws}.

\textit{Per-Frame Confidence Masks.} Lightning Pose returns, for each keypoint and each frame, a continuous \emph{likelihood} score in $[0, 1]$ reflecting the network's confidence in the localization. We threshold these per-paw likelihoods at $\tau = 0.9$ to obtain a binary, per-frame reliability flag. Because the velocity (and hence speed) at frame $t$ is computed via central differences on the regularized position trace, each kinematic sample depends on its two temporal neighbors $t \pm 1$; a single low-confidence frame therefore contaminates the velocity estimates of its neighbors. To account for this, the binary likelihood mask is morphologically eroded by one sample in time, yielding the final per-paw confidence masks (\texttt{is\_left\_paw\_confident}, \texttt{is\_right\_paw\_confident}). These masks are propagated alongside the kinematic signals on the 50~Hz grid and are used downstream both to gate normalization statistics for the paw signals (see \textbf{Behavioral Normalization}) and to decode  unreliable frames in TS1 evaluation.

\paragraph{Whisker Motion Energy (\texttt{whisker\_motion\_energy}).} Whisker motion energy was extracted from the left camera video feed. The motion energy is a single scalar per frame, computed by taking the sum of the absolute value of pixel-level differences across consecutive frames~\cite{ibl2022video}. Because native video sampling rates varied across sessions ($\sim$60~Hz or $\sim$150~Hz), the time series was first regularized onto a uniform timestamp grid and then resampled to the 50~Hz target rate. 
The whisker motion energy trace is exposed as \texttt{data.whisker}.

\paragraph{Licking Behavior (\texttt{licking\_rate}).}
Licking events were recorded as discrete timestamps, precomputed on the IBL public database using pose tracking outputs on the tongue~\citep{ibl2022video}. To align this point-process data with the continuous behavioral features, lick timestamps were binned directly onto the regularized 50~Hz timestamp grid established by the video data. The resulting binned counts were multiplied by the target sampling rate to yield a continuous lick rate (in Hz). The continuous lick rate is exposed as \texttt{data.licks}.

\paragraph{Behavioral Normalization.}
Normalization is applied to the pose- and video-derived signals (\texttt{left\_paw\_speed}, \texttt{right\_paw\_speed}, and \texttt{whisker\_motion\_energy}), which are expressed in raw pixel units. Their absolute scale is camera- and session-dependent and can take very large numerical values that would otherwise dominate the loss and gradient dynamics across modalities; \textit{z-scoring} places these heterogeneously scaled signals on a common, well-conditioned numerical footing. By contrast, \texttt{wheel\_speed} and \texttt{licking\_rate} are \emph{not} normalized: \texttt{wheel\_speed} is already in physical units (rad/s) from the rotary encoder, and \texttt{licking\_rate} is scale-controlled by construction, obtained by binning lick events on the 50~Hz grid and multiplying by the sampling rate to yield a count-process rate in Hz. Both are consumed in their native units.

Per-signal \emph{z-scoring} was computed as $\tilde{x}(t) = (x(t) - \mu) / \max(\sigma, \varepsilon)$, with $\varepsilon = 10^{-8}$ guarding against zero-variance signals. To prevent leakage and to ensure that the normalization scale is representative of the data actually scored at evaluation, the statistics $(\mu, \sigma)$ are accumulated over the same task-aligned movement intervals that define the evaluation windows, restricted to the training split only. In other words, $(\mu, \sigma)$ are estimated on the train-split portion of the exact temporal windows on which the regression metrics are computed at test time, rather than on the full continuous session.
For the paw kinematics, only frames passing the per-paw confidence mask (\texttt{is\_left\_paw\_confident}, \texttt{is\_right\_paw\_confident}) contribute to $(\mu, \sigma)$. This prevents low-confidence tracking artifacts from inflating the estimated scale and distorting the standardized signal.

\paragraph{Discrete Behavioral Labels.}
Three behavioral variables are extracted as trial-level targets for classification: stimulus contrast, choice, and reward. They are taken directly from the dataset, kept unchanged, and exposed under \texttt{data.task\_aligned\_intervals} as \texttt{.stimulus\_contrast}, \texttt{.choice}, and \texttt{.reward}.

\paragraph{Dataset Builds.}
The dataset is released in two builds that differ only in the unit population they
store; the sessions, the behavioral streams, and the temporal splits are identical
across both. \texttt{all\_units} applies no unit filtering. \texttt{selected\_units}
applies three filters before the spike data is written to disk: \texttt{firing\_rate}
(mean firing rate above 1.0~Hz), \texttt{unit\_qc} (IBL unit QC label equal to 1.0,
i.e.\ well-isolated units), and \texttt{probe\_qc} (probe insertions carrying a passing
QC label, applied to evaluation sessions only). Because the filters run at processing
time, they are baked into the HDF5 files and cannot be changed without reprocessing.
Which build a suite uses is dictated by what it predicts. TS2 and TS3 attach their
targets to individual units: TS2 reconstructs a held-out neuron's activity and TS3
classifies a unit's brain region, so a poorly isolated unit is a corrupted target or a
corrupted label rather than merely a noisy input, and both suites require
\texttt{selected\_units}. TS1 decodes behavior from the whole population, where each
unit is only an input feature and filtering discards usable signal; all reported TS1
baselines therefore use \texttt{all\_units}. Pretraining imposes no requirement and may
use either build. Table~\ref{tab:build_overview} reports the composition of each build.

\begin{table}[t]
\centering
\caption{Composition of the two dataset builds. Sessions, subjects, and recording
duration are identical across builds; only the stored unit population differs. Probes
counts insertions that still contribute at least one unit. Duration is total session
recording time.}
\label{tab:build_overview}
\small
\begin{tabular}{llrrrrrr}
\toprule
Build & Regime & Sessions & Subjects & Probes & Units & Spikes & Duration (h) \\
\midrule
\texttt{all\_units}      & pretrain & 423 & 126 & 643 & 567{,}514 & 19.21\,B & 561.5 \\
\texttt{all\_units}      & eval     &  29 &  13 &  45 &  43{,}187 &  1.47\,B &  40.7 \\
\midrule
\texttt{selected\_units} & pretrain & 423 & 126 & 642 &  63{,}382 &  3.75\,B & 561.5 \\
\texttt{selected\_units} & eval     &  29 &  13 &  37 &   3{,}778 &  0.21\,B &  40.7 \\
\bottomrule
\end{tabular}
\end{table}

%% file: appendix/data_splits.tex
\section{Data Splits}
\label{app:data_splits}

\begin{figure}[!t]
    \centering
  \includegraphics[width=0.9\linewidth]{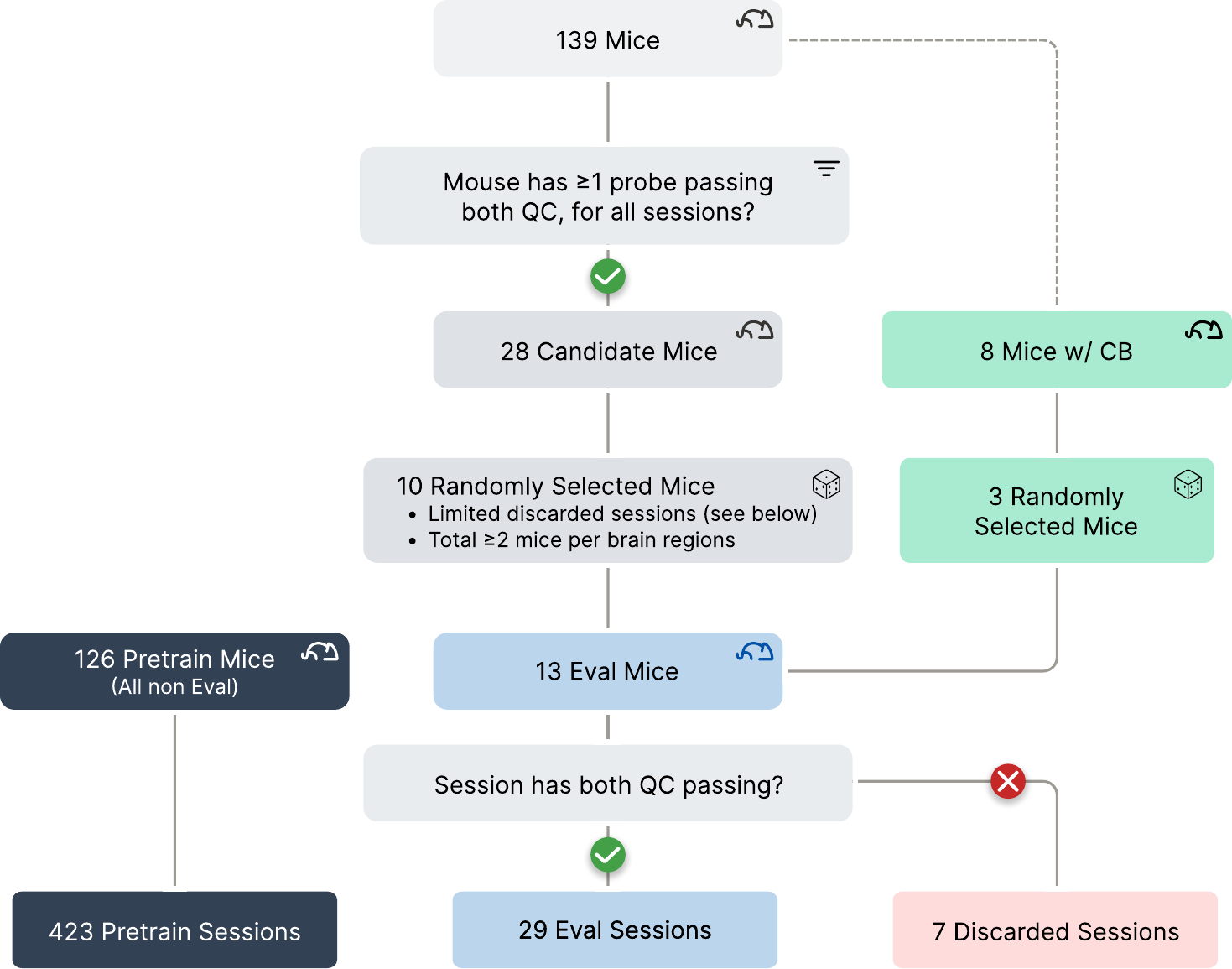}
    \caption{\textbf{Pretrain/Evaluation Mice Split.} Selection pipeline for the 13 evaluation mice from the full cohort of 139 mice. Starting from all 139 mice, 28 candidates are identified for whom every session has at least one probe passing both neural and alignment quality control and full behavioral quality control. From this pool, 10 mice are selected via a seeded greedy search optimizing for broad brain region coverage and minimal within-cohort session discards. Separately, 3 mice with confirmed cerebellum coverage are drawn from a curated pool of 8 cerebellum candidates, yielding 13 evaluation mice in total. All remaining 126 mice are assigned to pretraining. Within the evaluation cohort, 7 individual sessions failing quality control are discarded, resulting in 29 evaluation sessions and 423 pretraining sessions.
    } \label{fig:mice_split}
\end{figure}

\subsection{Pretrain and Evaluation Split}

The overarching design principle of the \brainwidebench~pretraining/evaluation split is strict animal-level separation: no animal contributes data to both the pretraining and evaluation sets. 
This ensures that all three task suites evaluate genuine cross-animal generalization rather than within-animal memorization.

Two main criteria were considered to ensure effective transfer. First, the neural and behavioral data in the evaluation set must meet quality standards (passing the quality control considerations detailed in Appendix~\ref{app:data_qc}). Second, the selected animals must provide broad coverage of brain regions, particularly for TS3 (Neuron Identity Prediction), where representative sampling across areas is essential for a meaningful evaluation of generalization. The overall selection process is illustrated in Figure~\ref{fig:mice_split} and described in detail in the following paragraphs.

Following the paper submission, we intend to host a submission-based leaderboard to accelerate community adoption of \brainwidebench~and to maintain a living record of the state of the art. We note that \brainwidebench~does not rely on a separate held-out evaluation set. Instead, all pretraining and evaluation sessions are openly released. This decision is deliberate. The benchmark is built on the Brainwide Map dataset, which is already publicly available, making any withheld sessions or signals straightforward to identify. Furthermore, the task suites impose heterogeneous requirements on the data: for example, withholding behavioral signals to prevent leakage in TS1 (Neural Decoding) would preclude their use as supervision for constructing neural embeddings in TS3 (Neuron Identity Prediction). We therefore release all data in full and trust the community to engage with the benchmark in the spirit in which it is intended.

\paragraph{Identifying the evaluation candidate pool with Quality Control.}
We first identify 28 candidate mice by applying a two-stage quality filter. A session is considered usable if it passes behavioral quality control (\texttt{qc\_behavior = PASS}) and contains at least one probe passing both neural quality control (\texttt{qc\_neural = PASS}) and neural alignment quality control (\texttt{qc\_neural\_alignment = PASS}). We retain only mice for whom \textit{every} recording session satisfies these criteria, yielding 28 animal-level candidates. This guarantee ensures that no general evaluation candidate contributes unusable sessions to the evaluation set. All remaining mice are assigned to the pretraining set (except for a cohort of mice with cerebellum recordings; see below).

\paragraph{Selecting evaluation mice via greedy search.}
From the 28 candidates, we apply a seeded greedy search to select 10 evaluation mice, jointly optimizing two criteria: (1) each selected mouse must contribute probe recordings spanning at least two distinct Cosmos-level brain regions, ensuring meaningful anatomical coverage per held-out subject; and (2) the number of probes discarded due to failing quality control within the evaluation cohort is minimized. Because cerebellum recordings are rare and scientifically valuable, 3 additional mice are sampled separately from a curated list of 8 cerebellum candidates. These cerebellum mice are not required to satisfy the animal-level criterion above; instead, sessions are filtered post-hoc, retaining only those that pass behavioral quality control and contain at least one probe passing both neural quality control and alignment quality control. This post-hoc filtering accounts for the 7 discarded sessions that differentiate the \brainwidebench's 452 sessions from the original Brainwide Map's 459 sessions. The resulting evaluation set comprises 13 held-out mice and 29 sessions, with all remaining 126 mice forming the pretraining set of 423 sessions.

\begin{figure}[!t]
    \centering
  \includegraphics[width=\linewidth]{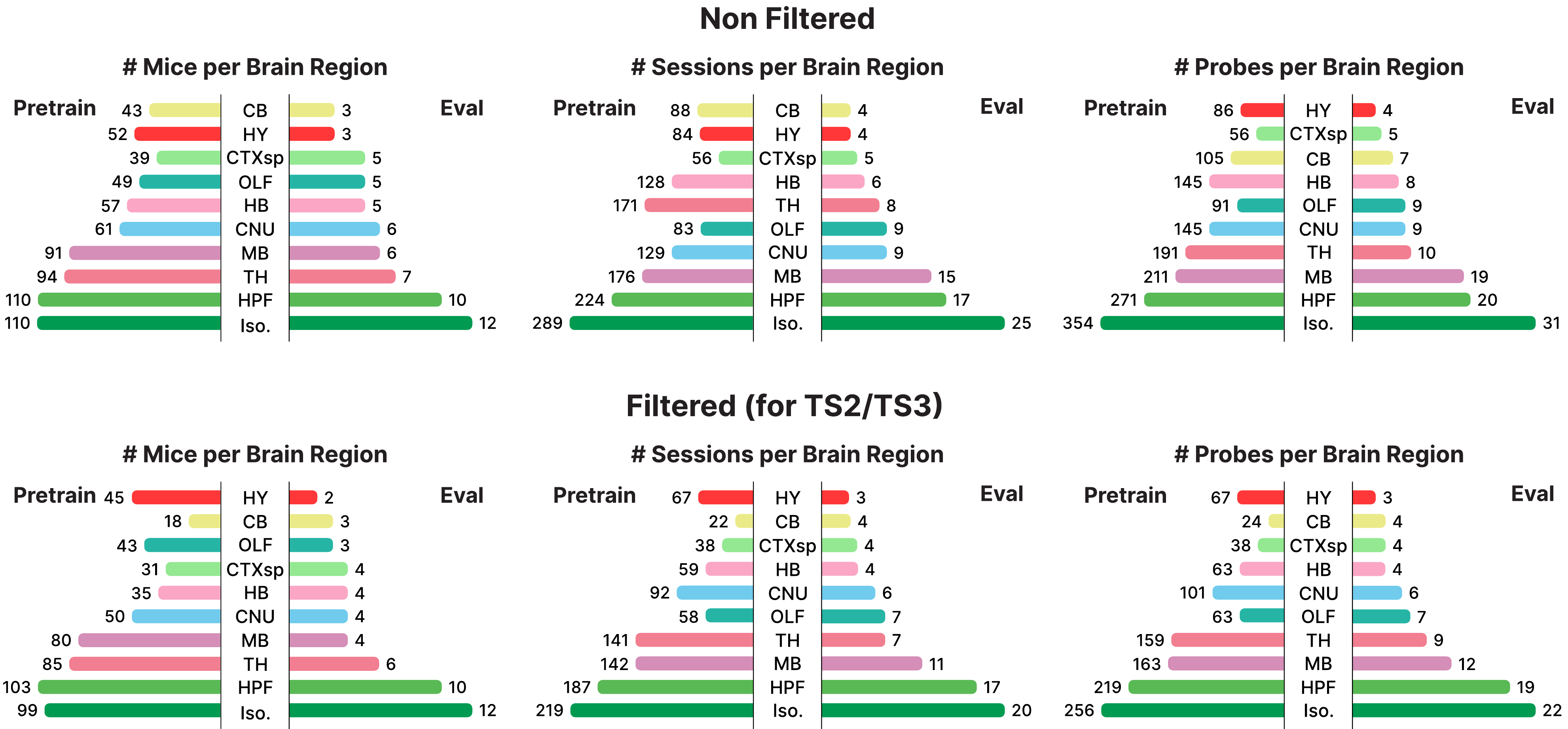}
    \caption{{\bf Brain Region Distributions between Pretrain and Evaluation Sets.}
    Bar plots show the number of mice, sessions, and probes per Cosmos-level brain region, split by pretrain versus evaluation set. The top row shows the \texttt{all\_units} build used for Task Suite 1, while the bottom row shows the \texttt{selected\_units} build used for Task Suites 2 and 3.
    } \label{fig:brain_region_dist}
\end{figure}

\begin{figure}[!t]
    \centering
  \includegraphics[width=\linewidth]{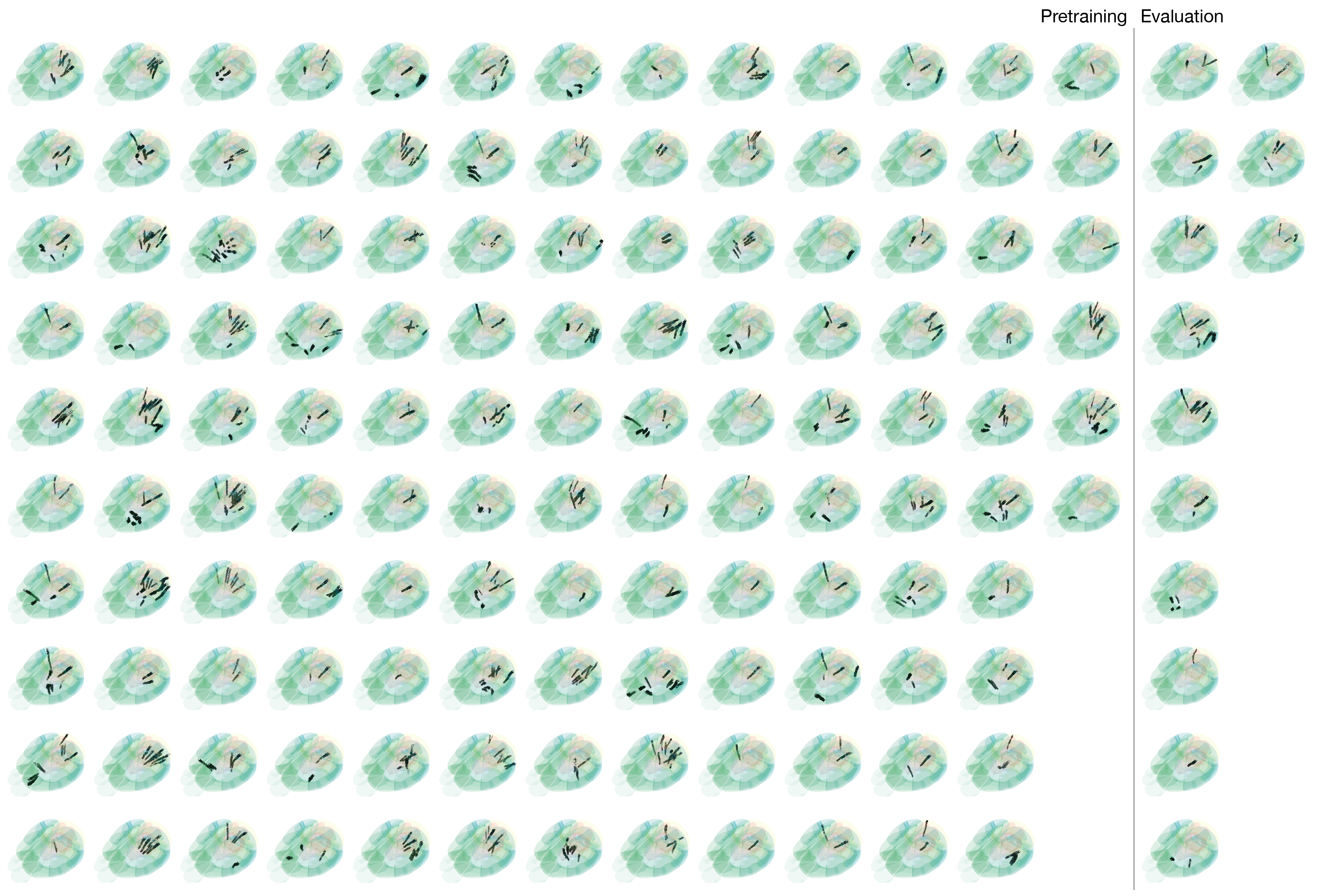}
    \caption{{\bf Probe locations across 139 subjects in \brainwidebench.}
     Each brain shows the histology-aligned location of neurons in one mouse subject’s brain across all acquisitions performed with that subject. Number of probe insertions ranges from 1 to 16 across subjects. Probe locations include added jitter (mean 0 s.d. 300 $\mu$m Gaussian) for better visualization.
    } \label{fig:brain_region_atlases}
\end{figure}

The resulting pretrain/evaluation split yields a brain region coverage distribution that is shown in Figure~\ref{fig:brain_region_dist}. Despite the strict animal-level separation and quality-driven selection, the evaluation set maintains broad anatomical coverage across all major brain regions with the number of mice, sessions, and probes per region remaining proportionally consistent between the two sets, in both the \texttt{all\_units} and \texttt{selected\_units} builds (Appendix~\ref{app:data_preprocessing}). Figure~\ref{fig:brain_region_atlases} illustrates the spatial distribution of individual neurons across the pretrain and evaluation mice.

\paragraph{Pretraining Scale.}
Table~\ref{tab:pretrain_scale} reports the size of the pretraining corpus under three
tokenization schemes: spike tokens, used by event-based models such as POYO; unit-bin
tokens, used by patch-based models such as NDT2; and time-step tokens, used by models
that keep units in the channel dimension, such as NDT-Stitch. Counts cover the full spike
domain of every pretraining session, with no
restriction to task-aligned intervals, and assume 1\,s windows and 20\,ms bins. They are
given for both builds; time-step
tokens are near-identical across builds because the unit dimension never enters the token axis.

\begin{table}[h]
\centering
\caption{Pretraining corpus size over the full spike domain of all 423 sessions. Spike
tokens count one token per spike; unit-bin tokens count one per (unit, 20\,ms bin);
time-step tokens count one per 20\,ms bin with units in the channel dimension. Binned
counts tile the domain with non-overlapping 1\,s windows at 20\,ms resolution.}
\label{tab:pretrain_scale}
\small
\begin{tabular}{lrrrrrr}
\toprule
& & & & \multicolumn{3}{c}{Tokens} \\
\cmidrule(lr){5-7}
Build & Sessions & Units & Duration (h) & Spike & Unit-bin & Time-step \\
\midrule
\texttt{all\_units}      & 423 & 567{,}514 & 561.5 & 19.21\,B & 138.42\,B & 101{,}082{,}400 \\
\texttt{selected\_units} & 423 &  63{,}382 & 561.5 &  3.75\,B &  15.36\,B & 101{,}080{,}400 \\
\bottomrule
\end{tabular}
\end{table}

\newpage

\subsection{Within-session Splits}
\label{app:within_session_splits}
\begin{figure}[th]
\centering
\includegraphics[width=\linewidth]{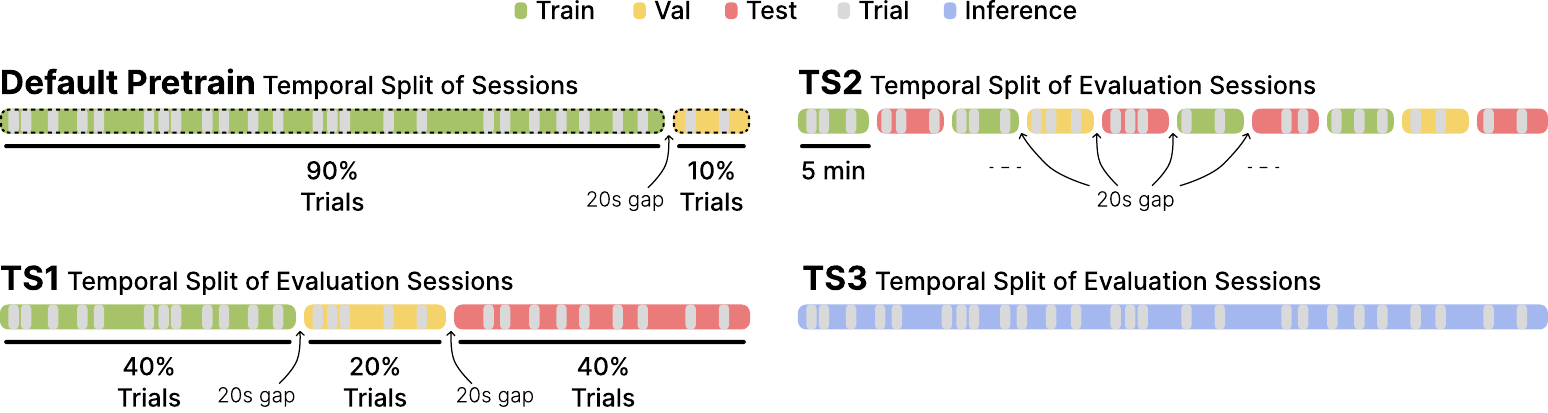}
\caption{\textbf{Train/Validation/Test Splits Across Pretraining and Evaluation Sessions.} Each bar represents a single session, with colored segments indicating the temporal assignment of trials to training (green), validation (yellow), test (red) and inference (blue) splits. Splits are separated by 20-second gaps to prevent leakage and to accommodate longer context windows. Trial windows are shown in gray. Pretraining sessions use a 90/10 train/val split with no test set. Each evaluation task suite uses a split configuration tailored to its objectives: TS1 uses a causal 40/20/40 train/val/test split; TS2 uses interleaved 5-minute blocks; and TS3 uses the full session for inference.}
\label{fig:temporal_splits}
\end{figure}

Following the pretraining/evaluation split at the subject level, each session is further partitioned into training, validation, and test subsets at the trial level and specifically for each task suite as shown in Figure~\ref{fig:temporal_splits}.

\paragraph{Pretraining sessions.}
For pretraining sessions, the first 90\% of trials are assigned to the training split and the remaining 10\% to the validation split. No test split is defined for pretraining sessions. This split is applied by default and can be updated to accommodate more advanced strategies, such as curriculum learning.

\paragraph{Evaluation sessions.}
For evaluation sessions, trials are divided into training (40\%), validation (20\%), and test (40\%) splits.
The relatively small training fraction favors models capable of learning from limited data, while the large test fraction maximizes the number of trials available for evaluation.

\paragraph{Task Suite 1.} This setting utilizes a causal (non-interleaved) split: trials are assigned to splits in temporal order, such that all training trials precede all validation trials, which in turn precede all test trials. This design prevents decoding models from exploiting long-term signal drift, rather than learning the underlying structure of the data~\citep{harris2020nonsense}. Placing the validation split between training and test maximizes the temporal buffer between the two, further limiting the ability of models to exploit session-level drift. To enable longer context windows extending beyond 2 seconds (see Figure~\ref{fig:context_window}), a 20-second gap is enforced between adjacent splits.

\begin{figure}[ht]
    \centering
    \includegraphics[width=\linewidth]{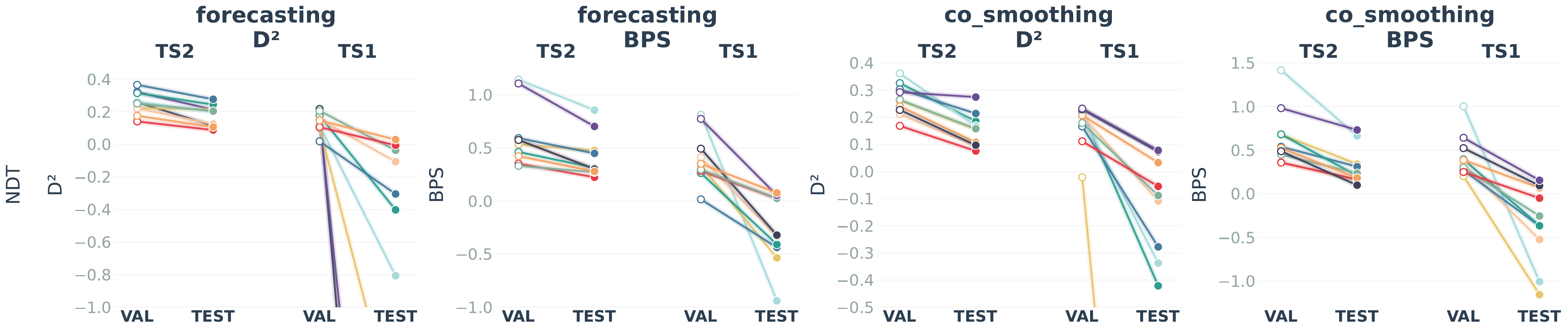}
    \caption{\textbf{TS2 Performance Under Causal (TS1) versus Interleaved (TS2) Split Strategies.} Each panel shows validation and test performance for a single-session NDT-Stitch model finetuned on 10 randomly selected evaluation sessions, across four task and metric combinations: forecasting ($D^2$), forecasting (BPS), co-smoothing ($D^2$), and co-smoothing (BPS). Each line connects the validation and test performance of a single session, with the left half of each panel showing the interleaved (TS2) split and the right half showing the causal (TS1) split. Under the interleaved split, validation and test performance are broadly consistent, reflecting the stationarity of the estimation problem. Under the causal split, performance drops sharply from validation to test across nearly all sessions and metrics, illustrating the detrimental effect of neural non-stationarity over long timescales. The slight reduction in validation performance between the TS2 and TS1 regimes is attributable to hyperparameters having been tuned for the interleaved split setting.}
    \label{fig:ts_analysis_splits}
\end{figure}

\paragraph{Task Suite 2.} This setting utilizes an interleaved split of 5-minute temporal blocks following the pattern {\splitbox{trainColor}{train}\splitbox{testColor}{test}\splitbox{trainColor}{train}\splitbox{valColor}{val}\splitbox{testColor}{test}}, with a 20-second gap between blocks. 
This block structure reduces autocorrelation leakage that would arise from fully randomized trial-level splits while ensuring that each subset is large enough to contain representative samples of neural dynamics. Due to the strong non-stationary drift of neural recordings~\citep{ibl2022spikesorting}, we found that replicating the causal split of TS1 led to a sharp drop in performance between validation and test sets (see Figure~\ref{fig:ts_analysis_splits}). As TS2 probes dynamical structure of neural populations, models trained on this task are more susceptible to non-stationary dynamics.

To motivate this choice, we compare TS2 task performance under two split strategies: causal (TS1) and interleaved (TS2) (Figure~\ref{fig:ts_analysis_splits}). 
Under the causal split, models are trained on the earlier portions of a recording and evaluated on the final held-out portion (Figure~\ref{fig:temporal_splits}). We use the best-performing TS2 model (finetuned NDT-Stitch) across both tasks (forecasting and co-smoothing) and both metrics ($D^2$ and bps). We observe performance drops from validation to test under this strategy. 
This likely reflects non-stationarity in neural recordings~\citep{ibl2022spikesorting}: as the temporal gap between train and test grows, probe drift and spike sorting artifacts cause neural dynamics to shift, making prediction of a fixed set of neurons increasingly ill-defined. Interleaved splits mitigate this by mixing train, validation, and test windows throughout the session, yielding more stable performance across splits.

\begin{figure}[ht]
    \centering
    \includegraphics[width=\linewidth]{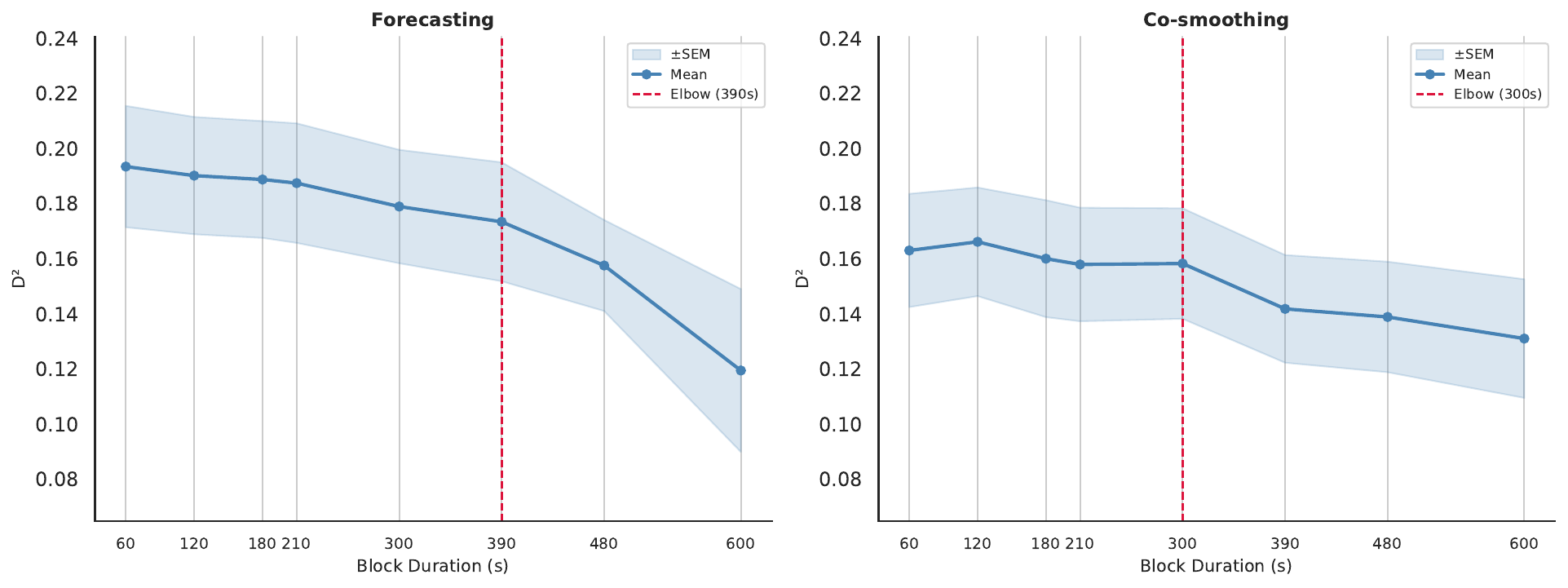}
    \caption{\textbf{Effect of Block Duration on Test $D^{2}$ Score.} Mean test $D^{2}$ ($\pm$ SEM) across 10 fine-tuned NDT models as a function of block duration, shown for the Forecasting (left) and Co-smoothing (right) tasks. Block duration is the length, in seconds, of the contiguous recording segments used to define the non-causal train/val/test splits for TS2. The dashed line marks the geometric elbow, defined as the block duration whose mean $D^{2}$ is at maximum perpendicular distance from the line connecting the first and last points.}
    \label{fig:block_duration_exp}
\end{figure}

We additionally examine how the choice of block duration affects test performance (Figure~\ref{fig:block_duration_exp}). Performance varies monotonically with block size: longer blocks reduce autocorrelation leakage but increase the difficulty of generalization across splits, while shorter blocks have the opposite effect. We select a block duration of 5 minutes (300 s) as a compromise between the two tasks: the geometric elbow occurs at 300 s for co-smoothing and 390 s for forecasting, and performance decreases only marginally between these two values.

\paragraph{Task Suite 3.} This setting involves classifying the brain region of each recorded unit in a zero-shot setting and, by nature, does not depend on any particular temporal splitting. In the inductive setting, a model may use all available session data at test time (inference only, no gradient updates) to predict unit labels. For transductive models that first calibrate on a downstream task (e.g., behavior prediction in TS1 or neural activity prediction in TS2), the standard convention applies: the training split is used for gradient updates and the validation split for early stopping. The result is a set of embeddings for the new neural population that was learned on the evaluation data using a training objective external to the brain region task. Note that in our baseline evaluation, models operating in the transductive setting have access to less session data for embedding generation, since only the training split contributed to updating the unit embeddings and validation was used to select a set of embeddings. Nonetheless, nothing prevents such models from calibrating on the full session prior to transfer, but doing so requires an additional run.

%% file: appendix/task_suites.tex
\section{Task Suites}
\label{app:task_suites}

In this section, we provide details on the three task suites incorporated into our benchmark.

\input{appendix/task_suites/ts1}

\input{appendix/task_suites/ts2}

\input{appendix/task_suites/ts3}

%% file: appendix/task_suites/ts1.tex

\subsection{TS1 Details: Behavioral Decoding}
\label{app:ts1}

\begin{figure}[t]
    \centering
    \includegraphics[width=\linewidth]{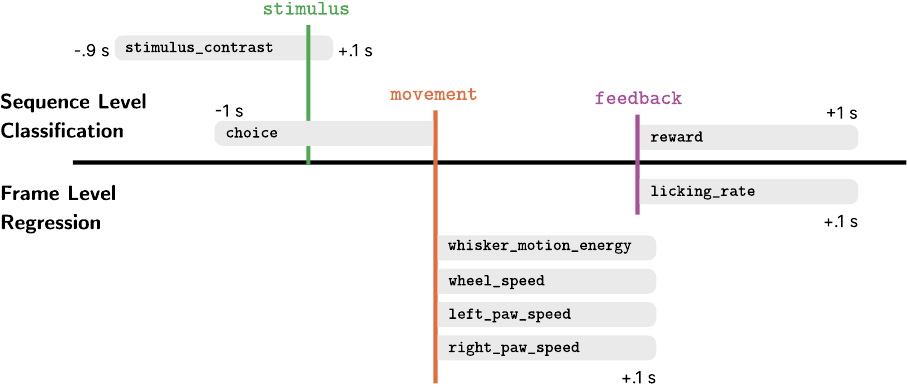}
    \caption{\textbf{Task-Aligned Decoding Windows for Behavior Prediction (TS1).} Illustration of the time windows used for each behavioral decoding target, aligned to three key trial events: stimulus onset (green), movement onset (orange), and feedback onset (purple). Sequence-level classification targets (top) include \texttt{stimulus\_contrast}, decoded from a window around stimulus onset, and \texttt{choice} and \texttt{reward}, decoded from windows ending at and starting from movement onset and feedback onset, respectively. Frame-level regression targets (bottom) include \texttt{whisker\_motion\_energy}, \texttt{wheel\_speed}, \texttt{left\_paw\_speed}, and \texttt{right\_paw\_speed}, decoded from a 1 second window starting at movement onset, and \texttt{licking\_rate}, decoded from a 1 second window starting at feedback onset. Windows are designed to avoid leakage between task phases while capturing the most task-relevant period of neural and behavioral activity for each target.}
    \label{fig:task_intervals_ts1}
\end{figure}

\paragraph{Task Structure Overview.} We define 8 decoding tasks in total: 5 at the frame level and 3 at the sequence level (see Table~\ref{tab:ts1_details}). To minimize confounds and maximize the ability to decode intent, we standardize all target windows to a fixed duration of 1 second. An overview of these windows is provided in Figure~\ref{fig:task_intervals_ts1}.

\input{tables/table_task_infos}

\paragraph{Frame-Level Tasks.} 

For the movement interval, we use a 1 second window of behavioral signals (\texttt{wheel\_speed}, \texttt{left\_paw\_speed}, \texttt{right\_paw\_speed}, and \texttt{whisker\_motion\_energy}) starting from movement onset (defined as the beginning time of the first sustained change in wheel position greater than 0.1 radians after the quiescent period). Decoding these movement-oriented signals in a window aligned to first movement onset ensures robust variation and task-relevant changes in the target signals. Note that right paw speed may provide an easier target than left paw due to the orientation of the camera from which the data is derived, consistent with results in Table~\ref{tab:ts1_main}.

For licking rate, which occurs after the feedback phase (water is delivered via solenoid upon reward, or a noise burst is played over a speaker upon non-reward), we define a 1 second window starting at feedback onset.

\paragraph{Sequence-Level Tasks.} For stimulus contrast, we decode the contrast level of the visual stimulus presented to the mouse, which can take one of five values: 100\%, 25\%, 12.5\%, 6.25\%, and 0\%. We use a 100\,ms window starting at stimulus onset (defined as the appearance of the Gabor patch on the screen), consistent with the decoding window used in~\cite{international2025brain}. This window is chosen carefully: a longer window risks leakage from the movement or reward phase, which could be spuriously correlated with higher stimulus contrast, since the mouse is more likely to move and be rewarded when the stimulus is more visible. At the same time, 100\,ms provides sufficient neural signal to decode a response without overlapping substantially with reward and decision-making activity. We also hypothesize that reaction time may itself be correlated with stimulus contrast; using a fixed window rather than one that varies with reaction time therefore reduces the risk of spurious correlations. Because we use a 1 second window overall, the stimulus contrast window also includes 900\,ms prior to stimulus onset, a period during which the screen is blank and no task-relevant information is present. We do not consider this a problem: this pre-stimulus baseline may aid decoding by providing a reference level of neural activity, and we expect models to be capable of selecting the relevant portion of the input even when additional context is provided.

For choice, we use a 1 second window ending at movement onset, ensuring that no information from the movement phase leaks into the decoding window. In~\cite{international2025brain}, this window was only 100\,ms; we extend it to 1 second for standardization and apply the same reasoning as above: models are expected to be robust and to attend selectively to the task-relevant portion of the window.

For reward, we use a 1 second window starting at feedback onset, matching the licking rate window.

For sequence-level tasks we report the class distribution across all splits except the evaluation test set in Table~\ref{tab:class_distribution}.

\begin{table}[h]
\centering
\caption{Class distribution of sequence-level tasks per split.}
\label{tab:class_distribution}
\resizebox{0.99\columnwidth}{!}{
\begin{tabular}{lccccccccc}
\toprule
\textbf{Split} & \multicolumn{2}{c}{\textbf{Choice}} & \multicolumn{5}{c}{\textbf{Contrast}} & \multicolumn{2}{c}{\textbf{Reward}} \\
\cmidrule(lr){2-3} \cmidrule(lr){4-8} \cmidrule(lr){9-10}
 & \shortstack{Right\\ ($c=0$)} & \shortstack{Left \\ ($c=1$)}& \shortstack{0\% \\ ($c=0$)} & \shortstack{6\% \\ ($c=1$)} & \shortstack{12.5\% \\ ($c=2$)} & \shortstack{25\% \\ ($c=3$)} & \shortstack{100\% \\ ($c=4$)} & \shortstack{False \\ ($c=0$)} & \shortstack{True \\ ($c=1$)} \\
\midrule
Pretrain & 0.497 & 0.503 & 0.116 & 0.219 & 0.219 & 0.221 & 0.225 & 0.168 & 0.832 \\
Eval Train & 0.499 & 0.501 & 0.114 & 0.217 & 0.219 & 0.225 & 0.224 & 0.158 & 0.842 \\
Eval Val & 0.462 & 0.538 & 0.111 & 0.233 & 0.228 & 0.205 & 0.223 & 0.147 & 0.853 \\
\bottomrule
\end{tabular}}
\end{table}

Several observations are worth noting. First, reward is a binary classification task and its distribution is heavily imbalanced: the mouse receives a reward in over 82\% of trials across all splits. Second, choice is also binary and its distribution is approximately balanced ($\sim$50\% per class), which reflects the trial structure: trials alternate between blocks (known as block priors) in which the stimulus appears on one side 80\% of the time and on the other 20\% of the time. Since these blocks are balanced across the session, the overall marginal distribution is near-uniform.

For stimulus contrast, the distribution across the 5 classes is approximately uniform across the four non-zero contrast levels ($\sim$22\% each), but the 0\% contrast class is underrepresented at 11\%. This is due to our trial exclusion criteria, which remove trials with reaction times that are too long or with no wheel movement; mice also tend to respond less when no visual stimulus is present. We note that the 0\% contrast condition is closely entangled with the block prior structure: in this condition, the mouse has no visual guidance and is expected to rely on its internalized block prior to decide which way to turn the wheel. This introduces a level of complexity that is out of scope for this benchmark and has been studied in depth in~\cite{findling2025brain}.

\emph{A Note on Window Overlap.} Some windows can overlap (for example, the stimulus contrast and choice windows), but we do not consider excluding them as our goal is to maximize the number of valid trials while ensuring that all tasks are included within each trial. In extreme cases, when the mouse's reaction time is shorter than 100\,ms (which occurs in 18\% of trials of train and val split from the evaluation sessions), the choice window precedes the stimulus contrast window entirely.

\emph{Various Difficulty of Behavioral Targets.}
Some behavioral variables, such as reward or coarse movement signals, are relatively accessible, while others like licking rate remain substantially more difficult. This range reveals both where simple models are already sufficient and where large-scale pretraining provides meaningful gains. 

\paragraph{Evaluation Paradigm.}
While pretraining can leverage both frame-level and sequence-level behavioral signals in a flexible setting (single- or multi-task, with optional modalities such as anatomical labels), we impose stricter constraints during downstream evaluation. All evaluation is performed on a single session and a single task at a time, without cross-session calibration. This reflects our target use case: a single pretrained model that generalizes to new animals and sessions unseen during pretraining. For the baselines we implemented are all calibrated using the train and validation splits of the evaluation session via fine-tuning. Note that our splits support further research in modified evaluation regimes, such as few-shot or zero-shot adaptation, through using a restricted set of the training split. All reported numbers correspond to single-task inference on the held-out test split of each session.

\paragraph{What Is Enforced, Recommended, and Left Open.}
We distinguish between three levels of constraint in the evaluation protocol.

\textit{Enforced.} The test set is always the last 40\% of trials in the session, reflecting the temporal structure of the data and ensuring consistency across all reported results. The decoding windows and task definitions described above are fixed and must not be modified. In TS1, these windows are sampled directly from the trial-aligned task intervals using the \texttt{TrialSampler} from \texttt{torch\_brain}.

\textit{Recommended.} The train and validation splits cover the first 40\% and middle 20\% of trials, respectively. When calibration is performed, we recommend fine-tuning with task-specific losses: cross-entropy for classification tasks, mean squared error (MSE) for continuous behavioral signals, and negative log-likelihood (NLL) under a Poisson distribution for licking rate. Regarding unit selection, we conducted a preliminary analysis to test the effect of unit filtering on decoding. Our initial results showed that, generally, the less filtering the better. We think this makes sense, since (1) decoders can learn to ignore units with poor signal, and (2) several of the QC filtering criteria are related to spike sorting, but even if a unit is not sorted correctly a decodeable signal may still be present. While we did not conduct a more extensive analysis with other baselines, we've generally observed this trend to be consistent. Hence, for all of the baselines we present in Table~\ref{tab:ts1_main}, we employ no unit filtering. That said, QC filtering remains a valid alternative, and benchmark users are free to experiment with filtering options that work best for their models.

\textit{Left open.} The context window size and QC filtering are left open for the community to explore, during pretraining and evaluation. During evaluation, while the target window is fixed, the context window can be expanded from the default $1$s up to $20$s. The train and validation splits may also be reduced or removed entirely to probe few-shot or zero-shot generalization, provided the test split remains unchanged.

\paragraph{Split Statistics.}
Table~\ref{tab:ts1_splits} reports the size of each split under three tokenization
schemes: spike tokens, used by event-based models such as POYO; unit-bin tokens, used by
patch-based models such as NDT2; and time-step tokens, used by models that keep units in
the channel dimension, such as NDT-Stitch. Token counts assume the default 1\,s context
window and 20\,ms bins; expanding the context window as permitted above scales them
proportionally. Spike tokens are given as a range because each task aligns its 1\,s
window to a different event, so the same number of windows contains different amounts of
neural activity.

\begin{table}[h]
\centering
\caption{TS1 split statistics on the \texttt{all\_units} build. Samples are 1\,s
task-aligned windows, disjoint within a task. Spike tokens count one token per spike;
unit-bin tokens count one per (unit, 20\,ms bin); time-step tokens count one per 20\,ms
bin with units in the channel dimension.}
\label{tab:ts1_splits}
\small
\begin{tabular}{lrrrrrr}
\toprule
& & & & \multicolumn{3}{c}{Tokens} \\
\cmidrule(lr){5-7}
Split & Units & Samples & Duration (h) & Spike & Unit-bin & Time-step \\
\midrule
train & 43{,}187 & 7{,}795 & 2.17 & 66.3 -- 81.8\,M & 583.9\,M & 389{,}750 \\
val   & 43{,}187 & 3{,}471 & 0.96 & 31.2 -- 37.9\,M & 254.6\,M & 173{,}550 \\
test  & 43{,}187 & 5{,}894 & 1.64 & 55.7 -- 67.1\,M & 424.8\,M & 294{,}700 \\
\bottomrule
\end{tabular}
\end{table}

\paragraph{Baselines.}
For our single-session baselines, we fix the input to neural activity only (no anatomical labels or auxiliary signals), using the unit selection from each model's original configuration. We use the recommended train and validation splits and evaluate in the single-task inductive setting unless otherwise specified. We apply no unit filtering in TS1. All baselines in Table~\ref{tab:ts1_main} are run on $5$ seeds.

\paragraph{Metrics.}
We report task-specific metrics for each decoding target. For frame-level regression tasks (\texttt{whisker\_motion\_energy}, \texttt{wheel\_speed}, \texttt{left\_paw\_speed}, and \texttt{right\_paw\_speed}), we report the coefficient of determination ($R^2$) and Pearson's $r$. For \texttt{licking\_rate}, which follows a count distribution, we report the pseudo-$R^2$ measure $D^2$. For all sequence-level classification tasks (\texttt{stimulus\_contrast}, \texttt{choice}, and \texttt{reward}), we report balanced accuracy and $F1$ score, accounting for the class imbalances noted in Table~\ref{tab:class_distribution}.

%% file: tables/table_task_infos.tex
\begin{table}[h]
\centering
\caption{Summary of decoding tasks with associated intervals, class counts, and evaluation metrics for TS1 (Behavior Prediction). Note: $D^2$ is defined in Appendix~\ref{app:metrics}.}
\label{tab:ts1_details}
\resizebox{0.99\columnwidth}{!}{
\begin{tabular}{clcccc}
\toprule
\textbf{Level} & \textbf{Variable Name} & \textbf{Interval Start} & \textbf{Interval End} & \textbf{Num Classes} & \textbf{Metric} \\

\midrule

\multirow{5}{*}{\rotatebox[origin=c]{90}{\textbf{Frame}}} 
& \texttt{whisker\_motion\_energy} & $ \textcolor{movementOnset}{\texttt{movement\_onset\_time}} $   & $ \textcolor{movementOnset}{\texttt{movement\_onset\_time}} + 1.0 $ & --            & $\boldsymbol{R^2}$, Pearson's $r$ \\

& \texttt{wheel\_speed}         & $ \textcolor{movementOnset}{\texttt{movement\_onset\_time}} $   & $ \textcolor{movementOnset}{\texttt{movement\_onset\_time}} + 1.0 $ & --            & $\boldsymbol{R^2}$, Pearson's $r$ \\
& \texttt{left\_paw\_speed}    & $ \textcolor{movementOnset}{\texttt{movement\_onset\_time}} $   & $ \textcolor{movementOnset}{\texttt{movement\_onset\_time}} + 1.0 $ & --            & $\boldsymbol{R^2}$, Pearson's $r$ \\

& \texttt{right\_paw\_speed}    & $ \textcolor{movementOnset}{\texttt{movement\_onset\_time}} $   & $ \textcolor{movementOnset}{\texttt{movement\_onset\_time}} + 1.0 $ & --            & $\boldsymbol{R^2}$, Pearson's $r$ \\

& \texttt{licking\_rate}        & $ \textcolor{feedbackOnset}{\texttt{feedback\_time}} $   & $ \textcolor{feedbackOnset}{\texttt{feedback\_time}} + 1.0 $ & --            & $\boldsymbol{D^2}$ \\

\midrule

\multirow{3}{*}{\rotatebox[origin=c]{90}{\textbf{Seq.}}} 
& \texttt{stimulus\_contrast} & $ \textcolor{StimulusOnset}{\texttt{stim\_on\_time}} - 0.9 $ & $ \textcolor{StimulusOnset}{\texttt{stim\_on\_time}} + 0.1 $ & 5 & \textbf{Balanced Acc}, $F1$ Score, AP \\ 

& \texttt{choice} & $ \textcolor{movementOnset}{\texttt{movement\_onset\_time}} $ & $\textcolor{movementOnset}{\texttt{movement\_onset\_time}} + 1.0 $ & 2 & \textbf{Balanced Acc}, $F1$ Score, AP \\ 

& \texttt{reward} & $ \textcolor{feedbackOnset}{\texttt{feedback\_time}}$ & $\textcolor{feedbackOnset}{\texttt{feedback\_time}} + 1.0 $ & 2 & \textbf{Balanced Acc}, $F1$ Score, AP \\
\bottomrule
\end{tabular}
}
\end{table}

%% file: appendix/task_suites/ts2.tex
\subsection{TS2 Details: Neural Activity Prediction}
\label{app:ts2}

\begin{table}[h]
\centering
\caption{Summary of pretraining evaluation task associated with TS2 (Neural Activity Prediction).}
\resizebox{0.99\columnwidth}{!}{
\begin{tabular}{clcccccc}
\toprule
\textbf{Task} & \textbf{Window Type} & \textbf{Interval Start} & \textbf{Interval End} & \textbf{Mask Ratio} & \textbf{Mask Dimension} & \textbf{Criterion} & \textbf{Metric} \\
\midrule
\texttt{co-smoothing} & rolling (20\,ms) & $\textcolor{StimulusOnset}{\texttt{stim\_on\_time}} - 1.0$ & $\textcolor{feedbackOnset}{\texttt{feedback\_time}} + 1.0$ & 10\% & neuron & random held-out$^{\dagger}$ & $\boldsymbol{D^2}$, bps \\
\texttt{forecasting}  & rolling (20\,ms) & $\textcolor{StimulusOnset}{\texttt{stim\_on\_time}} - 1.0$ & $\textcolor{feedbackOnset}{\texttt{feedback\_time}} + 1.0$ & 10\% & time   & deterministic last timestep & $\boldsymbol{D^2}$, bps \\
\bottomrule
\end{tabular}
}
\vspace{0.5em}
\begin{minipage}{0.99\columnwidth}
{\footnotesize $^{\dagger}$The held-out neuron mask used for evaluation is fixed and accessible only on the test set, to prevent optimization against the evaluation criterion during training.}\\
{\footnotesize Note: $D^2$ and bps are defined in Appendix~\ref{app:metrics}.}
\end{minipage}
\end{table}

\begin{figure}[ht]
    \centering
    \includegraphics[width=\linewidth]{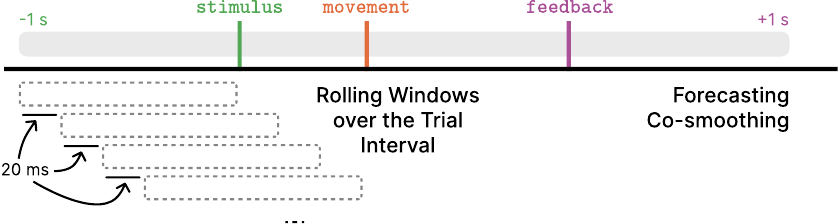}
    \caption{\textbf{Evaluation Windows for TS2.} 
    Each trial spans from 1\,s before stimulus onset to 1\,s after feedback, 
    with key events (stimulus, movement, feedback) marked along the timeline. 
    Context windows of 1\,s are extracted using a rolling procedure with a 20\,ms 
    stride across the full trial interval, and serve as inputs for both the 
    co-smoothing and forecasting tasks.}
    \label{fig:task_intervals_ts2}
\end{figure}

\paragraph{Tasks.}
Task Suite 2 (TS2) evaluates models on neural activity prediction: the ability to reconstruct masked portions of population spiking activity from the observed remainder, without any behavioral supervision. TS2 comprises two complementary tasks (see Table~\ref{fig:task_intervals_ts2}). In \emph{co-smoothing}, a fixed 10\% of units per session are designated as held-out at data preparation time (using a fixed random seed). During inference, the model observes the remaining 90\% of units and must predict the held-out units' spike counts. In \emph{forecasting}, the model observes the first 90\% of each context window and must predict the final 10\% of time bins. Both tasks use 20\,ms bins and a 1\,s context window, yielding 50-bin inputs and 5-bin targets for forecasting, and a per-session unit mask for co-smoothing.

\paragraph{Splits and Provided Labels.}
TS2 uses a different temporal splitting strategy than TS1: trials are assigned to interleaved 5-minute blocks (see Figure~\ref{fig:temporal_splits} and Appendix~\ref{app:within_session_splits} for details). Unlike TS1, no behavioral labels are provided: the prediction targets are the binned spike counts of the held-out units (co-smoothing) or future time bins (forecasting), derived directly from the neural data.

\paragraph{Evaluation Paradigm.}

TS2 evaluation measures how well a model predicts unobserved neural activity within a context window sampled from the test split of each session. Evaluation windows are extracted using a rolling procedure across the full trial interval, rather than being anchored to a specific task epoch, to avoid conflating  prediction performance with the stereotyped and highly predictable neural dynamics that tend to dominate task-aligned windows. For the baselines reported here, models are fully fine-tuned on the training and validation splits of each session. All reported numbers are computed on windows drawn sequentially from the test split with a stride equal to the bin size 20\,ms (see Figure~\ref{fig:task_intervals_ts2}).

\paragraph{What Is Enforced, Recommended, and Left Open.}
We distinguish between three levels of constraint in the evaluation protocol.

\textit{Enforced.} The test split always corresponds to the designated test block, ensuring temporal consistency across methods. For co-smoothing, the held-out unit mask is fixed at data preparation time (10\% of units per session, drawn with a fixed seed) and must not be modified.  For forecasting, the masked portion is always the final 10\% of the context window. The bin size (20\,ms), the held-out masks, and the context window length at test time (1\,s) must all remain unchanged. To ensure that only high-quality spikes contribute to the reconstruction target, both tasks are required to be evaluated with quality-controlled (QC) units (see Appendix~\ref{app:data_qc}.

\textit{Recommended.} We recommend training with a Poisson negative log-likelihood loss (with log-rate inputs).

\textit{Left open.} The context window size during training is left open; our baselines use 1\,s, but longer windows may benefit temporal forecasting.

\paragraph{Split Statistics.}
Table~\ref{tab:ts2_splits} reports the size of each split under three tokenization
schemes: spike tokens, used by event-based models such as POYO; unit-bin tokens, used by patch-based models such as NDT2; and time-step tokens, used by models that keep units in the channel dimension, such as NDT-Stitch. 
Every baseline we report in TS2 uses the time-step scheme. Token counts assume the default 1\,s context window and 20\,ms bins.
Unlike TS1, whose windows are disjoint trials, the validation and test samplers step through each session at 0.5\,s and at one bin respectively, so their windows overlap and each spike is tokenized roughly twice at validation and thirty-five times at test.
Task-specific masking reduces the input further at evaluation time.

\begin{table}[h]
\centering
\caption{TS2 split statistics on the \texttt{selected\_units} build. A sample is a 1\,s
window drawn by the split's sampler, and duration is the total span of the sampling
intervals the sampler draws from. Spike tokens count one token per spike; unit-bin tokens
count one per (unit, 20\,ms bin); time-step tokens count one per 20\,ms bin with units in
the channel dimension. All three are counted as sampled, so overlapping windows count the
same spike more than once.}
\label{tab:ts2_splits}
\small
\begin{tabular}{lrrrrrr}
\toprule
& & & & \multicolumn{3}{c}{Tokens} \\
\cmidrule(lr){5-7}
Split & Units & Samples & Duration (h) & Spike & Unit-bin & Time-step \\
\midrule
train & 3{,}778 &  19{,}033 & 6.05 &    28.7\,M &   122.5\,M &      951{,}650 \\
val   & 3{,}778 &  16{,}952 & 2.57 &    25.8\,M &   109.8\,M &      847{,}600 \\
test  & 3{,}778 & 700{,}471 & 5.59 & 1067.5\,M & 4529.5\,M & 35{,}023{,}550 \\
\bottomrule
\end{tabular}
\end{table}

\paragraph{Baselines.}
All TS2 baselines utilize neural activity as the sole input, without auxiliary signals or anatomical labels. An exception is MtM, which leverages brain region designations to define its masking schema. For pretrain model, we report results following full fine-tuning on the training and validation splits, with performance averaged across 5 random seeds.

\paragraph{Metrics.}
Both tasks are evaluated using two complementary metrics that assume a Poisson spiking model: the Fraction of Deviance Explained ($D^2$) and Bits Per Spike (bps). Both metrics are defined in detail in Appendix~\ref{app:metrics}.

%% file: appendix/task_suites/ts3.tex
\subsection{TS3 Details: Neuron Identity Prediction}
\label{app:ts3}

\paragraph{Tasks.}
Task Suite 3 (TS3) assesses whether models can infer anatomical position of neurons in a zero-shot manner. Such a task simultaneously probes representations for captured anatomical structure, and serves as a proxy for in vivo histology as a downstream application~\cite{liu2021accurate}.

\paragraph{Splits and Provided Labels.}
TS3 is unique in that splits are defined over units instead of time. In particular, we aggregate all units over animals in the pretraining corpus to define the training set of brain region labels. Then the evaluation set forms the test labels.

\paragraph{Evaluation Paradigm.}
For a model to be evaluated on TS3, it must produce neuronal unit-level embeddings that can be probed on the brain region classification task. These embeddings are derived directly from neural activity, including spike trains used as input for TS1 and TS2, as well as metadata such as spike waveforms. Different architectures can represent neuronal identity in different ways, hence every model has its own designation for what constitutes unit embeddings. However, in the end, the model must be able to produce two sets of embeddings $E_{\text{train}}\in\mathbb{R}^{N_\text{pretrain}\times D}$ and $E_{\text{test}}\in\mathbb{R}^{N_\text{eval}\times D}$, where $N_\text{pretrain}$ and $N_\text{eval}$ are the number of units in the pretraining and evaluation corpora, respectively, and $D$ is an embedding dimension defined by the model.

\paragraph{What is enforced in TS3?} To ensure that only high-quality spikes contribute to the reconstruction target and that the neural identities are well-defined, TS3 requires high quality-controlled (QC) over units (Appendix~\ref{app:data_qc}). Aside from this requirement, the mechanism by which embeddings are generated is fully flexible.

\paragraph{Types of Probes.}
Given embeddings, TS3 provides a standardized set of probes from which a brain region classifier can be fit. Following \cite{yu2025in}, these probes are organized along two axes:
\begin{enumerate}
    \item Linear vs MLP: we either fit a Linear probe via the \texttt{LogisticRegression} classifier provided by \texttt{sklearn}, or train an MLP probe in \texttt{torch}. For the latter, we optimize hyperparameters using the Tree-structured Parzen Estimator (TPE) algorithm~\citep{bergstra2011tpe} implemented in Optuna~\citep{akiba2019optuna}. Refer to Table~\ref{tab:ts3_linear_probe_hp} for details on the hyperparameters of the standard Linear probe, and Table~\ref{tab:ts3_mlp_probe_hp} for the hyperparameter search space used for tuning the MLP probe.
    \item Single- vs multi-unit: in~\cite{yu2025in} it was found that by aggregating consensus from nearby neurons, region classification accuracy can be improved. We implement this approach in a standardized evaluation pipeline for which we call the \textit{multi-unit} probe. In particular, class predictions are averaged over all neurons in the vicinity of the target neuron, using probe depth to measure proximity.
\end{enumerate}

\paragraph{Evaluation Regimes.}
Although the brain region classification task is designed to test zero-shot transfer with respect to region labels, models may vary in how they are able to generate embeddings. If a model requires gradients to generate embeddings on a new set of neurons, then it needs to be fine-tuned on the evaluation sessions using its pretraining task. Since then the model's weights were updated to calibrate on the new session, it has become exposed to the new neural population. We call this the \textit{transductive zero-shot setting}. If, on the other hand, a model is able to generate embeddings in a forward pass without having to calibrate, we call this the \textit{inductive zero-shot setting}.

\paragraph{Metrics.}
Due to imbalance in the region distribution (see Figure~\ref{fig:brain_region_dist}), we report macro-F1 score as the primary metric.

\begin{table*}[t]
\caption{Hyperparameters used for TS3 Linear probe. Implemented as \texttt{LogisticRegression} classifier from \texttt{sklearn}. No hyperparameter tuning is performed as the solver is deterministic given fixed inputs.}
\label{tab:ts3_linear_probe_hp}
\begin{center}
\begin{small}
\begin{sc}
\begin{tabular}{lcr}
\toprule
Hyperparameter & Value \\
\midrule
Preprocessing               & \texttt{StandardScaler}                \\
Solver                      & \texttt{lbfgs}               \\
Inverse regularization $C$  & $1.0$                                  \\
Max iterations              & $1000$                                 \\
Tolerance                   & $10^{-5}$                              \\
Class weight                & \texttt{balanced}                      \\
\bottomrule
\end{tabular}
\end{sc}
\end{small}
\end{center}
\vskip -0.1in
\end{table*}

\begin{table*}[t]
\caption{Hyperparameter search space for TS3 MLP probe. Tuned for $100$ trials with $4$ concurrent trials at a time.}
\label{tab:ts3_mlp_probe_hp}
\begin{center}
\begin{small}
\begin{sc}
\begin{tabular}{lccr}
\toprule

Hyperparameter & Search Space & Sampling type \\
\midrule
\multicolumn{3}{l}{\scriptsize Model parameters} \\
Bin size (ms)               & $\{5,\, 10,\, 20,\, 40\}$                      & Categorical \\
Depth                       & $\{1,\, 2,\, 3\}$                              & Discrete \\
Hidden dimension            & $\{32,\, 64,\, 128,\, 256,\, 512\}$            & Discrete ($\log_2$) \\
Dropout                     & $\{0.0,\, 0.2,\, 0.4,\, 0.6\}$                 & Discrete (step $0.2$) \\
Batch norm                  & $\{\text{True},\, \text{False}\}$              & Categorical \\
\midrule
\multicolumn{3}{l}{\scriptsize Training parameters} \\
Number of epochs            & $\{1000,\, 3000,\, 5000\}$                     & Discrete (step $2000$) \\
Batch size                  & $\{64,\, 128,\, 256,\, 512,\, 1024\}$          & Discrete ($\log_2$) \\
Weight decay                & $[10^{-6},\, 10^{-1}]$                         & Loguniform \\
Base learning rate$^\dagger$ & $[10^{-6},\, 10^{-2}]$                        & Loguniform \\

\bottomrule
\end{tabular}
\end{sc}
\end{small}
\end{center}
\vskip -0.1in
\end{table*}

%% file: appendix/metrics.tex
\section{Metrics and Ranking}
\label{app:metrics}

\subsection{Fraction of Deviance Explained under Poisson Model (\texorpdfstring{$D^2$}{D2})}

\paragraph{Poisson negative log-likelihood.}
For a single observation with predicted log-rate $\rho_i$ and observed spike count
$n_i \in \mathbb{N}_{\geq 0}$, the Poisson negative log-likelihood (NLL) is
\begin{equation}
    \ell(\rho_i;\, n_i)
    = e^{\rho_i} - n_i\,\rho_i + \ln(n_i!).
\end{equation}
Summed over all $T$ time bins, the total NLL of a model is
$\mathcal{L} = \sum_{i=1}^{T} \ell(\rho_i;\, n_i)$.

\paragraph{Three reference models.}
Fraction of Deviance Explained~\cite{cameron1997r,xia2025inpaintingneuralpictureinferring} is defined with respect to three nested models.

\begin{enumerate}
    \item \textbf{Predicted model.}
    The model of interest outputs log-rates $\{\rho_i\}$, giving
    \begin{equation}
        \mathcal{L}_{\mathrm{pred}}
        = \sum_{i=1}^{T}\bigl(e^{\rho_i} - n_i\,\rho_i + \ln(n_i!)\bigr).
    \end{equation}

    \item \textbf{Saturated model.}
    The ideal model sets $\hat{\lambda}_i = n_i$ exactly, achieving the
    minimum achievable NLL:
    \begin{equation}
        \mathcal{L}_{\mathrm{sat}}
        = \sum_{i=1}^{T}\bigl(n_i - n_i\ln n_i + \ln(n_i!)\bigr),
    \end{equation}

    \item \textbf{Null model.}
    The baseline predicts a constant rate equal to the empirical mean
    $\bar{n} = N/T$, where $N = \sum_{i=1}^T n_i$:
    \begin{equation}
        \mathcal{L}_{\mathrm{null}}
        = \sum_{i=1}^{T}\bigl(\bar{n} - n_i\ln\bar{n} + \ln(n_i!)\bigr).
    \end{equation}
\end{enumerate}

\paragraph{Fraction of Deviance Explained-\texorpdfstring{$D^2$}{D2}.}
The metric is defined as
\begin{equation}
    D^2
    = 1 - \frac{\mathcal{L}_{\mathrm{pred}} - \mathcal{L}_{\mathrm{sat}}}
                {\mathcal{L}_{\mathrm{null}} - \mathcal{L}_{\mathrm{sat}}}.
\end{equation}
It equals $1$ when the predicted model matches the saturated model, $0$ when
it matches the null model, and is negative when the model is worse than the
null. Reported $D^2$ values are clipped to 0.

\subsection{Bits Per Spike (\texorpdfstring{bps}{bps})}
\label{app:metric_bps}




Bits per spike (bps) measures the log-likelihood gain of the predicted model over the null model,
normalized by the total spike count and converted to base-2 bits:
\begin{equation}
    \mathrm{bps}
    = \frac{\mathcal{L}_{\mathrm{null}} - \mathcal{L}_{\mathrm{pred}}}
           {N_{\mathrm{sp}}\,\ln 2}.
\end{equation}
A positive bps indicates the model predicts spike timing more accurately than
the mean-rate baseline; $\mathrm{bps} = 0$ corresponds to the null model.



\subsection{Coefficient of Determination ($R^2$)}
For a continuous target with observations $y_i$, predictions $\hat{y}_i$, and mean $\bar{y} = \frac{1}{T}\sum_{i=1}^T y_i$,
the coefficient of determination is
\begin{equation}
R^2 = 1 - \frac{\sum_{i=1}^T (y_i - \hat{y}_i)^2}{\sum_{i=1}^T (y_i - \bar{y})^2}.
\end{equation}
It equals 1 for a perfect prediction, 0 when the model matches the mean baseline, and is negative when
the model is worse than the mean baseline. Reported $R^2$ values are clipped to 0.

Note that $R^2$ is the $D^2$ metric under a Gaussian observation model~\cite{mccullagh1989generalized}. Under this model, the notion of bits-per-spike (\ref{app:metric_bps}) becomes proportional to MSE, and the saturated model is $0$ (achieved under Gaussian noise). Hence, indeed $D^2$ reduces to $R^2$.

\subsection{Rank Aggregation Procedure}
\label{app:rank_procedure}

To summarize performance across the many task-specific metrics reported within each task suite, we compute an average rank for each model that reflects statistical significance in point estimates. This follows the general practice of rank-based aggregation across evaluation sets~\citep{demsar2006statistical}, gating rank assignment on pairwise significance testing rather than raw score differences alone~\citep{dror2018hitchhiker}. We treat each evaluation session as a separate dataset: for a fixed task and session, we rank models using the step-down procedure in Algorithm~\ref{alg:rank_step_down}, and average the resulting ranks across sessions to obtain the per-task rankings reported in Tables~\ref{tab:ts1_avg_ranks_app},~\ref{tab:ts2_avg_ranks_app_nostat}. These per-task rankings are in turn averaged across tasks to obtain the overall average rank reported in Tables~\ref{tab:ts1_main} and~\ref{tab:table2}.

For each model, we take the point estimate to be the mean score over 5 random seeds, and use the standard error of the mean (SEM) over seeds to quantify variance for the significance test in Algorithm~\ref{alg:rank_step_down}.

\begin{algorithm}[ht]
\caption{Step-down rank assignment for a single task and session}
\label{alg:rank_step_down}
\begin{algorithmic}[1]
\Require Models $\{m_1, \dots, m_K\}$ with per-seed scores; significance level $\alpha$
\State Compute mean score $\bar{s}_k$ and SEM $\sigma_k$ over seeds for each model $m_k$
\State Sort models by $\bar{s}_k$ so that $m_{(1)}$ is best-performing; let $\pi$ denote this order
\State $\text{rank}[\pi(1)] \gets 1$
\State $a \gets \pi(1)$ \Comment{current anchor}
\For{$i = 2$ \textbf{to} $K$}
  \State $m \gets \pi(i)$
  \State $p \gets$ one-sided Welch's $t$-test, $H_1$: anchor $a$ outperforms $m$
  \If{$p \geq \alpha$}
    \State $\text{rank}[m] \gets \text{rank}[a]$ \Comment{not significantly worse than the anchor}
  \Else
    \State $\text{rank}[m] \gets i$ \Comment{standard competition (``1224'') ranking}
    \State $a \gets m$ \Comment{$m$ becomes the new anchor}
  \EndIf
\EndFor
\State \Return $\text{rank}[m_1], \dots, \text{rank}[m_K]$
\end{algorithmic}
\end{algorithm}

Each comparison is made only against the current anchor rather than against all previously ranked models; a model is assigned a new, strictly lower rank only once it is shown to be significantly worse than the best remaining anchor. This anchor-based, sequential structure means a group of models found statistically indistinguishable from one another all receive the same rank, and the next distinct rank is set to the group's position in the sorted order (line 10) rather than incrementing by one (e.g., a tied pair at rank~2 is followed by a rank of~4, not~3). This grouping behavior parallels compact letter displays used to summarize sets of pairwise comparisons in applied statistics~\citep{piepho2004algorithm}, though we note the same caveat that applies to that literature: a shared rank here reflects a failure to reject the null hypothesis of no difference at $\alpha$, not demonstrated equivalence between models.

We assign the tied rank itself using standard competition (``1224'') ranking, computed via \texttt{scipy.stats.rankdata(..., method=`min')}~\citep{2020SciPy-NMeth}, rather than the fractional (mid-rank) ranking conventionally used when averaging ranks in the style of~\cite{demsar2006statistical}. We make this choice because competition ranking directly encodes the quantity of interest for this benchmark: a model's rank equals one plus the number of models shown to be significantly better than it. Fractional ranking would instead average a tied group into the ranks its members would have occupied absent ties, which can obscure the significance information. We use $\alpha = 0.05$ throughout. Algorithm~\ref{alg:rank_step_down} is applied independently per task and session; the resulting ranks are averaged first over sessions (Appendix~\ref{app:additional_results}) and then over tasks to yield the average rank columns reported in the main text.

%% file: appendix/model_overview.tex
\section{Overview of Baseline Models}
\label{app:model_overview}





We implement a panel of baselines reflecting two complementary goals. First, we establish task-specific reference points using specialized single-session baselines, which serve as points of comparison on individual tasks.  Second, we pretrain typical neural data methods on a shared corpus and evaluate them downstream on applicable tasks, providing a starting point for the community to compare approaches and identify gaps.    

In the following sections, we provide background on the various aspects of design involved in constructing models for neural data. These include tokenization schemes or input processing, model architectures, and training objectives. In Section~\ref{app:hyperparameters}, we describe how we defined and tuned hyperparameters for the different task suites, including single-session baselines, pretraining, and finetuning on individual tasks.

\subsection{Tokenization Schemes}
\label{app:tokenization}

Models of neural data can accept various modalities of input, including spike trains (encoding exact timing information along with neuron identity), spike waveforms, and local field potential (LFP). A major part of the design space involves choosing how input data should be structured while the model extracts representations, in such a way that is compatible with their choice of architecture, training objective, and underlying assumptions. This choice is often denoted \textit{tokenization}, popularized by its use in other Machine Learning domains such as Natural Language Processing~\cite{sennrich2016neural} and Computer Vision~\cite{dosovitskiy2020image} to indicate the fundamental units of information as seen by the model.

Note that while the current benchmark does not include a task suite or baseline that utilizes LFP data, it is an interesting direction for future extensions of the benchmark.

\subsubsection{Binning and Patching}
\label{app:tokenization_binning}

A natural choice for representing neural spike trains is to convert the discrete event sequences into bin counts in a regularly spaced grid. The idea is to derive an approximation of the instantaneous firing rate, motivated by the assumption that neural information is carried via rate coding~\cite{adrian1926impulses}. The choice of bin size modulates the tradeoff between smooth rate estimates and precise timing.

Given a raster of $T$ bin counts over a fixed context length from $N$ neurons, the next choice is how to ingest this spatiotemporal grid into a model. One idea is to flatten the bins into $N\cdot T$ tokens, treating them all the same throughout computations. This approach loses precise temporal and spatial information unless it is explicitly included via positional information, and could result in large computational complexity given many neurons. At the same time, this approach introduces minimal inductive biases and is a typical choice for simple baselines such as Linear and MLP.

Several works have explored alternative tokenization schemes given binned spikes. NDT~\cite{ye2021representation} extracts tokens as temporal \textit{patches} by first slicing multiple timestamps at a time, then projecting the spatial dimension into a shared embedding space. STNDT~\cite{le2022stndt} separately tokenizes spatial and temporal slices, applies independent attention modules, and then fuses the features. NDT2~\cite{ye2024neural} adopts a ViT~\cite{dosovitskiy2020image} style tokenization scheme by treating the $N\times T$ raster as an image and applying both spatial and temporal patching. Different approaches yield tradeoffs in terms of computational complexity and spatial or temporal resolution throughout the architecture.

\subsubsection{Spike Tokenization}
\label{app:tokenization_spike}

An alternative philosophy is to retain exact timing information throughout computations, motivated by the idea that neural information could be carried via temporal coding~\cite{thorpe1998rank}. From the modeling perspective, this approach treats every unit of communication between neurons, i.e. exact spike times along with the identity of the neuron, as a token. Such an approach was introduced by POYO~\cite{azabou2024unified} and extended to follow up works~\cite{ryoo2025generalizable}. The advantage of this tokenization scheme is that it is not reliant on a fixed bin size, considering optimal temporal resolutions can vary between neural circuits and some circuits may even exhibit temporal multiplexing at multiple timescales~\cite{panzeri2010sensory}. On the other hand, such an approach requires unit-level tokenization resulting in very long token sequences, especially with a large number of neurons or very active neurons.

\subsubsection{Inter-Spike Interval (ISI) Distributions}
\label{app:tokenization_isi}

Another perspective is that information lies in relative timing, which can be captured by the inter-spike interval (ISI) distribution \citep{perkel1967neuronal}. For a given neuron's spike train, its ISIs are computed as the time differences between consecutive spikes within a recording window and then binned into a histogram. The choice of binning approach should aim to distribute the probability mass of the ISIs uniformly across bins. ISI distributions are typically heavily right-skewed.

A linearly spaced binning approach, which assigns equal resolution across the entire ISI range, concentrates the majority of the probability mass into the first few bins, creating a highly non-uniform distribution that obscures fine-grained dynamics. Conversely, log-spaced binning grants finer resolution to shorter intervals, resulting in a more uniform representation that preserves the fine-grained temporal precision of short-interval events \citep{dorval2008probability}.

The recording window length introduces another tradeoff. A longer window, such as a per-session window, can provide a more stable estimate of the ISI distribution as it contains more spikes, and therefore more intervals \citep{gerstner2014neuronal}. However, because a per-session window pools ISIs across an entire session, it may obscure fine-grained local temporal dynamics. In contrast, per-trial or fixed-context windows can better preserve local spiking patterns, though they may yield noisier ISI estimates.

\subsubsection{Autocorrelograms (ACG)}
\label{app:tokenization_acg}

Similar to the ISI distribution, an autocorrelogram (ACG) also captures relative timing information by treating each spike as a reference event and constructing a histogram of pairwise time lags to other spikes from the same neuron \citep{perkel1967neuronal}. Unlike the ISI distribution, which only considers intervals between consecutive spikes, the ACG includes delays between every pair of spikes within a fixed lag window. This allows the ACG to capture repeated temporal patterns in the neuron's spike train.

As with ISI histograms, the lag bins may be linearly or logarithmically spaced, with log-spaced bins preserving finer resolution at short lags while compressing longer lags into broader bins \citep{dorval2008probability}. This is useful because bins near zero lag can capture closely spaced spike pairs, including burst firing patterns and refractory-period effects.

\subsubsection{Spike Waveforms}
\label{app:tokenization_waveform}

Neural activity can also be represented through spike waveforms, where each waveform captures the extracellular action potential (EAP) measured by extracellular recording electrodes during a short time window around each detected spike \citep{gold2006origin}. Whereas tokenization schemes such as ISI distributions and ACGs capture spike timing, EAP waveforms encode the shape of the recorded spike itself.

The shape of a spike waveform is often characterized by a prominent negative and positive peak, reflecting the extracellular signature of a neuron's action potential \citep{gold2006origin}. Because this waveform shape can vary greatly across neuron classes, common neuroscience problems such as cell-type classification often use waveform-derived features, such as waveform width and peak-to-trough amplitude \citep{yu2025in}.


\subsection{Models}

\subsubsection{Linear and Multi-Layer Perceptron (MLP)}
\label{app:model_overview_linmlp}

A straightforward baseline for decoding involves mapping flattened bin counts directly into the target readout. Due to the complexity of loss functions for certain decoding tasks (e.g.,~masked MSE for paw kinematics, Poisson NLL for licking rate), we standardized training linear models using gradient descent.

We can introduce nonlinearity into the mapping from flattened bins to target readout via a Multi-Layer Perceptron (MLP). Note that in initial experiments with the MLP, we found this fully flattened binned counts representation performed better than taking temporal or spatial slices. These baselines introduce very little inductive bias on the input and output structures, which can make them flexible but difficult to train with limited data.

See Appendix~\ref{app:hyperparameters_linear} and~\ref{app:hyperparameters_mlp} for hyperparameter selection of Linear and MLP models, respectively.

\subsubsection{Convolutional Neural Network (CNN)}
\label{app:model_overview_tcn}

A standard approach for extracting local, translation-invariant temporal features is the Convolutional Neural Network (CNN). We start with a typical implementation of convolutional layers on time-series data using the Temporal Convolutional Network (TCN), which achieves wide receptive fields via stacked dilated convolutions~\cite{lea2016temporal}. However, we retain a dilation factor of $1$ for simplicity while defining the search space.
Input features are temporal slices over a binned raster, where the bin size is a model hyperparameter. The result is a temporal resolution that is directly modulated by the bin size throughout the model architecture. To support both sequence- and frame-level tasks, we attach adaptive average pooling before a final linear readout.

In the CNN, convolutions are interleaved with nonlinear activations providing a structure similar to the MLP but with strong inductive biases of locality and translation invariance along the temporal axis. Note, however, that if the activations were replaced with \textit{identities}, the convolutions would collapse to a single linear map from a fixed context window of binned counts to the readout.

A key benefit of this representation is that the CNN then expresses a standard linear baseline for neural decoding of continuous variables, wherein a short context window is used to readout the next timestep in an autoregressive fashion. For example, in the IBL BrainwideMap~\cite{international2025brain}, continuous targets like whisker motion energy and wheel speed are regressed in this way using a $200$ms sliding context window of population activity. The CNN thus strictly contains this linear baseline as a special case.

See Appendix~\ref{app:hyperparameters_tcn} for more details on hyperparameter selection.

\subsubsection{Gated Recurrent Unit (GRU)}
\label{app:model_overview_gru}

Recurrent architectures offer a natural inductive bias for sequential neural data by maintaining a hidden state that is updated at each timestep, implicitly integrating information over arbitrary temporal horizons. The earliest such model, the Recurrent Neural Network (RNN)~\cite{elman1990finding}, suffers from well-known training instabilities: gradients either vanish or explode when backpropagated through long sequences, making it difficult to capture dependencies beyond a short effective context window~\cite{279181}. The Long Short-Term Memory (LSTM)~\cite{6795963} addressed this by introducing a dedicated cell state governed by input, forget, and output gates, which create additive gradient paths that resist vanishing. The Gated Recurrent Unit (GRU)~\cite{cho2014learningphraserepresentationsusing}  simplified this design by merging the cell and hidden states into a single vector controlled by just two gates, a reset gate and an update gate, retaining comparable expressive power at reduced computational cost. 

This gating mechanism is closely related to State Space Models (SSMs)~\cite{gu2022efficiently}, as both learn the transition matrices of a linear dynamical system, with gating providing a data-dependent, nonlinear modulation of the dynamics. The SSM perspective has a long tradition in systems neuroscience, where latent linear dynamical systems and their nonlinear extensions have been used to infer low-dimensional population dynamics from high-dimensional spike trains~\cite{NIPS2008_ad972f10, NIPS2011_7143d7fb}, including approaches that couple Gaussian Process priors with RNN transitions for smooth latent inference~\cite{pmlr-v115-she20a}. The broader sequence-modelling literature has revisited linear recurrences for efficient long-range modelling (e.g., S4~\cite{gu2022efficiently}, Mamba~\cite{gu2024mamba}), a trend that has begun to influence neural decoding through architectures that exploit their fast inference properties (see Section~\ref{app:model_overview_possm}).

As with the CNN, input features are temporal slices over a binned raster with bin size as a hyperparameter. Binned spike counts are first projected to a fixed hidden dimension via a linear layer, then processed sequentially by a multi-layer GRU. To support both sequence- and frame-level tasks, adaptive average pooling is applied over the temporal dimension before a final linear readout.

See Appendix~\ref{app:hyperparameters_gru} for hyperparameter selection of the GRU model.

\subsubsection{TS2 Statistical Baselines}
\label{app:model_overview_stat_baseline}

Alongside the trained single-session baselines, TS2 includes a panel of statistical baselines: closed-form, zero-parameter methods that fit a handful of summary statistics on the training windows and select one or two hyperparameters on validation. Nothing is optimized by gradient descent, so these serve as a floor that quantifies how much of each TS2 task is solved by stationary firing statistics, instantaneous population coupling, or simple linear extrapolation, before any learned representation is involved. All methods emit log-rates and are scored with the same Poisson metrics as the trained models.

The two TS2 tasks hold out along orthogonal axes: co-smoothing holds out a subset of units and leaves the rest of the population simultaneously observed, whereas forecasting holds out the trailing timesteps and leaves only each unit's own past. A given method exploits exactly one of these structures, so each is declared for one task. Every method reads only observed entries, making them leakage-free by construction.

\textbf{Co-smoothing methods.}

\begin{itemize}
    \item \textbf{Population coupling.} A rank-one model in which each unit's rate is modulated by a single global population signal, $r_i(t) = \bar{r}_i \, p(t)^{\gamma}$, where $p(t)$ is a Gaussian-smoothed, mean-one normalized trace of the observed population activity and $\gamma$ controls the strength of coupling. The smoothing width $\sigma$ and the exponent $\gamma$ are selected on validation, and $\gamma = 0$ recovers the mean rate. This tests whether a single shared gain, with no unit-specific structure, explains the held-out activity.
    \item \textbf{RRR.} A reduced-rank ridge regression mapping the observed units to the held-out units at each timestep, with the rank $k$ and the ridge strength $\lambda$ selected on validation. Unlike population coupling, this captures unit-specific, multi-dimensional structure in the instantaneous population code, and it is the natural linear reference point for what a learned latent variable model must beat. 
    \item \textbf{RRR (w/ ISI features).} The same rank-constrained ridge regression, but with each observed unit contributing, in addition to its binned count, two timing features per bin: the time since its last spike and its most recent inter-spike interval (see Appendix~\ref{app:tokenization_isi}), computed from the raw spike times. Because the regressors now mix counts with seconds, features are standardized before the solve. This isolates the value of the sub-bin timing information that binning discards, holding the regression itself fixed.
\end{itemize}

For both reduced-rank readouts, the held-out set used at test is disjoint from the observed set, so predicting one from the other is leakage-free. Hyperparameters are selected on an independent validation hold-out draw and the readout is then refit for the test draw, so the scored units take no part in selection.

\textbf{Forecasting methods.}
\begin{itemize}
    \item \textbf{Trailing mean.} Each unit's horizon is predicted by the average of its own last $K$ observed bins, held flat across the forecast window, with $K$ selected on validation. This measures whether recent activity alone, without reference to session-level statistics, carries predictive signal about the immediate future.
    \item \textbf{Shrinkage.} A blend of each unit's trial-local rate with its global mean, $r_i = (1 - \alpha)\,\bar{r}_i + \alpha\, \ell_i$, where $\ell_i$ is the mean over the observed portion of the current window and $\alpha$ is selected on validation. The two endpoints are exactly the mean rate ($\alpha = 0$) and the trailing mean over the full observed window ($\alpha = 1$), so an interior optimum indicates that trial-to-trial rate fluctuations are informative but noisy enough to require regularization toward the session average.
    \item \textbf{Ridge autoregression.} A per-unit ridge regression from the last $L$ observed bins to each forecast bin, solved separately for every bin in the horizon, with $L$ and $\lambda$ selected on validation. Because each future bin gets its own weights, this captures the recent trend of a unit's rate rather than only its level, making it the strongest purely linear, single-unit forecaster in the panel.
\end{itemize}

See Appendix~\ref{app:hyperparameters_stat_baseline} for the search grids and selection protocol used for each method.

\subsubsection{Autoencoder (AE)}
\label{app:model_overview_ae}

The Autoencoder (AE) extends the MLP baseline toward latent variable modeling: rather than mapping directly from input to readout, it introduces a bottleneck that encourages the network to capture low-dimensional structure in the population response. The model is trained to reconstruct its own input under a Poisson negative log-likelihood loss, treating the decoder outputs as log-rates of a Poisson spike-count distribution. Following the same tokenization scheme as the Linear and MLP baselines 
(Section~\ref{app:model_overview_linmlp}), the input is flattened and passed through an encoder-decoder MLP pair. The architecture is a symmetric counterpart to the MLP: the encoder projects the 
representation down by a factor of $2$ at each layer, the decoder 
mirrors this by projecting back up by a factor of $2$, and a final 
linear readout maps the reconstruction to the full output 
dimensionality. This symmetric bottleneck can be seen as a nonlinear generalization of PCA applied jointly across time and neurons.
\subsubsection{LFADS}
\label{app:model_overview_lfads}

Latent Factor Analysis via Dynamical Systems (LFADS)~\cite{pandarinath2018inferring} is a sequential variational autoencoder that models a trial of population activity as the output of a low-dimensional nonlinear dynamical system. A bidirectional GRU encoder maps the binned spike raster to a posterior over the generator's initial condition, and a second bidirectional encoder feeds a controller GRU that infers a time-varying input sequence, capturing structure the autonomous dynamics cannot produce. A generator GRU is rolled out from the sampled initial condition under these inferred inputs, projected to a small set of latent factors, and read out linearly to per-unit Poisson log-rates. Training maximizes an ELBO with a Gaussian prior on the initial condition and an autoregressive prior on the inferred inputs, the latter keeping those inputs temporally smooth rather than pushing them toward white noise.

We follow AutoLFADS~\cite{keshtkaran2022large} in treating the regularizers as search parameters rather than fixed constants: the two KL weights, the L2 penalties on the generator and controller recurrent weights, dropout, and the coordinated dropout rate are all tuned per session and task. Coordinated dropout, which corrupts a random subset of input entries and grades the reconstruction only on them, is what blocks the identity solution available to an autoencoder with a per-unit readout. For co-smoothing we additionally blank whole unit rows at a tuned rate, matching the shape the hold-out takes at evaluation. For forecasting, the reconstruction gradient is taken either from the held-out tail alone or from the whole window, which is itself a tuned choice.

See Appendix~\ref{app:hyperparameters_lfads} for hyperparameter selection.

\subsubsection{LOLCAT}
\label{app:model_overview_lolcat}
Local Latent Concatenated
Attention (LOLCAT)~\cite{schneider2023transcriptomic} is a supervised model predicting neuron identity from neuronal activity timing data. Originally, it was used to predict a neuron's transcriptomic cell-type, but we are using it for brain region classification. For each neuron, spike trains within trial windows are converted into inter-spike interval (ISI) (see Appendix~\ref{app:tokenization_isi}) histograms using log-spaced bins. These histograms are passed through a per-trial encoder (i.e., MLP), and then aggregated across trials into a single per-neuron embedding via multi-head attention pooling. Finally, an MLP classifier is used to predict the brain region of each neuron.

To address class imbalance, LOLCAT utilizes an adaptive sampler that reweights classes during training. The sampler maintains class-specific sampling factors that are updated at each validation epoch using class-wise training and validation losses. It computes two scores: an overfitting score, which quantifies overfitting for a given class, and an undertraining score, which quantifies how well the model is performing on that class relative to the entire dataset.

LOLCAT also implements adaptive trial dropout, with the probability of dropping a trial being proportional to a specified hyperparameter $p$ and inversely proportional to the neuron's firing rate in that trial.

See Appendix~\ref{app:hyperparameters_lolcat} for more details on the implementation and hyperparameters used

\subsubsection{CEBRA}
\label{app:model_overview_cebra}

CEBRA~\citep{schneider2023cebra} is a self-supervised method that learns low-dimensional embeddings of neural data by contrasting neural activity against behavioral labels. We use the single-session variant of CEBRA, fitting a separate model for each session and training a MLP probe on the resulting embeddings to predict behavior. We do not use multi-session CEBRA, as it is incompatible with the \texttt{ts1} evaluation protocol: multi-session CEBRA requires all target sessions to be included during model fitting, and provides no standard mechanism for transferring pretrained model to held-out sessions. CEBRA supports three contrastive modes: a time mode, which contrasts neural activity based on temporal proximity alone; a behavior mode, which contrasts neural activity against behavioral labels; and a hybrid mode, which combines both time-contrastive and behavior-contrastive objectives. We treat the contrastive mode as a hyperparameter and select it per task and per session based on validation performance. In practice, we find that behavior mode is often detrimental for sequence-level tasks, whereas hybrid mode, which uses both temporal structure and behavioral supervision, tends to perform better on frame level tasks. Refer to Appendix~\ref{app:hyperparameters_cebra} for hyperparameter selection of the CEBRA model.

\subsubsection{POYO}
\label{app:model_overview_poyo}

POYO~\cite{azabou2024unified} is a method for pretraining on a large amount of neural spiking activity. It introduces spike tokenization (see Appendix~\ref{app:tokenization_spike}) in a Transformer-based backbone. In particular, it uses a PerceiverIO~\cite{jaegle2021perceiverio} backbone featuring a cross-attention layer on model inputs to compress the input token sequence (which can be quite long with a large number of units) into a relatively small latent token sequence. Following this compression, the operations no longer scale quadratically in time and space with the input size, allowing memory-efficient processing of individual spikes. The most expensive operation is the first cross-attention layer, hence computational budgets are primarily allocated with this initial layer in mind. The architecture uses a final cross-attention decoder to readout the target sequence. POYO is typically trained in a supervised fashion, where the target readout specifies the query sequence in the cross-attention decoder.

In the single-session, single-task case, we adapt the standard implementation~\cite{azabou2024unified} into our evaluation pipeline. In particular, we tokenize input spikes as $(u_i,t_i)$, where $u_i$ represents the unit identity mapping to a lookup table containing \textit{unit embedding}, and $t_i$ represents timing information relative to the context window. This input token sequence is compressed into the latent sequence via an input cross-attention layer (where learnable latent tokens are queries and the input tokens are keys/values), followed by several layers of self-attention on the latent sequence. On the decoder, a learnable \textit{session embedding} is repeated over the target sequence (e.g.,~the number of timestamps for frame-level tasks, or a single timestamp at the middle of the context window for sequence-level) to form the query sequence for a cross-attention decoder, where the latent tokens outputted from the final self-attention layer form the keys/values. After the final cross-attention layer the output tokens are projected via a final linear layer into the target dimension, depending on the task. POYO uses Rotary Position Embeddings (ROPE)~\cite{su2024roformer} to incorporate timing information into attention layers.

Note that in the single-session case, the set of learnable session embeddings is a singleton, hence the decoder is simply performing attention pooling where the session embedding is the learnable query. The notion of session embeddings becomes relevant in the multi-session setting. In the context of our benchmark, we only use multi-task POYO+~\cite{azabou2025multisession} for unified pretraining that can be applied downstream to any of the decoding tasks, and single-session POYO for tuning and for finetuning. The multi-session, single-task POYO variant is not relevant to the benchmark since it does not fit within our evaluation protocol, and in pretraining it cannot produce a single model that can be applied to any task.

Still, to benchmark how the multi-task variant compares to single-task variants on each task, we pretrained POYO single-task (POYO-ST) and finetuned on each evaluation session in a cursory study. This study mirrors experiments done in the POYO+~\cite{azabou2025multisession} paper on the Allen Institute's Brain Observatory dataset~\cite{de2020allenobs}, where the multi-task model is compared to single-task variants to observe the effect of multi-task learning. See Appendix~\ref{tab:poyo_add_results} for results and discussion on this study on our benchmark tasks and splits.

Note that when finetuning POYO+ on the evaluation sessions, we do retain the task-specific head and learnable task embeddings used in POYO+ pretraining, but restricted to the specific task at hand. This is an architectural addition to what is used for the single-session POYO baseline necessary to effectively finetune the pretrained variant. See Appendix~\ref{app:hyperparameters_poyo+} for more details on the finetuning implementation and hyperparameters used.

\subsubsection{POSSM}
\label{app:model_overview_possm}

POSSM~\cite{ryoo2025generalizable} is a hybrid architecture for real-time, causal neural decoding that pairs POYO-style spike tokenization and input/output cross-attention with a recurrent state-space model (SSM) backbone. The key departure from POYO is that the self-attention layers over the full latent sequence are replaced with a recurrent backbone that operates over short, contiguous, non-overlapping time chunks (typically 50~ms). The input cross-attention operates within each chunk, compressing variable-length spike token sequences $(u_i, t_i)$ into a single latent vector $z^{(t)}$, with ROPE~\cite{su2024roformer} encoding spike timing relative to the chunk and unit embeddings encoding unit identity, exactly as in POYO. This latent is then passed to the recurrent backbone whose hidden state $h^{(t)} = f_{\text{SSM}}(z^{(t)}, h^{(t-1)})$ propagates information across chunks. This enables constant-time updates as neural activity is streamed in, in contrast to POYO which reprocesses an entire context window at each new prediction. The output cross-attention decoder mirrors POYO's: queries built from a repeated learnable \textit{session embedding} combined with target timestamps (via ROPE) attend over the $k$ most recent hidden states ($k=3$ in the original work) to produce behavioral predictions, preserving POYO's flexibility to emit multiple, irregularly-spaced targets per chunk and to query timestamps beyond the current input chunk.

We adapt the standard implementation~\cite{ryoo2025generalizable} into our evaluation pipeline, with the only modification being that we use a non-causal, bi-directional recurrent backbone in place of the original uni-directional one to match other non-causal models used in TS1; all other architectural and tokenization choices follow~\cite{ryoo2025generalizable} exactly. In the single-session, single-task case, the set of learnable session embeddings is again a singleton and the decoder reduces to attention pooling, analogously to single-session POYO. For cross-session/cross-dataset pretraining, the unit and session embedding tables span all training sessions, yielding the pretrained \textit{o-POSSM} model that can be applied downstream to any of the benchmark decoding tasks. As with POYO+, the multi-session, multi-task pretrained variant produces a single model applicable to any task in the benchmark.

See Appendix~\ref{app:hyperparameters_possm} for more details on POSSM finetuning implementation and hyperparameters used.

\subsubsection{NDT}
\label{app:model_overview_ndt}

The Neural Data Transformer (NDT)~\cite{ye2021representation} is a standard transformer model operating on binned spikes. The original method was designed for latent variable modeling~\cite{pei2021neural} where input spike trains are encoded into latent factors that summarize the neural population codes. These latent factors can then be used to produce predicted firing rates and decode behavior. Hence, the original method  is inspired by BERT encoder~\cite{devlin2019bert} and used masked modeling as a training objective. See Figure~\ref{fig:masking_scheme} for a comparison of masking schemes used by various models that employ masked modeling on bins.

Our treatment of NDT goes beyond the training objective to consider the architecture as a fundamental approach for processing binned spike tokens in a transformer. Drawing on the comparison of tokenization schemes in Appendix~\ref{app:tokenization}, we consider a supervised variant of NDT (NDT Supervised) that can be directly used to perform on a downstream task of TS1, and as a single-session baseline it can be compared to the single-session POYO variant forming the two fundamental transformer baselines in TS1 (see Table~\ref{tab:ts1_main}). 

NDT is also used as a single-session baseline in TS2, but 
there we use the original BERT-inspired masking scheme for latent variable modelling, which can then be evaluated on both co-smoothing and forecasting tasks. Masking is applied along the temporal dimension across all neurons simultaneously, such that contiguous blocks of timesteps are masked from the model. Masked bins are replaced by one of three augmentations sampled during training: the spike count is zeroed out, replaced by a random value, or left unchanged, following the token corruption strategy of BERT~\cite{devlin2019bert}.

Finally, we pretrain NDT (NDT-stitch) with its masked modeling 
objective across all pretraining sessions with downstream evaluation on 
TS1 and TS2. To generalize across sessions and animals that have 
varying numbers of units, common in practice due to spike sorting, we 
adopt the session-specific stitcher technique 
introduced in~\cite{ye2024neural, zhang2024towards}. The stitcher is a 
lightweight linear or MLP layer that projects the variable neuron 
counts of each session into a fixed-dimensional latent space, allowing 
the transformer backbone to share weights across all sessions while the 
stitcher weights remain session-specific. When transferring to a new 
session or animal during fine-tuning, a new stitcher is initialized 
from scratch while the pretrained backbone is retained.

For further details on hyperparameter selection, see 
Appendix~\ref{app:hyperparameters_ndt_ts1}
 for the supervised 
single-session NDT baseline (TS1), Appendix~\ref{app:hyperparameters_ndt_ts2} for the single-session NDT 
baseline (TS2), and Appendix~\ref{app:hyperparameters_ndt_stitch} for the pretrained NDT-stitch model.

\subsubsection{MtM}
\label{app:model_overview_mtm}

Multi-task masking (MtM) \cite{zhang2024towards} is a self-supervised transformer model for neural activity reconstruction. MtM uses the same architecture as NDT, but replaces the standard random timestep masking objective with multiple masking strategies designed to capture the spatiotemporal structure of neural population activity (see Figure~\ref{fig:masking_scheme}). These objectives include: (1) causal masking, which masks future time steps and predicts them from past activity, corresponding to forecasting in TS2; (2) neuron masking, which randomly masks neurons and reconstructs them from the remaining neurons, corresponding to co-smoothing in TS2; (3) intra-region masking, which masks randomly sampled neurons within a randomly selected brain region and reconstructs them using unmasked neurons from the same region; and (4) inter-region masking, which masks all neurons within a brain region and reconstructs them using neurons from other regions. For each masking scheme, we apply input masking by zeroing out the masked portions of the data. Each of these masking schemes teaches the model about different structure in neural populations. During training, the model alternates across these objectives, enabling it to learn latent neural representations that support multiple inference-time tasks. In addition, we prepend a learnable prompt token to the transformer to indicate the masking scheme on which the model is being trained. This token can also be provided at test time to adapt the model to different downstream tasks. 

We pretrain MtM with its designed masked modeling objective across all pretraining sessions with downstream evaluation on TS1 and TS2. We evaluate transfer both within the pretraining distribution (TS2) and in a more challenging out-of-distribution setting (TS1). For multi-session pretraining, we prepend a session identity token to indicate the session from which the neural data originates. To accommodate varying neuron counts across sessions, we adopt a similar strategy to NDT-Stitch. Specifically, we use a session-specific linear read-in layer to map neural activity from each session into a shared hidden dimension, followed by a shared MLP applied across all sessions (this shared MLP differs from our NDT-Stitch implementation). We additionally employ session-specific read-out stitchers that map the learned latent representations back to the original neuron count of each session. 

See Appendix~\ref{app:hyperparameters_mtm} for more details on MtM finetuning implementation and hyperparameters used.

\subsubsection{NuCLR}
\label{app:model_overview_nuclr}

NuCLR~\cite{arora2025know} is a self-supervised framework for learning neuron-level representations from large-scale neural population activity, designed to capture identity-relevant attributes such as cell type and brain region. NuCLR produces a representation \textit{per neuron} directly from population activity, without assuming any fixed neuron ordering or requiring session-specific alignment. Inputs are constructed by binning each neuron's spike train at a fixed bin-size (20~ms by default) and partitioning the bins into non-overlapping temporal patches of length $T_{\text{patch}}$, which are linearly projected into a $D$-dimensional latent space to yield a token sequence per neuron. These tokens are then processed by a spatiotemporal transformer that alternates between two types of attention: \textit{temporal attention layers} that operate independently on each neuron's patch sequence (with ROPE~\cite{su2024roformer} encoding relative patch timing), and \textit{spatial attention layers} that, at each time index, attend across neurons in the population to inject population-level context into each neuron's representation. The model uses $L_T$ purely temporal layers followed by $L_{\text{ST}}$ spatiotemporal layers (each combining a spatial block with a temporal block), and finally mean-pools each neuron's tokens over the temporal axis to yield a fixed-dimensional vector per neuron. The architecture is permutation-equivariant across neurons and accepts populations of arbitrary size, allowing the same encoder to be applied to recordings with different numbers of neurons.

NuCLR is trained with a sample-wise contrastive objective inspired by SimCLR~\cite{chen2020simpleframeworkcontrastivelearning}. Two temporally-spaced views of the same population are sampled within $\Delta T_{\max}$ of each other and independently subjected to neuron dropout (up to 50\%) to encourage robustness to partial observations. Both views are passed through the spatiotemporal encoder and a projection head $g(\cdot)$, and an InfoNCE-style loss treats representations of the same neuron across the two views as positives while treating all other neurons in the same population as negatives. Crucially, negatives are restricted to within-recording neurons (and, for electrophysiology, within a single probe insertion) to avoid the trivial-negative problem that arises when neurons from unrelated recordings are pooled. We adapt the standard implementation~\cite{arora2025know} into our evaluation pipeline without modification. Following the linear-evaluation protocol used in the original work, we freeze the pretrained encoder, compute a single representation per neuron by averaging encoder outputs over windows sampled from each session, and train a linear classifier on top for downstream cell-type and brain-region classification.

\subsubsection{NEMO}
\label{app:model_overview_nemo}
Neuronal Embeddings via Multimodal contrastive learning (NEMO)~\cite{yu2025in} is a multimodal contrastive learning framework for learning representations of individual neurons from neurophysiological data. Specifically, it jointly embeds two complementary views of the same neuron: activity autocorrelograms (ACGs) and average extracellular waveforms (see appendices~\ref{app:tokenization_acg} and \ref{app:tokenization_waveform}). This method is motivated by the assumption that these two modalities carry shared information that is more discriminative of neuron identity than either modality alone, yielding representations useful for downstream cell-type and brain region classification tasks.

To get individual ACG and waveform representations, NEMO utilizes separate encoders for the two modalities: a 2-layer CNN for ACGs and a 2-layer MLP for waveforms. Then, both representations are projected into a shared embedding space via linear projection heads to a common dimensionality. A CLIP-style contrastive loss~\cite{radford2021learning} is used to treat each unit's (waveform, ACG) pair as a positive pair and all cross-unit pairs as negatives. For downstream evaluation, the projection heads are discarded, and the pre-projection encoder representations are concatenated, with the ACG representation first, to produce a final unit embedding. The resulting embeddings are then evaluated in a supervised manner via linear and MLP probing for cell-type and brain region classification.

The original NEMO model applies modality-specific data augmentations: ACG images are augmented with temporal Gaussian smoothing, temporal jitter, amplitude scaling, additive Gaussian noise, and multiplicative pepper noise, whereas waveform templates use additive Gaussian noise.

See Appendix~\ref{app:hyperparameters_nemo} for more details on the implementation and hyperparameters used.

\subsubsection{NDT2}
\label{app:model_overview_ndt2}

NDT2~\cite{ye2024neural} is a transformer model for multi-context pretraining on binned neural spiking activity. It adopts an asymmetric encoder-decoder architecture: a large transformer encoder processes the masked spike bin sequence into latent representations, while a lightweight 2-layer transformer decoder reconstructs the masked tokens. This asymmetry concentrates representational capacity in the encoder while keeping the reconstruction decoder computationally cheap, following the design philosophy of masked autoencoders~\cite{he2022masked}. Prior to encoding, spike bins are grouped into patches along the temporal dimension, compressing the representation spatially and reducing the effective sequence length seen by the encoder (analogously to the patch embedding step in vision transformers~\cite{dosovitskiy2020image}). To handle variable neuron counts across sessions and animals, NDT2 uses a session-specific linear \textit{stitcher} layer that projects each session's spike bins into a shared
 hidden dimension before the encoder and context embeddings.

NDT2 is pretrained with a masked autoencoding objective across a large collection of sessions (see Figure~\ref{fig:masking_scheme} for an illustration of the masking scheme). For adaptation to a new session, the model first undergoes calibration: the model is finetuned end-to-end on unlabeled neural activity from the evaluation session using the same masked autoencoding objective. The calibrated encoder is then finetuned on the labeled downstream decoding task.

We omit NDT2 from the main comparison table and analysis, as our evaluation revealed the model to be in a failure mode: performance consistently remains at or below the single-session linear baseline across tasks (Table~\ref{tab:neds_ndt2_results}), indicating that the pretrained representations do not provide meaningful nonlinear structure beyond what a simple linear decoder already captures. 

We conducted a hyperparameter exploration detailed in Appendix~\ref{app:hyperparameters_ndt2}.

\subsubsection{NEDS}
\label{app:model_overview_neds}

Neural Encoding and Decoding at Scale (NEDS)~\cite{zhangneural} is a large-scale semi-supervised model for jointly learning to encode neural population activity and decode behavioral variables across many sessions and animals. A key distinguishing feature of NEDS is its multimodal tokenizer: rather than processing neural activity alone, NEDS jointly tokenizes binned spike bins and behavioral variables into a single unified sequence of tokens. Neural population activity and behavioral signals are each embedded into a shared token space and concatenated into a long multimodal token sequence that is processed by a single transformer backbone. This design allows the model to learn rich associations between neural and behavioral representations within a single forward pass.

The backbone is a transformer with Rotary Position Embeddings (RoPE)~\cite{su2024roformer} to encode the temporal ordering of tokens within the multimodal sequence. On the output side, a session-specific linear \textit{stitcher} layer acts as a decoder head, projecting the transformer's latent representations back into the original neuron or behavioral variable space of each session, symmetrically to the input stitcher. 

To exploit this multimodal token sequence, NEDS applies several masking schemes during pretraining: neural tokens can be masked and reconstructed from behavioral tokens (encoding), behavioral tokens can be masked and reconstructed from neural tokens (decoding), or both modalities can be jointly masked and reconstructed from the unmasked context. This multi-objective masked modeling encourages the model to learn bidirectional mappings between neural population activity and behavior, going beyond unidirectional decoding. Like NDT models, NEDS relies on bin spikes and session-specific linear \textit{stitcher} layers on the
input and output sides to project variable neuron counts into and out of a shared hidden dimension.

As shown in Table~\ref{tab:neds_ndt2_results}, NEDS performance remains close to the single-session linear baseline and falls below it on several tasks (e.g., Wheel, Reward, Choice), indicating a failure to leverage the nonlinear representational capacity of the pretrained encoder beyond what a simple linear decoder already recovers. We therefore omit NEDS from the main
comparison table and analysis. Appendix~\ref{app:hyperparameters_neds} provides more details of the hyperparameter exploration conducted.

%% file: appendix/hyperparameters.tex
\section{Hyperparameter Selection}
\label{app:hyperparameters}

In this section, we detail the implementation and tuning configurations for the different models as they were used in the 3 task suites. We start by introducing how we tuned the single-session, single-task baselines. In all of the sweeps, we optimize hyperparameters using the Tree-structured Parzen Estimator (TPE) algorithm~\citep{bergstra2011tpe}
implemented in Optuna~\citep{akiba2019optuna}. Exact hyperparameter spaces and number of trials vary depending on the model. For a given trial, models are optimized using AdamW~\citep{loshchilov2019decoupledweightdecayregularization} with a one-cycle LR scheduler~\citep{smith2018superconvergencefasttrainingneural}. In the search spaces, the LR scheduler's division (div) factor determines the initial learning rate and is sampled from a loguniform distribution to bias towards 1. Pretrained models are also optimized with AdamW. Following common practice~\cite{he2019bag,jia2018highly}, weight decay is applied to non-embedding, non-bias parameters only; embedding, bias, and normalization layers' scale/shift parameters are excluded.

\subsection{Single-session Baselines for TS1}
\label{app:ts1-baseline-intro}

\begin{figure}[ht]
    \centering
    \includegraphics[width=0.8\linewidth]{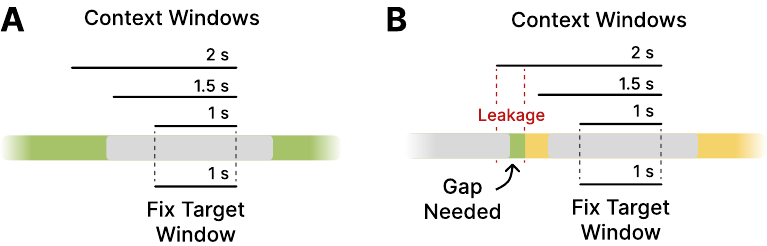}
    \caption{{\bf Context Window Selection and Data Leakage Prevention.} (A) While our benchmark uses a fixed 1s default for both input and target windows, the framework supports variable input context lengths (e.g., 1.5s, 2s). (B) When extending the context, a temporal gap is required to prevent data leakage from outside the valid trial boundaries.}
    \label{fig:context_window}
\end{figure}

We implement a set of single-session baselines for TS1 to benchmark performance without any pretraining. Each model is tuned on a single task and single session at a time in a fully supervised manner. In TS1, models must be able to map from neural activity across many units to a given decoding target. Evaluation on any task from TS1 consists of producing a prediction that conforms to the standard shape of the target: either sequence-level logits or frame-level continuous predictions. While our benchmark does enforce a specific target shape to ensure consistent and fair comparison across models, including the number of timestamps for frame-level tasks, we do not enforce a particular context window for model \textit{input} (see Figure~\ref{fig:context_window}). Nonetheless, for simplicity we use a fixed $1s$ context window for all our baselines. We encourage exploration of a more suitable context length for each task for future works and hope the community will converge on this design choice.

\newpage

\input{appendix/tuning/ts1_linear}

\newpage

\input{appendix/tuning/ts1_mlp}

\newpage

\input{appendix/tuning/ts1_tcn}

\newpage

\input{appendix/tuning/ts1_gru}

\newpage

\input{appendix/tuning/ts1_cebra}

\newpage

\input{appendix/tuning/ts1_poyo}

\newpage

\input{appendix/tuning/ts1_ndt}

\newpage

\subsection{Single-session Baselines for TS2}

\input{appendix/tuning/ts2_stats}

\newpage

\input{appendix/tuning/ts2_autoencoder}

\newpage

\input{appendix/tuning/ts2_ndt}

\newpage

\input{appendix/tuning/ts2_lfads}

\newpage

\subsection{Single-session Baselines for TS3}

\input{appendix/tuning/ts3_isi}

\newpage

\input{appendix/tuning/ts3_lolcat}

\newpage

\input{appendix/tuning/ts3_nemo}

\newpage

\subsection{Pretrained Baselines}

\begin{figure}[ht]
    \centering
    \includegraphics[width=\linewidth]{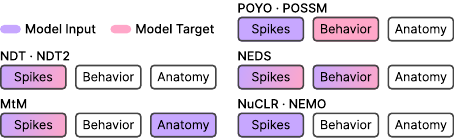}
    \caption{{\bf Input and Target Modalities for Pretrained Baselines}. We illustrate which data types (Spikes, Behavior, Anatomy) serve as model inputs (purple) or prediction targets (pink) for each baseline model during pretraining.}
    \label{fig:overview_pretrain_models}
\end{figure}

We now outline pretraining and evaluation procedures for the models summarized in Figure~\ref{fig:overview_pretrain_models}. We detail any deviations in our implementations from their respective original papers, as well as tuning procedures we followed for the pretraining and finetuning of the models, if applicable. For fairness of comparison, we scaled the pretrained models to approximately 10M parameters (excluding stitcher and embedding tables) and used bf16 precision.

\input{appendix/tuning/poyo_plus}

\input{appendix/tuning/possm}

\newpage

\begin{figure}[ht]
    \centering
    \includegraphics[width=\linewidth]{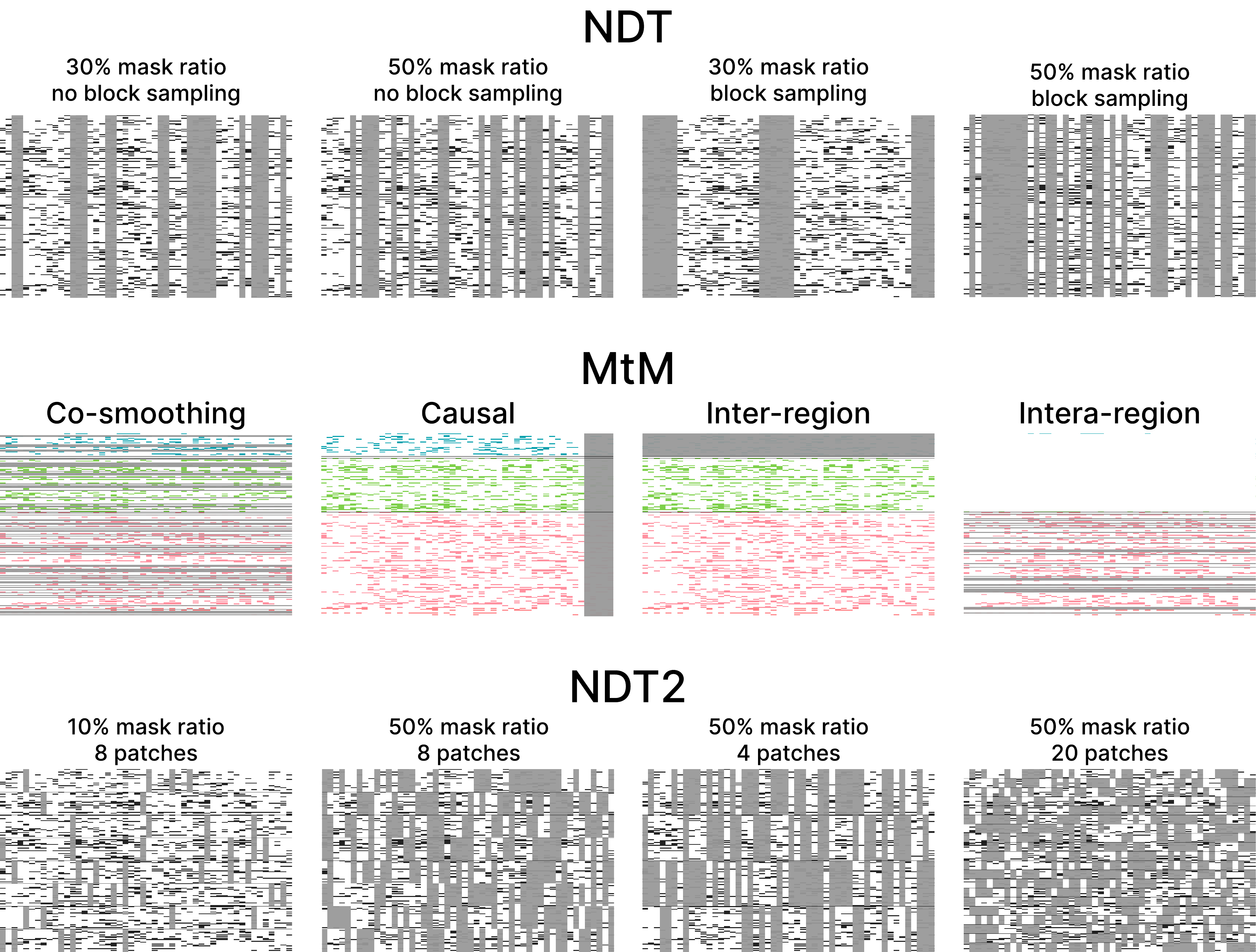}
    \caption{{\bf Visualizing Masking Strategies for Pretrained Neural Models}. Comparison of masking schemes used during pretraining for NDT, MtM, and NDT2. Gray shaded areas represent masked spikes (targets for reconstruction).}
    \label{fig:masking_scheme}
\end{figure}
\newpage

\input{appendix/tuning/ndt_stitch}
\newpage

\input{appendix/tuning/mtm}
\newpage

\input{appendix/tuning/ndt2}
\newpage

\input{appendix/tuning/neds}
\newpage


\newpage

%% file: appendix/tuning/ts1_linear.tex
\subsubsection{Linear}
\label{app:hyperparameters_linear}

The linear baseline accepts the full input context at once, bins spike trains according to a fixed bin size, and maps directly to the target readout. We detail the tuning parameters for the linear decoding baseline in Table~\ref{tab:linear_hp_space}.

\begin{table*}[h]
\caption{Hyperparameter search space for linear decoding baselines. Linear models are tuned on individual sessions and tasks for $100$ trials with $4$ concurrent trials at a time.}
\label{tab:linear_hp_space}
\begin{center}
\begin{small}
\begin{sc}
\begin{tabular}{lccr}
\toprule

Hyperparameter & Search Space & Sampling type \\

\midrule
\multicolumn{3}{l}{\scriptsize Model parameters} \\
Bin size (ms)               & $\{1,\, 5,\, 10,\, 20,\, 40\}$         & Categorical \\
\midrule
\multicolumn{3}{l}{\scriptsize Training parameters} \\
Number of epochs            & $\{100,\, 300,\, 500\}$                & Discrete (step $200$) \\
Batch size                  & $\{16,\, 32,\, 64\}$                   & Discrete ($\log_2$) \\
Weight decay                & $[10^{-6},\, 10^{-1}]$                 & Loguniform \\
Base learning rate          & $[10^{-5},\, 10^{-2}]$                 & Loguniform \\
LR scheduler -- pct start   & $[0.1,\, 0.9]$                         & Uniform \\
LR scheduler -- div factor  & $[1,\, 5]$                             & Loguniform \\

\bottomrule
\end{tabular}
\end{sc}
\end{small}
\end{center}
\vskip -0.1in
\end{table*}

%% file: appendix/tuning/ts1_mlp.tex
\subsubsection{Multi-Layer Perceptron (MLP)}
\label{app:hyperparameters_mlp}

We parameterize our MLP baseline based on network depth and the width of the first hidden layer. Successive hidden layers project inputs down by a factor of $2$, followed by a final readout layer. In initial experiments we found this was a reasonable design choice to balance performance enabled by expressivity of the search space with combinatorial complexity (e.g.,~an alternative is to allow Optuna to select a width for every layer, but the search space becomes much larger). Like the linear baseline, the MLP accepts the full context window at once and bins at a fixed bin size. We detail the tuning parameters for the MLP decoding baseline in Table~\ref{tab:mlp_hp_space}.

\begin{table*}[h]
\caption{Hyperparameter search space for MLP decoding baselines. MLP's are tuned on individual sessions and tasks for $100$ trials with $4$ concurrent trials at a time.}
\label{tab:mlp_hp_space}
\begin{center}
\begin{small}
\begin{sc}
\begin{tabular}{lccr}
\toprule

Hyperparameter & Search Space & Sampling type \\
\midrule
\multicolumn{3}{l}{\scriptsize Model parameters} \\
Bin size (ms)               & $\{5,\, 10,\, 20,\, 40\}$              & Categorical \\
Depth                       & $\{1,\, 2,\, 3\}$                      & Discrete \\
Hidden dimension            & $\{32,\, 64,\, 128,\, 256,\, 512\}$    & Discrete ($\log_2$) \\
Dropout                     & $\{0.0,\, 0.2,\, 0.4,\, 0.6\}$         & Discrete (step $0.2$) \\
Batch norm                  & $\{\text{True},\, \text{False}\}$      & Categorical \\
\midrule
\multicolumn{3}{l}{\scriptsize Training parameters} \\
Number of epochs            & $\{100,\, 300,\, 500\}$                & Discrete (step $200$) \\
Batch size                  & $\{16,\, 32,\, 64\}$                   & Discrete ($\log_2$) \\
Weight decay                & $[10^{-6},\, 10^{-1}]$                 & Loguniform \\
Base learning rate$^\dagger$ & $[10^{-5},\, 10^{-2}]$                & Loguniform \\
LR scheduler -- pct start   & $[0.1,\, 0.9]$                         & Uniform \\
LR scheduler -- div factor  & $[1,\, 5]$                             & Loguniform \\

\bottomrule
\end{tabular}
\end{sc}
\end{small}
\end{center}
{\footnotesize $^\dagger$ When batch normalization is disabled, the upper bound is lowered to $10^{-3}$ to mitigate training instability.}
\vskip -0.1in
\end{table*}

%% file: appendix/tuning/ts1_tcn.tex
\subsubsection{Convolutional Neural Network (CNN)}
\label{app:hyperparameters_tcn}

The CNN baseline is implemented on a binned raster of fixed bin size, and has a final adaptive average pooling layer before readout. The search space allows for identity activation when depth is 1, which degenerates to the rolling window linear readout baseline (see Appendix~\ref{app:model_overview_tcn}). Note, however, that to enable fair comparison with other methods that receive context in a bidirectional fashion (e.g.,~by ingesting the full context window at a time), we utilize non-causal convolutions by applying padding on both the left and right of the kernels. The padding size is deterministic based on kernel size. As mentioned in Appendix~\ref{app:model_overview_tcn}, we start with a typical implementation of the TCN~\cite{lea2016temporal} and fix the dilation factor to $1$ for simplicity. We allow various bin sizes to improve the expressivity of the CNN baseline. 

Note that CNN and GRU are typical baselines in various other fields of ML that we adapt to neural decoding, hence we treat bin size as a tunable hyperparameter to retain expressivity on par with the linear and MLP baselines. On the other hand, methods designed to work with neural data (e.g.,~CEBRA, NDT) are used with their original designs, including fixed bin sizes, to preserve their intended inductive biases. 

We detail the tuning parameters for the CNN decoding baseline in Table~\ref{tab:tcn_hp_space}.

\begin{table*}[h]
\caption{Hyperparameter search space for CNN decoding baselines. CNN's are tuned on individual sessions and tasks for $100$ trials with $4$ concurrent trials at a time.}
\label{tab:tcn_hp_space}
\begin{center}
\begin{small}
\begin{sc}
\begin{tabular}{lccr}
\toprule

Hyperparameter & Search Space & Sampling type \\
\midrule
\multicolumn{3}{l}{\scriptsize Model parameters} \\
Bin size (ms)               & $\{1,\, 5,\, 10,\, 20,\, 40\}$         & Categorical \\
Depth$^\dagger$             & $\{1,\, 2,\, 4,\, 6,\, 8,\, 10\}$      & Discrete \\
Hidden dimension            & $\{32,\, 64,\, 128,\, 256\}$           & Discrete ($\log_2$) \\
Kernel size                 & $\{3,\, 5,\, 7\}$                      & Discrete (step $2$) \\
Dropout                     & $\{0.0,\, 0.2,\, 0.4,\, 0.6\}$         & Discrete (step $0.2$) \\
Batch norm                  & $\{\text{True},\, \text{False}\}$      & Categorical \\
Activation$^\ddagger$       & $\{\text{ReLU},\, \text{Identity}\}$   & Categorical \\
\midrule
\multicolumn{3}{l}{\scriptsize Training parameters} \\
Number of epochs            & $\{100,\, 300,\, 500\}$                & Discrete (step $200$) \\
Batch size                  & $\{16,\, 32,\, 64\}$                   & Discrete ($\log_2$) \\
Weight decay                & $[10^{-6},\, 10^{-1}]$                 & Loguniform \\
Base learning rate$^\S$     & $[10^{-5},\, 10^{-2}]$                 & Loguniform \\
LR scheduler -- pct start   & $[0.1,\, 0.9]$                         & Uniform \\
LR scheduler -- div factor  & $[1,\, 5]$                             & Loguniform \\

\bottomrule
\end{tabular}
\end{sc}
\end{small}
\end{center}

{\footnotesize $^\dagger$ Sampled via an index $i \in \{0, \dots, 5\}$, mapped to depth $1$ if $i=0$ and $2i$ otherwise.

\footnotesize $^\ddagger$ Only sampled when depth $= 1$ to allow the degenerate linear case; deeper models use ReLU.

\footnotesize $^\S$ Upper bound is reduced to $10^{-3}$ for the licking rate task to mitigate training instability.}

\vskip -0.1in
\end{table*}

%% file: appendix/tuning/ts1_gru.tex
\subsubsection{Gated Recurrent Unit (GRU)}
\label{app:hyperparameters_gru}

The GRU baseline operates on a binned raster, projecting spike counts 
at each timestep into a fixed hidden dimension via a linear layer 
before processing the sequence through a multi-layer GRU. The context window is processed 
in strides determined by the bin size, rather than ingested all at 
once, reflecting the sequential nature of the recurrent architecture. 
Note that dropout is disabled within the GRU when depth is $1$, as 
it is only meaningful across stacked recurrent layers. We include 
bidirectionality as a tunable hyperparameter: when enabled, separate 
forward and backward passes are concatenated, allowing the model to 
exploit future context symmetrically with the past, on par with the 
non-causal convolutions used in the CNN baseline. Adaptive average 
pooling is applied over the temporal dimension before the final linear 
readout to support both sequence- and frame-level tasks. The upper 
bound on hidden dimension is reduced from $512$ to $256$ at a bin size 
of $1$\,ms to satisfy memory constraints, and the learning rate upper 
bound is reduced to $10^{-3}$ on the licking rate task to mitigate 
training instability.

Note that GRU and CNN are typical baselines in various other fields of 
ML that we adapt to neural decoding, hence we treat bin size as a 
tunable hyperparameter to retain expressivity on par with the linear 
and MLP baselines. On the other hand, methods designed specifically for 
neural data (e.g.,~CEBRA, NDT) are used with their original designs, 
including fixed bin sizes, to preserve their intended inductive biases.

We detail the tuning parameters for the GRU decoding baseline in 
Table~\ref{tab:gru_hp_space}.

\begin{table*}[h]
\caption{Hyperparameter search space for GRU decoding baselines. GRU's are tuned on individual sessions and tasks for $100$ trials with $4$ concurrent trials at a time.}
\label{tab:gru_hp_space}
\begin{center}
\begin{small}
\begin{sc}
\begin{tabular}{lccr}
\toprule
Hyperparameter & Search Space & Sampling type \\
\midrule
\multicolumn{3}{l}{\scriptsize Model parameters} \\
Bin size (ms)               & $\{1,\, 5,\, 10,\, 20,\, 40\}$         & Categorical \\
Depth                       & $\{1,\, 2,\, 3\}$                      & Discrete \\
Hidden dimension$^\dagger$  & $\{32,\, 64,\, 128,\, 256,\, 512\}$    & Discrete ($\log_2$) \\
Dropout                     & $\{0.0,\, 0.2,\, 0.4,\, 0.6\}$         & Discrete (step $0.2$) \\
Bidirectional               & $\{\text{True},\, \text{False}\}$      & Categorical \\
\midrule
\multicolumn{3}{l}{\scriptsize Training parameters} \\
Number of epochs            & $\{100,\, 300,\, 500\}$                & Discrete (step $200$) \\
Batch size                  & $\{16,\, 32,\, 64\}$                   & Discrete ($\log_2$) \\
Weight decay                & $[10^{-6},\, 10^{-1}]$                 & Loguniform \\
Base learning rate$^\ddagger$ & $[10^{-5},\, 10^{-2}]$               & Loguniform \\
LR scheduler -- pct start   & $[0.1,\, 0.9]$                         & Uniform \\
LR scheduler -- div factor  & $[1,\, 5]$                             & Loguniform \\

\bottomrule
\end{tabular}
\end{sc}
\end{small}
\end{center}

{\footnotesize $^\dagger$ Upper bound is reduced to $256$ when bin size $= 1$\,ms to satisfy memory constraints.

\footnotesize $^\ddagger$ Upper bound is reduced to $10^{-3}$ for the licking rate task to mitigate training instability.}

\vskip -0.1in
\end{table*}

%% file: appendix/tuning/ts1_cebra.tex
\subsubsection{CEBRA}
\label{app:hyperparameters_cebra}

CEBRA is applied in a two-step process. In the first step, the CEBRA encoder is fitted on the training split of each session using a contrastive objective, producing a low-dimensional embedding of binned neural activity. In the second step, a two-layer MLP probe, with hidden dimension equal to twice the CEBRA output dimension, a GELU activation, and dropout ($p = 0.2$), is trained on top of the frozen embeddings to predict the target behavior. The probe is optimized using the AdamW optimizer with a OneCycleLR scheduler, and early stopping is applied on the validation split.

Neural activity is binned at a fixed resolution of $20\,$ms prior to encoding and is not treated as a tunable hyperparameter following the original CEBRA paper~\citep{schneider2023cebra}. This stands in contrast to recurrent and convolutional baselines (e.g., GRU, CCN), which operate directly on spike trains at flexible temporal resolutions.

For sequence-level tasks, the CEBRA embeddings are mean-pooled across time before being passed to the probe, reducing the per-trial representation to a single vector. For frame-level tasks, the probe is applied at each time step independently.

The contrastive mode (\texttt{time}, \texttt{behavior}, or \texttt{hybrid}) is selected per task and per session during hyperparameter tuning. The number of hidden units in the MLP probe is coupled to the output dimension, fixed at $2\times$ the output dimension. 

We detail the tuning parameters for the CEBRA decoding baseline in Table~\ref{tab:cebra_hp_space}.

\begin{table*}[h]
\caption{Hyperparameter search space for CEBRA baselines. CEBRA is tuned on individual sessions and tasks for $100$ trials with $4$ concurrent trials at a time.}
\label{tab:cebra_hp_space}
\begin{center}
\begin{small}
\begin{sc}
\begin{tabular}{lccr}
\toprule

Hyperparameter & Search Space & Sampling type \\
\midrule
\multicolumn{3}{l}{\scriptsize Model parameters} \\
Output dimension            & $\{8,\, 16,\, 32,\, 64,\, 128,\, 256\}$ & Discrete ($\log_2$) \\
Time offsets                & $\{1,\, 2,\, 3,\, 4,\, 5\}$            & Discrete \\
Model learning rate         & $[10^{-4},\, 10^{-3}]$                 & Loguniform \\
Temperature                 & $[0.5,\, 2.0]$                         & Discrete (step $0.1$) \\
Mode                        & $\{\text{time},\, \text{behavior}\}$   & Categorical \\
Temperature mode            & $\{\text{constant}\}$                  & Categorical \\
Max iterations              & $\{5000,\, 7000,\, 9000\}$             & Discrete (step $2000$) \\
\midrule
\multicolumn{3}{l}{\scriptsize Training parameters} \\
Number of epochs            & $\{100,\, 200,\, 300,\, 400,\, 500\}$  & Discrete (step $100$) \\
Batch size                  & $\{16,\, 32,\, 64\}$                   & Discrete ($\log_2$) \\
Weight decay                & $[10^{-6},\, 10^{-1}]$                 & Loguniform \\
Base learning rate          & $[10^{-5},\, 10^{-2}]$                 & Loguniform \\
LR scheduler -- pct start   & $[0.1,\, 0.9]$                         & Uniform \\
LR scheduler -- div factor  & $[1,\, 5]$                             & Loguniform \\

\bottomrule
\end{tabular}
\end{sc}
\end{small}
\end{center}

{\footnotesize Note: Number of hidden units is set to $2 \times$ output dimension.}

\vskip -0.1in
\end{table*}

%% file: appendix/tuning/ts1_poyo.tex
\subsubsection{POYO}
\label{app:hyperparameters_poyo}  

POYO uses spike tokenization (see Appendix~\ref{app:tokenization_spike}), where each spike is represented as a tuple $(u_i, t_i)$ of unit identity and timestamp, making the model independent of any fixed bin size. The architecture compresses the input spike sequence into a fixed latent grid via cross-attention, processes it through several self-attention layers, 
and decodes predictions via a final cross-attention decoder. For sequence-level tasks, a single query token placed at the midpoint of the context window is used; for frame-level tasks, one query token is placed at each target timestamp. In both cases, the query tokens are constructed 
from a learnable session embedding combined with rotary position embeddings 
encoding the output timestamps.

Due to the memory overhead of Transformer models, POYO is tuned with 2 concurrent trials rather than the 4 used for lighter baselines. To ensure the number of Bayesian updates remains proportionally equivalent, the total trial budget is reduced to 50 accordingly, maintaining comparable hyperparameter optimization efficiency across all models.

We detail the tuning parameters for the POYO single-session baseline in Table~\ref{tab:poyo_hp_space}.

\begin{table*}[h]
\caption{Hyperparameter search space for POYO decoding baselines. POYO models are tuned on individual sessions and tasks for $50$ trials with $2$ concurrent trials at a time.}
\label{tab:poyo_hp_space}
\begin{center}
\begin{small}
\begin{sc}
\resizebox{\columnwidth}{!}{
\begin{tabular}{lcc}
\toprule
Hyperparameter & Search Space & Sampling type \\
\midrule

\multicolumn{3}{l}{\scriptsize Model parameters} \\
Latent step                 & $\{0.0625,\, 0.125\}$                  & Discrete (step $0.0625$) \\
Num. latents per step       & $\{8,\, 16,\, 32\}$                    & Discrete ($\log_2$; sampled as $2^k$, $k\in\{3,4,5\}$) \\
Depth                       & $\{2,\, 4,\, 6,\, 8,\, 10\}$           & Discrete (step $2$) \\
Dropout$^\dagger$           & $\{0.0,\, 0.2,\, 0.4,\, 0.6\}$         & Discrete (step $0.2$) \\
Dimension                   & $\{32,\, 64,\, 128,\, 256\}$           & Discrete ($\log_2$; sampled as $2^k$, $k\in\{5,6,7,8\}$) \\
Dim. head$^\ddagger$        & $\{32,\, 64\}$                         & Conditional categorical \\
Num. attention heads$^\S$   & $\{1,\, 2,\, 4,\, 8\}$                 & Derived as dimension / dim. head \\
\midrule

\multicolumn{3}{l}{\scriptsize Training parameters} \\
Number of epochs            & $\{100,\, 300,\, 500\}$                & Discrete (step $200$) \\
Batch size                  & $\{16,\, 32,\, 64\}$                   & Discrete ($\log_2$; sampled as $2^k$, $k\in\{4,5,6\}$) \\
Weight decay                & $[10^{-6},\, 10^{-1}]$                 & Loguniform \\
Base learning rate          & $[10^{-5},\, 10^{-2}]$                 & Loguniform \\
LR scheduler -- pct start   & $[0.1,\, 0.9]$                         & Uniform \\
LR scheduler -- div factor  & $[1,\, 5]$                             & Loguniform \\
\midrule

\multicolumn{3}{l}{\scriptsize Unit dropout parameters} \\
Min. units                  & $\{100,\, 200,\, 300,\, 400\}$         & Discrete (step $100$) \\
Max. units                  & $\{600,\, 800,\, 1000,\, 1200\}$       & Discrete (step $200$) \\
Mode units$^\P$             & $\left\lfloor \mathrm{min} + \frac{\mathrm{max}-\mathrm{min}}{3} \right\rfloor$ & Derived \\

\bottomrule
\end{tabular}
}
\end{sc}
\end{small}
\end{center}

\vspace{-0.5em}
\footnotesize{
$^\dagger$ The same sampled dropout value is applied to feed-forward, linear, and attention dropout: 
\texttt{ffn\_dropout}, \texttt{lin\_dropout}, and \texttt{atn\_dropout}. \\
$^\ddagger$ \texttt{dim\_head} is sampled only when dimension $>32$; otherwise it is set equal to the model dimension. \\
$^\S$ The same derived number of heads is used for both cross-attention and self-attention:
$\texttt{cross\_heads}=\texttt{self\_heads}=\texttt{dim}/\texttt{dim\_head}$. \\
$^\P$ The unit-dropout mode is not sampled directly; it is derived from the sampled minimum and maximum unit counts.
}
\vskip -0.1in
\end{table*}



\vskip -0.1in

%% file: appendix/tuning/ts1_ndt.tex
\subsubsection{NDT Supervised}
\label{app:hyperparameters_ndt_ts1}

NDT supervised adapts the Neural Data Transformer~\cite{ye2021representation} to a fully supervised decoding setting. Each timestep is tokenized as a population vector of binned spike counts at a fixed bin size of 20\,ms matching the continuous behavior sampling rate of 50\,Hz (following the same temporal tokenization scheme as the CNN and GRU; see Appendix~\ref{app:tokenization}). This stands in contrast to POYO, which tokenizes individual neuron spikes as events. NDT instead operates on binned population snapshots, making the bin size an architectural choice rather than a tunable preprocessing step. Each population token is projected into a fixed-dimensional embedding via a linear layer and a learnable positional embeddings are added before the sequence is processed by a standard transformer encoder with pre-norm layers. For sequence-level tasks, the output is mean-pooled over the temporal dimension before the final linear readout, while frame-level tasks read out at every timestep as the bin size match the sampling rate.

A single unified dropout rate is applied across the feed-forward, attention, and pre- and post-encoder layers, following the ablation study in the original NDT paper~\cite{ye2021representation}. Although the dataset and tasks differ from the original work, we retain this design choice to stay close to the validated configuration. An optional T-Fixup initialization scheme~\cite{pmlr-v119-huang20f} is included in the search space, which rescales attention value weights and linear layers to stabilize training without a warm-up schedule.

Due to the memory overhead of Transformer models, NDT supervised is tuned with $2$ concurrent trials rather than the $4$ used for lighter baselines. To ensure the number of Bayesian updates is proportionally equivalent, the total trial budget is reduced accordingly, maintaining comparable hyperparameter optimization efficiency across all models.

We detail the tuning parameters for the NDT supervised baseline in 
Table~\ref{tab:ndt_superv_hp_space}.

\begin{table*}[h]
\caption{Hyperparameter search space for NDT supervised decoding baselines. NDT supervised's are tuned on individual sessions and tasks for $50$ trials with $2$ concurrent trials at a time.}
\label{tab:ndt_superv_hp_space}
\begin{center}
\begin{small}
\begin{sc}
\begin{tabular}{lccr}
\toprule
Hyperparameter & Search Space & Sampling type \\
\midrule
\multicolumn{3}{l}{\scriptsize Model parameters} \\
Dimension                   & $\{32,\, 64,\, 128,\, 256\}$           & Discrete ($\log_2$) \\
Depth                       & $\{2,\, 4,\, 6,\, 8,\, 10\}$           & Discrete (step $2$) \\
Num. heads                  & $\{1,\, 2,\, 4\}$                      & Discrete ($\log_2$) \\
FFN factor       & $\{1,\, 2,\, 4\}$                      & Discrete ($\log_2$) \\
Dropout           & $\{0.2,\, 0.4,\, 0.6\}$                & Discrete (step $0.2$) \\
T-Fixup init           & $\{\text{True},\, \text{False}\}$      & Categorical \\
\midrule
\multicolumn{3}{l}{\scriptsize Training parameters} \\
Number of epochs            & $\{100,\, 300,\, 500\}$                & Discrete (step $200$) \\
Batch size                  & $\{16,\, 32,\, 64\}$                   & Discrete ($\log_2$) \\
Weight decay                & $[10^{-6},\, 10^{-1}]$                 & Loguniform \\
Base learning rate          & $[10^{-5},\, 5 \cdot 10^{-2}]$         & Loguniform \\
LR scheduler -- pct start   & $[0.1,\, 0.9]$                         & Uniform \\
LR scheduler -- div factor  & $[1,\, 5]$                             & Loguniform \\
\bottomrule
\end{tabular}
\end{sc}
\end{small}
\end{center}
{
\footnotesize Spikes are binned at $20$ ms and read in through a linear projection; neither is searched. No unit dropout is applied, as the model is single-session with a fixed unit set.}
\vskip -0.1in
\end{table*}

%% file: appendix/tuning/ts2_stats.tex
\subsubsection{Statistical Baselines}
\label{app:hyperparameters_stat_baseline}

The statistical baselines have no training parameters: each is fit in closed form on the training windows, and its one or two hyperparameters are selected by exhaustive grid search over the values in Table~\ref{tab:hp_space_stat_baseline}, scored on validation with the same Poisson metric used at test. Bin size is not tuned: all methods use the TS2 protocol's $20$~ms bins, giving a $50$-bin context window with a $5$-bin forecast horizon and $10\%$ of units held out for co-smoothing. The reduced-rank readouts select on an independent validation hold-out draw and are then refit for the test draw, so the scored units take no part in selection. The reduced rank is additionally capped at $\min(|O|, |H|)$, the number of observed and held-out units on the session.

\begin{table*}[h]
\caption{Hyperparameter search space for the statistical baselines. Each method is fit per session and task by exhaustive grid search on validation, with no gradient-based training.}
\label{tab:hp_space_stat_baseline}
\begin{center}
\begin{small}
\begin{sc}
\begin{tabular}{lccr}
\toprule

Hyperparameter & Search Space & Sampling type \\

\midrule
\multicolumn{2}{l}{\scriptsize Population coupling} \\

Smoothing width $\sigma$ (bins) & $\{1,\, 1.5,\, 2,\, 2.5\}$ & Categorical \\
Coupling exponent $\gamma$ & $\{0.5,\, 0.75,\, 1,\, 1.25,\, 1.5,\, 2\}$ & Categorical \\

\midrule
\multicolumn{2}{l}{\scriptsize RRR and RRR (w/ ISI features)} \\

Rank $k$ & $\{1,\, 2,\, 3,\, 4,\, 5,\, 8,\, 12,\, 16\}^\dagger$ & Categorical \\
Ridge strength $\lambda$ & $\{10^{0}, \dots, 10^{7}\}$ & Categorical ($\log_{10}$) \\

\midrule
\multicolumn{2}{l}{\scriptsize Trailing mean} \\

Window length $K$ (bins) & $\{1,\, 2,\, 3,\, 5,\, 8,\, 13,\, 21,\, 34,\, 45\}$ & Categorical \\

\midrule
\multicolumn{2}{l}{\scriptsize Shrinkage} \\

Blend weight $\alpha$ & $[0, 1]$ & Discrete (step $0.1$) \\

\midrule
\multicolumn{2}{l}{\scriptsize Ridge autoregression} \\

Window length $L$ (bins) & $\{3,\, 5,\, 10,\, 20,\, 45\}$ & Categorical \\
Ridge strength $\lambda$ & $\{10^{-1}, \dots, 10^{4}\}$ & Categorical ($\log_{10}$) \\

\bottomrule
\end{tabular}
\end{sc}
\end{small}
\end{center}
\vskip -0.1in
\end{table*}

{\footnotesize $^\dagger$ Capped at $\min(|O|, |H|)$, the number of observed and held-out units on the session.}

%% file: appendix/tuning/ts2_autoencoder.tex
\subsubsection{Autoencoder (AE)}
\label{app:hyperparameters_ae}

We parameterize the AE based on encoder depth, decoder depth, and the width of the first hidden layer. Successive encoder layers project the representation down by a factor of 2, while the decoder mirrors this by projecting back up by a factor of 2 at each layer, followed by a final linear readout to the full output dimensionality. As with the MLP, this symmetric bottleneck design balances expressivity and search space complexity. The AE is trained with a full reconstruction objective over all neurons and timesteps, encouraging the bottleneck to capture low-dimensional population latents, and is subsequently evaluated on co-smoothing and forecasting tasks. 

We detail the tuning parameters for the AE in Table~\ref{tab:hp_space_ae}.

\begin{table*}[h]
\caption{Hyperparameter search space for Autoencoder baselines. Autoencoders are tuned on individual sessions and tasks for $100$ trials with $4$ concurrent trials at a time.}
\label{tab:hp_space_ae}
\begin{center}
\begin{small}
\begin{sc}
\begin{tabular}{lccr}
\toprule

Hyperparameter & Search Space & Sampling type \\

\midrule
\multicolumn{2}{l}{\scriptsize Model parameters} \\

Bin size (ms) & $\{1, 5, 10, 20, 40\}$ & Categorical  \\

\midrule
\multicolumn{2}{l}{\scriptsize Training parameters} \\

Number of epochs & $\{100,\, 300,\, 500\}$  & Discrete (STEP 200) \\
Batch size & $[16, 128]$ & Discrete ($\log_2$) \\
Weight decay & $[10^{-6},10^{-1}]$ & Loguniform \\
Base learning rate & $[10^{-5}, 10^{-2}]$ & Loguniform \\
Lr scheduler - pct start & $[0.1, 0.9]$ & Uniform \\
Lr scheduler - div factor & $[1, 5]$ & Loguniform \\

\bottomrule
\end{tabular}
\end{sc}
\end{small}
\end{center}
\vskip -0.1in
\end{table*}

{\footnotesize $^\dagger$ Applied identically to encoder, pre-encoder, and post-encoder layers.}

%% file: appendix/tuning/ts2_ndt.tex
\subsubsection{NDT}
\label{app:hyperparameters_ndt_ts2}

The NDT baseline for TS2 shares the same Transformer encoder backbone 
described in Appendix~\ref{app:hyperparameters_ndt_ts1}, but departs 
from the supervised variant in both its tokenization and training 
objective. Rather than a linear projection from neuron counts to a 
fixed hidden dimension, the hidden dimension here is defined as 
$N \times \max(\texttt{unit\_emb\_dim}, 1)$, where $N$ is the number 
of units in the session. When \texttt{unit\_emb\_dim} $= 0$, spike 
counts are passed directly as raw floats. 
When \texttt{unit\_emb\_dim} $> 0$, a learned count embedding of that 
dimension is applied per unit and the resulting vectors are 
concatenated across neurons before being passed to the encoder. The 
number of attention heads is constrained to divide the hidden dimension 
evenly, and is only tuned when \texttt{unit\_emb\_dim} $= 2$, 
otherwise it is fixed to $1$. The bin size is fixed to 20\,ms matching the continuous behavior sampling rate.

The model is trained with a masked modelling objective in the spirit 
of BERT~\cite{devlin2019bert}, targeting full reconstruction of masked 
population activity under a Poisson negative log-likelihood loss. 
Masking is applied along the temporal dimension across all neurons 
simultaneously. The search space covers the overall mask ratio, the 
maximum contiguous block size, and the probability of block-structured 
versus random masking.

Dropout is sampled from a continuous range starting at $0.2$ and 
applied uniformly to the pre-encoder, post-encoder, and 
within-encoder (attention and feed-forward) layers, following the 
ablation study in the original NDT paper~\cite{ye2021representation}. 
Due to the memory overhead of transformer models, NDT is tuned with 
$2$ concurrent trials; the total trial budget is set to $50$ to 
maintain a number of Bayesian updates comparable to other baselines, 
as discussed in Appendix~\ref{app:hyperparameters_ndt_ts1}.

We detail the tuning parameters for the NDT baseline in 
Table~\ref{tab:ndt_ts2_hp_space}.

\begin{table*}[h]
\caption{Hyperparameter search space for NDT baselines. NDT is tuned on individual sessions for $50$ trials with $2$ concurrent trials at a time.}
\label{tab:ndt_ts2_hp_space}
\begin{center}
\begin{small}
\begin{sc}
\begin{tabular}{lccr}
\toprule
Hyperparameter & Search Space & Sampling type \\
\midrule
\multicolumn{3}{l}{\scriptsize Model parameters} \\
Unit emb. dim                   & $\{0,\, 1,\, 2\}$                       & Discrete \\
Num. attention heads$^\ddagger$ & $\{1,\, 2\}$                            & Discrete ($\log_2$) \\
FFN factor                      & $\{1,\, 2\}$                            & Discrete ($\log_2$) \\
Num. encoder layers             & $\{1,\, 2,\, 3,\, 4,\, 5,\, 6\}$        & Discrete \\
Dropout$^\dagger$               & $[0.2,\, 0.6]$                          & Uniform \\
Custom init                     & $\{\textsc{True},\, \textsc{False}\}$   & Categorical \\
\midrule
\multicolumn{3}{l}{\scriptsize Masking parameters} \\
Mask ratio                      & $[0.5,\, 0.9]$                          & Uniform \\
Max block size                  & $\{1,\, 2,\, \ldots,\, 7\}$             & Discrete \\
Block mask prob.                & $[0.0,\, 1.0]$                          & Uniform \\
\midrule
\multicolumn{3}{l}{\scriptsize Training parameters} \\
Number of epochs                & $\{100,\, 300,\, 500\}$                 & Discrete (step $200$) \\
Batch size                      & $\{16,\, 32,\, 64\}$                    & Discrete ($\log_2$) \\
Weight decay                    & $[10^{-6},\, 10^{-1}]$                  & Loguniform \\
Base learning rate              & $[10^{-5},\, 5\times 10^{-2}]$          & Loguniform \\
LR scheduler -- pct start       & $[0.1,\, 0.9]$                          & Uniform \\
LR scheduler -- div factor      & $[1,\, 5]$                              & Loguniform \\
\bottomrule
\end{tabular}
\end{sc}
\end{small}
\end{center}
{

\footnotesize $^\dagger$ Applied identically to pre-encoder, post-encoder, and within-encoder (attention and feed-forward) dropout.

\footnotesize $^\ddagger$ Only sampled when unit emb. dim $= 2$; otherwise set to $1$ (since the model dimension is $N \cdot \max(\text{unit emb. dim},\,1)$ and must be divisible by the number of heads).}
\vskip -0.1in
\end{table*}

%% file: appendix/tuning/ts2_lfads.tex
\subsubsection{Latent Factor Analysis via Dynamical Systems (LFADS)}
\label{app:hyperparameters_lfads}

Following AutoLFADS, the regularization strengths are searched alongside the architecture rather than fixed. The encoder, controller, and initial-condition dimensions are held at $64$, and bin size is not searched: LFADS uses the TS2 protocol's $20$~ms bins. The corruption parameters are task-gated, since each is only read by one task's forward pass: co-smoothing searches the coordinated dropout rates, forecasting searches the reconstruction gradient scope. We detail the tuning parameters in Table~\ref{tab:hp_space_lfads}.

\begin{table*}[h]
\caption{Hyperparameter search space for LFADS baselines. LFADS is tuned on individual sessions and tasks for $50$ trials with $4$ concurrent trials at a time.}
\label{tab:hp_space_lfads}
\begin{center}
\begin{small}
\begin{sc}
\begin{tabular}{lccr}
\toprule

Hyperparameter & Search Space & Sampling type \\

\midrule
\multicolumn{2}{l}{\scriptsize Model parameters} \\

Generator dimension & $\{100,\, 200,\, 400\}$ & Categorical \\
Factor dimension & $\{20,\, 40,\, 80\}$ & Categorical \\
Controller output dimension & $\{1,\, 2,\, 4,\, 8\}$ & Categorical \\
Dropout & $[0, 0.6]$ & Discrete (step $0.05$) \\
Coordinated dropout rate$^\dagger$ & $[0, 0.7]$ & Discrete (step $0.05$) \\
Unit dropout rate$^\dagger$ & $[0, 0.3]$ & Discrete (step $0.05$) \\
Forecast gradient scope$^\ddagger$ & $\{\textrm{tail},\, \textrm{window}\}$ & Categorical \\

\midrule
\multicolumn{2}{l}{\scriptsize Training parameters} \\

KL scale - initial condition & $[10^{-3}, 10^{1}]$ & Loguniform \\
KL scale - controller & $[10^{-3}, 10^{1}]$ & Loguniform \\
L2 scale - generator & $[10^{-4}, 10^{2}]$ & Loguniform \\
L2 scale - controller & $[10^{-4}, 10^{2}]$ & Loguniform \\
Number of epochs & $\{100,\, 300,\, 500\}$ & Discrete (STEP 200) \\
Batch size & $[16, 128]$ & Discrete ($\log_2$) \\
Weight decay & $[10^{-6},10^{-1}]$ & Loguniform \\
Base learning rate & $[10^{-5}, 10^{-2}]$ & Loguniform \\
Lr scheduler - pct start & $[0.1, 0.9]$ & Uniform \\
Lr scheduler - div factor & $[1, 5]$ & Loguniform \\

\bottomrule
\end{tabular}
\end{sc}
\end{small}
\end{center}
\vskip -0.1in
\end{table*}

{\footnotesize $^\dagger$ Searched on co-smoothing only. $^\ddagger$ Searched on forecasting only.}

%% file: appendix/tuning/ts3_isi.tex
\subsubsection{Inter-Spike Interval (ISI) }
\label{app:hyperparameters_isi}

We implemented the inter-spike interval (ISI) baseline as a two-stage pipeline: a feature extractor that computes a fixed length ISI histogram for each unit, followed by a classification probe (logistic regression or MLP) trained on those features to predict unit-level brain regions. To compute each unit's ISI histogram, we took the unit's full spike train over the entire recording session and calculated pairwise consecutive differences to obtain the ISIs. These ISIs were then binned into a 128-bin log-spaced histogram over the range $[10^{-3}, 3]$ seconds and L1-normalized to sum to one, yielding a fixed-length ISI feature vector for each unit.

We experimented with both linearly spaced and log-spaced bins for the ISI histogram. Using linearly spaced bins over the range $[0, 3]$ seconds resulted in a heavily right skewed histogram with nearly all of the probability mass being centered in the leftmost few bins. For example, with linear binning, only 2 bins covered the 0-50 ms range, whereas log-spaced binning allocated 63 bins to the same range, providing significantly finer temporal resolution at short intervals. Note that for the log-spaced binning approach, the lower bound of the range was $10^{-3}$ seconds, as the logarithm of 0 is undefined. Log-spaced binning yielded substantial improvement over linear binning in downstream brain region classification, with macro F1 improvement under the linear probe (single-unit: $0.215 \to 0.262$, multi-unit: $0.296 \to 0.354$) and MLP probe (single-unit: $0.288 \to 0.385$, multi-unit: $0.431 \to 0.544$).

We evaluated the ISI features using two classification probes described in~\ref{app:ts3}: logistic regression and an MLP probe. Since the ISI features are deterministic (unlike learned embeddings), we ran the linear probe once and estimated MLP variance across 5 additional seeds (43-47) using the best hyperparameters from a seed-42 MLP tuning run.

\begin{figure}[ht]
    \centering
    \includegraphics[width=\linewidth]{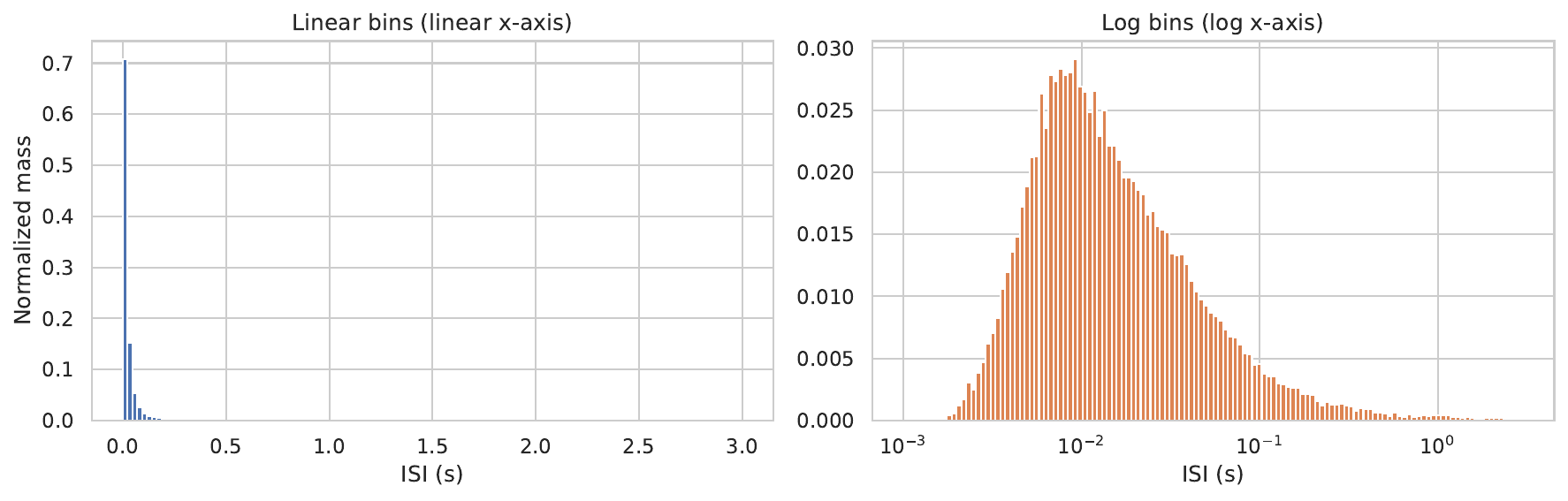}
    \caption{ISI histogram representations under linear and log-spaced binning}
    \label{fig:isi_binning_comparison}
\end{figure}

%% file: appendix/tuning/ts3_lolcat.tex
\subsubsection{LOLCAT}
\label{app:hyperparameters_lolcat}

We implemented the LOLCAT baseline as per the description in Appendix~\ref{app:model_overview_lolcat}, with various implementation-specific modifications. The original LOLCAT paper divided each recording session into 3-second windows aligned with each stimulus trial (drifting grating presentations), however \brainwidebench~does not have the same task structure. Hence, instead of using fixed-length 3-second trials, we used the task-aligned trial windows described in Figure~\ref{fig:task_intervals_ts1} to capture the behaviorally-relevant windows, rather than arbitrarily dividing the recording session into contiguous 3-second snippets. Task-aligned trial windows are variable in length, and the number of trials per recording session varies between sessions.

As per the original LOLCAT paper, we computed an ISI histogram for each task-aligned trial window, instead of a singular ISI histogram spanning the entire recording session as we had done in the ISI baseline model (see Appendix~\ref{app:hyperparameters_isi}). This produced a variable-length set of per-trial representations per unit, which were then aggregated by the multi-head attention pooling module into a single per-unit embedding. As described in~\cite{schneider2023transcriptomic}, computing per-trial histograms allows the multi-head attention module to selectively attend to the most informative trials, capturing neuron identity information at both local and global timescales. Following the same motivation as in the ISI model baseline (see Appendix~\ref{app:hyperparameters_isi}), we used log-spaced binning for the computation of the ISI histograms.

We carved out a validation set from the pretrain regime's recording sessions by using a subject-level split strategy. Recordings were grouped by subject, subjects were sorted, and the last 20\% of subjects were assigned to the validation set. This subject-level split ensured that no recordings from the same subject appeared in both the training and validation sets, emulating the zero-shot conditions of the evaluation (held-out) regime.

The original architecture in the LOLCAT paper used an MLP classifier head with a single hidden layer. However, after experimenting with both linear classifiers and MLP classifiers with a hidden layer dimension of 32 and 64, we found that the linear classifier outperformed both, and thus chose to use a linear classifier head instead.

The adaptive sampler that the original LOLCAT paper used, described in Appendix~\ref{app:model_overview_lolcat}, initialized the sampling factors for each class to balance the classes (majority class count / class count). However, we found that initializing the sampling factors for each class to a uniform value of 5, rather than the class-balancing initialization, improved performance on downstream brain region classification. 
Additionally, we increased the multiplicative step by which the sampler adjusts each class's sampling factor at every validation epoch, from $0.99$ and $1.01$ to $0.96$ and $1.04$. 
With the smaller step the factors could spread by at most $1.9\times$ over a run, far short of the $19\times$ class imbalance of \brainwidebench, leaving the sampler unable to correct the imbalance it was designed for. 
The larger step reaches $10$-$13\times$.

The original LOLCAT model had trial dropout (see Appendix~\ref{app:model_overview_lolcat}), but we found that it hurt brain region classification performance, hence we omitted the trial dropout from our final implementation. We experimented with setting the trial dropout probability as inversely proportional to the firing rate, as well as inversely proportional to the spike count of each trial, but both settings independently lowered performance. 

Unlike the ISI baseline, no hyperparameter sweep was performed for LOLCAT; all hyperparameters were taken directly from the original LOLCAT paper~\cite{schneider2023transcriptomic}, with the exception of the implementation-specific modifications mentioned above. 

We detail LOLCAT training parameters in Table~\ref{tab:lolcat_config}.

\begin{table*}[ht]
\caption{LOLCAT training configuration.}
\label{tab:lolcat_config}
\begin{center}
\begin{small}
\begin{sc}
\begin{tabular}{lr}
\toprule
Hyperparameter & Value \\
\midrule
\multicolumn{2}{l}{\scriptsize Model parameters} \\
Number of ISI bins                  & $128$ \\
ISI range (s)                       & $[10^{-3}, 3]$ \\
Encoder hidden dims                 & $[128, 64, 64, 32]$ \\
Encoder activation                  & \texttt{ReLU} \\
Encoder dropout                     & $0.5$ \\
Encoder batchnorm                   & \checkmark \\
Attention heads                     & $4$ \\
Attention hidden dim                & $16$ \\
\midrule
\multicolumn{2}{l}{\scriptsize Training parameters} \\
Number of epochs                    & $200$ \\
Batch size                          & $64$ \\
Weight decay                        & $10^{-5}$ \\
Base learning rate                  & $10^{-3}$ \\
LR scheduler -- LR decay            & $10^{-2}$ \\
Early stopping patience             & $10$ \\
Validation every $n$ epochs         & $5$ \\
Val fraction                        & $0.2$ \\
Number of seeds                     & $5$ \\
\midrule
\multicolumn{2}{l}{\scriptsize Sampler parameters} \\
Initial sampling factor             & $5$ \\
Sampling factor bounds              & $[0.8, 100]$ \\
Grow divisor                        & $0.96$ \\
Shrink divisor                      & $1.04$ \\
Overfitting threshold               & $1\sigma$ \\
Undertraining threshold             & $0.1$ \\

\bottomrule
\end{tabular}
\end{sc}
\end{small}
\end{center}
\vskip -0.1in
\end{table*}


%% file: appendix/tuning/ts3_nemo.tex
\subsubsection{NEMO}
\label{app:hyperparameters_nemo}

We used the NeuroPyxels Python library~\cite{beau2021neuropyxels} to compute ACGs over a 2-second window (-1 second to 1 second) with 1 ms bins, using a 250 ms boxcar filter as in the original paper to estimate instantaneous firing rate. Each unit's spike train is divided into 10 quantile-based firing rate deciles~\cite{beau2025deep}. A separate ACG is computed for each decile and log-compressed to 201 bins (100 log-spaced positive lags mirrored symmetrically around zero), yielding a $10 \times 201$ representation per unit.

Following the original NEMO paper, each unit's waveform template was obtained by averaging approximately 500 recorded waveforms. The peak-amplitude channel was then selected from the precomputed template, and the waveform was trough-aligned to a fixed 90-sample window (42 samples before the trough, 48 after), with zero-padding applied where the trough fell near the recording edge. Each waveform was subsequently amplitude-normalized to $[-1, 1]$ by dividing by its maximum absolute value. The resulting waveforms and ACGs were stored together in a NEMO cache, from which data loaders drew samples during training. The same ACG augmentations from the original paper were retained. For waveforms, we additionally included amplitude jitter (rescaling amplitude by a uniform factor in $[0.9, 1.1]$, $p=0.4$) alongside the Gaussian noise augmentation.

Rather than using the encoder architectures described in Section~\ref{app:model_overview_nemo}, we scaled up both encoders to a total of approximately 10M parameters to match the scale of other models in this task suite. While scaling up, we maintained the original ratio of approximately 1.5 between the waveform and ACG encoder sizes. Each encoder was built from residual blocks consisting of two 3$\times$3 convolutional layers with batch normalization, GeLU activations~\cite{hendrycks2016gelu}, and dropout, with a skip connection that used a 1$\times$1 convolution to match dimension when channel size or stride changed. The ACG and waveform encoder used 2D residual blocks and 1D residual blocks, respectively, to match the dimensionality of the ACG and waveform features.

We designed the waveform encoder as a 1D ResNet-34 style network~\cite{he2016deep} with a 7-sample stem convolution, followed by four stages of residual blocks with [2, 3, 3, 3] blocks. The channel widths progressed as [c, 2c, 4c, 8c], and the encoder utilized global average pooling, as well as a linear projection to the representation dimension. Similarly, we constructed the ACG encoder as a 2D ResNet-18 style network with an initial stem convolution of kernel size (3,7), followed by four stages with [2, 2, 2, 3] residual blocks and the same channel progression, with global average pooling and similar linear projection to the representation dimension.

Before finalizing the ResNet-18/ResNet-34 style encoders, we experimented with various architectures. We initially designed a simple ResNet with uniform channel widths of 256 and 128 for the waveform and ACG encoders respectively, and two residual blocks per encoder. Using the original NEMO embedding dimensions of 300 and 200, this yielded only around 1.5M parameters (868K for waveform, 618K for ACG). We then scaled these encoders up by doubling the uniform channel widths to 512 and 256, respectively, and adding a third residual block to each encoder, while scaling the embedding dimensions to 1200 and 800, reaching roughly 9.1M parameters (5.3M for waveform, 3.8M for ACG). However, we ultimately finalized the architecture as described above, since they follow established design principles of ResNet-18 and ResNet-34, such as progressive channel widening and structured block counts, rather than arbitrarily scaling channel widths or embedding dimensions.

We used early stopping based on the macro F1 score of a leave-one-session-out linear probe for brain region classification on the training set, with a patience of 250 epochs (5 validation cycles of 50 epochs each.

We detail NEMO training parameters in Table~\ref{tab:nemo_config}.

\begin{table*}[h]
\caption{NEMO training configuration.}
\label{tab:nemo_config}
\begin{center}
\begin{small}
\begin{sc}
\begin{tabular}{lr}
\toprule
Hyperparameter & Value \\
\midrule
\multicolumn{2}{l}{\scriptsize Model parameters} \\
WVF base channels                   & $64$ \\
ACG base channels                   & $32$ \\
WVF residual blocks                 & $[2, 3, 3, 3]$ \\
ACG residual blocks                 & $[2, 2, 2, 3]$ \\
WVF representation dim              & $512$ \\
ACG representation dim              & $256$ \\
Projection (embedding) dim          & $512$ \\
WVF dropout                         & $0.1$ \\
ACG dropout                         & $0.2$ \\
\midrule
\multicolumn{2}{l}{\scriptsize Training parameters} \\
Number of epochs                    & $100$ \\
Batch size                          & $1024$ \\
Weight decay                        & $10^{-4}$ \\
Base learning rate                  & $5 \times 10^{-4}$ \\
LR scheduler -- LR decay            & $10^{-2}$ \\
Contrastive loss temperature        & $0.5$ \\
Early stopping patience             & $5$ \\
Validation every $n$ epochs         & $10$ \\
Number of seeds                     & $5$ \\
\bottomrule
\end{tabular}
\end{sc}
\end{small}
\end{center}
\vskip -0.1in
\end{table*}

%% file: appendix/tuning/poyo_plus.tex
\subsubsection{POYO+}
\label{app:hyperparameters_poyo+}

\textbf{Multi-task pretraining and task weighting.}
We pretrained POYO+ with a weighted multi-task objective,
\begin{equation}
\mathcal{L}_{\mathrm{total}}=\sum_{t}\lambda_t\mathcal{L}_t.
\end{equation}
Pretraining was conducted in an 8-task setting with the POYO+ 10M architecture for 100 epochs, using batch size $1024$ and base learning rate $5\times10^{-5}$ on the nonfiltered processed dataset. To choose the task weights, we first ran a 10-epoch pilot training run with equal weights, i.e., $\lambda_t=1$ for all tasks. We then grouped tasks according to their observed overfitting behavior in the pilot run. In particular, we monitored whether the training loss continued to decrease while the validation loss started to increase. Tasks that showed stronger early overfitting were assigned smaller weights. For example, the choice task began to overfit around the third epoch, so we down-weighted it substantially to $\lambda_{\mathrm{choice}}=0.01$. In contrast, the regression tasks did not show clear overfitting within the 10-epoch pilot run, and their validation losses decreased at similar rates; therefore, we assigned them the same weight of $1.0$. 

The final task weights are the following:

\begin{alignat*}{2}
    &\text{\textbf{Sequence-level Classification}} && \\
    &\quad \lambda_{\text{choice}}             &&= 0.01 \\
    &\quad \lambda_{\text{reward}}             &&= 0.5  \\
    &\quad \lambda_{\text{stimulus\_contrast}} &&= 0.1  \\[10pt]
    &\text{\textbf{Frame-level Regression}}    && \\
    &\quad \lambda_{\text{whisker\_motion}}    &&= 1.0  \\
    &\quad \lambda_{\text{wheel\_speed}}       &&= 1.0  \\
    &\quad \lambda_{\text{right\_paw\_speed}}  &&= 1.0  \\
    &\quad \lambda_{\text{left\_paw\_speed}}   &&= 1.0  \\
    &\quad \lambda_{\text{licking\_rate}}      &&= 1.0 
\end{alignat*}

\textbf{Finetuning on novel sessions.} We pretrained POYO+ for $35$ epochs and selected the checkpoint with the highest average validation metric across all pretraining tasks as the pretrained model initialization. We then evaluated transfer to novel sessions following the same protocol as the single-task, single-session supervised baseline models: for each novel session and each target task, we finetuned and evaluated a separate POYO+ model from the pretrained checkpoint. Finetuning used the POYO+ 10M model with batch size $32$, base learning rate $5\times10^{-4}$ and up to $300$ epochs with early stopping.

During finetuning, we used a gradual unfreezing strategy. At the beginning of finetuning, we fixed the pretrained backbone and only trained the \texttt{session\_embs} and \texttt{unit\_embs}. The full model was then unfrozen after epoch $40$. This design allows POYO+ to first adapt its session- and unit-specific representations to each novel recording session, before updating the shared pretrained backbone. This reduces early optimization instability and helps preserve the pretrained multi-task representation during adaptation to novel sessions.

\textbf{Calibration and embedding extraction for TS3.} TS3 evaluates unit-level
representations on sessions absent from pretraining. For POYO+, no row exists for a held-out
unit in the unit embedding table, so we must \emph{calibrate}: for each held-out session we initialize from the same
pretrained checkpoint and continue the pretraining objective on that session alone,
creating and fitting its unit and session embeddings. Unlike the TS1 protocol above,
one calibration covers all eight tasks jointly with the same weights $\lambda_t$, and
its only purpose is to produce embeddings. We calibrated one model per (session, seed) using batch size $32$, base learning rate $8.0\times10^{-4}$,
\texttt{pct\_start}$=0.5$, \texttt{div\_factor}$=1.0$, weight decay $10^{-4}$, bf16 precision,
and $100$ epochs, with the TS1 gradual unfreezing strategy unchanged (backbone frozen initially, then unfrozen after epoch
$40$). Within each session we defined a causal $90/10$ split of its own domain, and drew windows from the session's task-aligned
intervals intersected with the union of the tasks' domains, following the pretraining setup. Checkpoints were selected on validation loss with patience $50$.

A unit's representation is its row of \texttt{unit\_emb}: the pretraining part of each
embeddings file is read from the shared pretrained checkpoint, and the held-out part from
the calibration checkpoint of that unit's session. Nothing in the objective constrains
a row's magnitude, only that it be discriminative, so the two parts differ
systematically in length: mean $\|x\|$ of $4.42$ for pretrained rows against
$1.15$--$1.23$ for calibrated ones, a factor of roughly $3.7$. Because the
probes standardize using pretraining statistics, an affine map fitted on one part
preserves this ratio. We therefore apply a \emph{scale-only} normalization before
probing: the held-out half is multiplied by a single scalar so both halves have the
same mean vector length, leaving every direction and the held-out set's shape
unchanged. The scalar uses held-out norms but no held-out labels, which the
transductive setting permits since calibration has already accessed these sessions.

%% file: appendix/tuning/possm.tex
\subsubsection{POSSM}
\label{app:hyperparameters_possm}

\textbf{Multi-task pretraining and task weighting.}
We pretrained POSSM with the same weighted multi-task objective as POYO+, i.e.,
\begin{equation}
\mathcal{L}_{\mathrm{total}}=\sum_{t}\lambda_t\mathcal{L}_t.
\end{equation}
Pretraining was conducted in an 8-task setting with a 10M parameter version of POSSM for 100 epochs, using batch size $256$ and base learning rate $5\times10^{-5}$ on the nonfiltered processed dataset. We use the same task weights as POYO+ to pretrain POSSM:

\begin{alignat*}{2}
    &\text{\textbf{Sequence-level Classification}} && \\
    &\quad \lambda_{\text{choice}}             &&= 0.01 \\
    &\quad \lambda_{\text{reward}}             &&= 0.5  \\
    &\quad \lambda_{\text{stimulus\_contrast}} &&= 0.1  \\[10pt]
    &\text{\textbf{Frame-level Regression}}    && \\
    &\quad \lambda_{\text{whisker\_motion}}    &&= 1.0  \\
    &\quad \lambda_{\text{wheel\_speed}}       &&= 1.0  \\
    &\quad \lambda_{\text{right\_paw\_speed}}  &&= 1.0  \\
    &\quad \lambda_{\text{left\_paw\_speed}}   &&= 1.0  \\
    &\quad \lambda_{\text{licking\_rate}}      &&= 1.0 
\end{alignat*}

\textbf{Finetuning on novel sessions.} We pretrained POSSM for $100$ epochs and selected the checkpoint with the highest average validation metric across all pretraining tasks as the pretrained model initialization. We then evaluated transfer to novel sessions following the same protocol as the single-task, single-session supervised baseline models: for each novel session and each target task, we finetuned and evaluated a separate POSSM model from the pretrained checkpoint. Finetuning used the 10M parameter POSSM model with batch size $32$, base learning rate $5\times10^{-4}$ and up to $300$ epochs with early stopping

During finetuning, we used a gradual unfreezing strategy like POYO+. At the beginning of finetuning, we fixed the pretrained backbone and only trained the \texttt{session\_embs} and \texttt{unit\_embs}. The full model was then unfrozen after epoch $40$. As with POYO+, his design allows POSSM to first adapt its session- and unit-specific representations to each novel recording session, before updating the shared pretrained backbone. This reduces early optimization instability and helps preserve the pretrained multi-task representation during adaptation to novel sessions.

\textbf{Calibration and embedding extraction for TS3.} TS3 evaluates unit-level
representations on sessions absent from pretraining. For POSSM, no row exists for a
held-out unit in the unit embedding table, so we must \emph{calibrate}: for each
held-out session we initialize from the same pretrained checkpoint and continue the
pretraining objective on that session alone, creating and fitting its unit and session
embeddings. Unlike the TS1 protocol above, one calibration covers all eight tasks
jointly with the same weights $\lambda_t$, and its only purpose is to produce
embeddings. We calibrated one model per (session, seed) using batch size $32$, base
learning rate $8.0\times10^{-4}$,
\texttt{pct\_start}$=0.5$, \texttt{div\_factor}$=1.0$, weight decay $10^{-4}$, bf16
precision, and $100$ epochs, with the gradual unfreezing strategy (backbone
frozen initially, then unfrozen after epoch $20$; note this differs from the finetuning strategy to unfreeze at $40$, which we found in initial experiments to underperform). Within each session we defined a
causal $90/10$ split of its own domain, and drew windows from the session's
task-aligned intervals intersected with the union of the tasks' domains, following the
pretraining setup. Checkpoints were selected on validation loss with patience $50$.

A unit's representation is its row of \texttt{unit\_emb}: the pretraining part of each
embeddings file is read from the shared pretrained checkpoint, and the held-out part
from the calibration checkpoint of that unit's session. Nothing in the objective
constrains a row's magnitude, only that it be discriminative, so the two parts differ
systematically in length: mean $\|x\|$ of $4.67$ for pretrained rows against
$0.90$--$0.91$ for calibrated ones, a factor of roughly $5.2$. Because the probes
standardize using pretraining statistics, an affine map fitted on one part preserves
this ratio. We therefore apply a \emph{scale-only} normalization before probing: the
held-out half is multiplied by a single scalar so both halves have the same mean vector
length, leaving every direction and the held-out set's shape unchanged. The scalar uses
held-out norms but no held-out labels, which the transductive setting permits since
calibration has already accessed these sessions. Also note that we do not apply this scaling on linear probes as we found it to underperform compared to taking raw embedding scales, whereas with MLP probes it is necessary to prevent collapse.

%% file: appendix/tuning/ndt_stitch.tex
\subsubsection{NDT-Stitch}
\label{app:hyperparameters_ndt_stitch}

The model is pretrained with masked prediction using block masking, then fine-tuned on each task.
A key finding during development was that reducing the batch size to $16$ was critical to obtain good performance; larger batch sizes ($32$, $64$, $128$) consistently underperformed even after extensive learning rate sweeping across several orders of magnitude. After pretraining, we sweep the learning rate in finetuning with these values $\{10^{-5},2\times 10^{-5}, 5\times 10^{-5},10^{-4}, 2\times10^{-4}\}$.

We detail NDT-Stitch training parameters in Table~\ref{tab:ndt_stitch_config}.

\begin{table*}[h]
\caption{Pretraining configuration.}
\label{tab:ndt_stitch_config}
\begin{center}
\begin{small}
\begin{sc}
\begin{tabular}{lr}
\toprule
Hyperparameter & Value \\
\midrule
\multicolumn{2}{l}{\scriptsize Model parameters} \\
Backbone parameters                 & $10.42$M \\
Total parameters                    & $43.04$M \\
Bin size (s)                        & $0.02$ \\
Hidden dimension                    & $256$ \\
Encoder num.\ layers                & $13$ \\
Encoder num.\ heads                 & $8$ \\
Encoder FF dimension                & $1028$ \\
Encoder activation                  & \texttt{ReLU} \\
Encoder dropout                     & $0.2$ \\
Pre-encoder dropout                 & $0.2$ \\
Post-encoder dropout                & $0.2$ \\
T-Fixup scale base                  & $0.67$ \\
T-Fixup $V$ scale factor            & $\sqrt{2}$ \\
Custom initialization               & \checkmark \\
\midrule
\multicolumn{2}{l}{\scriptsize Masking parameters} \\
Mask ratio                          & $0.5$ \\
Max block size                      & $7$ \\
Block mask probability              & $0.5$ \\
\midrule
\multicolumn{2}{l}{\scriptsize Training parameters} \\
Number of epochs                    & $200$ \\
Batch size                          & $16$ \\
Weight decay                        & $10^{-4}$ \\
Base learning rate                  & $10^{-4}$ \\
LR scheduler -- pct start           & $0.5$ \\
LR scheduler -- div factor          & $1$ \\
\bottomrule
\end{tabular}
\end{sc}
\end{small}
\end{center}
\vskip -0.1in
\end{table*}

\textbf{Calibration and embedding extraction for TS3.} TS3 evaluates unit-level
representations on sessions absent from pretraining, which for NDT-Stitch requires a
stitcher for each held-out session. We therefore \emph{calibrate}: for each held-out
session we initialize the backbone from the pretrained checkpoint, initialize that
session's input and output stitchers from scratch, and continue the masked-prediction
objective on that session alone, with the same masking parameters used in pretraining
(mask ratio $0.5$, max block size $7$, block mask probability $0.5$). No gradual unfreezing is used: the entire model trains from the first epoch. We
calibrated one model per (session, seed) using batch size $32$, base learning rate
$10^{-4}$ (one-cycle peak $\eta_{\max}=\eta_{\mathrm{base}}\sqrt{B}=5.7\times10^{-4}$),
\texttt{pct\_start}$=0.5$, \texttt{div\_factor}$=1.0$, weight decay $10^{-4}$, bf16
precision, and up to $500$ epochs. Within each session we defined a causal $90/10$
split of its own domain, and drew windows from the session's task-aligned intervals,
following the pretraining setup. Checkpoints were selected on validation loss with
patience $50$.

A unit's representation is the concatenation of its column of that session's input
stitcher and its row of the output stitcher. Both the pretraining and held-out parts of
each embeddings file are read this way, from the pretrained checkpoint and from each
session's calibration checkpoint respectively. Because a stitcher must map its
session's spike counts onto the scale the backbone expects, the objective pins its
magnitude per session, and the two parts agree in length without intervention (mean
$\|x\|$ of $1.44$ against $1.47$), hence no normalization is required on the embedding scales.

%% file: appendix/tuning/mtm.tex
\subsubsection{MtM}
\label{app:hyperparameters_mtm}

MtM shares a very similar architecture to NDT-Stitch, consistent with NDT-Stitch, we found batch size to be critical: reducing it to $16$ was necessary to observe stable transfer to downstream tasks, and larger batch sizes failed to match this performance even after extensive learning rate sweeping. We applied the same fine-tuning strategy as NDT-Stitch.

We detail MtM training parameters in Table~\ref{tab:mtm_config}.

\begin{table*}[h]
\caption{Model and Training Configuration.}
\label{tab:mtm_config}
\begin{center}
\begin{small}
\begin{sc}
\begin{tabular}{lr}
\toprule
Hyperparameter & Value \\
\midrule
\multicolumn{2}{l}{\scriptsize Model parameters} \\
Bin size (s)                        & $0.02$ \\
Hidden dimension                    & $256$ \\
Encoder num.\ layers                & $12$ \\
Encoder num.\ heads                 & $8$ \\
Encoder FF dimension                & $1028$ \\
Encoder activation                  & \texttt{GELU} \\
Encoder dropout                     & $0.2$ \\
Pre-encoder dropout                 & $0.2$ \\
Post-encoder dropout                & $0.2$ \\
T-Fixup scale base                  & $0.67$ \\
T-Fixup $V$ scale factor            & $\sqrt{2}$ \\
Custom initialization               & \checkmark \\
\midrule
\multicolumn{2}{l}{\scriptsize Masking parameters} \\
Mask ratio                          & $0.3$ \\
Num. mask regions                   & $1$ \\
Causal ratio                        & $0.1$ \\
\midrule
\multicolumn{2}{l}{\scriptsize Training parameters} \\
Number of epochs                    & $200$ \\
Batch size                          & $16$ \\
Weight decay                        & $10^{-4}$ \\
Base learning rate                  & $10^{-4}$ \\
LR scheduler -- pct start           & $0.15$ \\
LR scheduler -- div factor          & $10$ \\
Precision                           & \texttt{bf16} \\
\bottomrule
\end{tabular}
\end{sc}
\end{small}
\end{center}
\vskip -0.1in
\end{table*}

\textbf{Calibration and embedding extraction for TS3.} TS3 evaluates unit-level
representations on sessions absent from pretraining, which for MtM requires a stitcher
for each held-out session. We therefore \emph{calibrate}: for each held-out session we
initialize the backbone from the pretrained checkpoint, initialize that session's input
and output stitchers from scratch, and continue the masked-prediction objective on that
session alone. Following the finetuning recipe rather than the pretraining
configuration above, calibration uses a harder mask (mask ratio $0.5$, causal ratio
$0.5$, sampling over neuron, causal, inter-region and intra-region mask types) and a
flat one-cycle schedule. No gradual unfreezing is used: the entire model trains from
the first epoch. We calibrated one model per (session, seed) using batch size $32$,
base learning rate $10^{-4}$ (one-cycle peak
$\eta_{\max}=\eta_{\mathrm{base}}\sqrt{B}=5.7\times10^{-4}$),
\texttt{pct\_start}$=0.5$, \texttt{div\_factor}$=1.0$, weight decay $10^{-4}$, bf16
precision, and up to $500$ epochs. Within each session we defined a causal $90/10$
split of its own domain, and drew windows from the session's task-aligned intervals,
following the pretraining setup. Checkpoints were selected on validation loss with
patience $50$.

A unit's representation is the concatenation of its column of that session's input
stitcher and its row of the output stitcher. Both the pretraining and held-out parts of
each embeddings file are read this way, from the pretrained checkpoint and from each
session's calibration checkpoint respectively. Because a stitcher must map its
session's spike counts onto the scale the backbone expects, the objective pins its
magnitude per session, and the two parts remain within a small factor of each other
(mean $\|x\|$ of $1.70$ against $2.22$, the held-out part being the longer), hence no
normalization is required on the embedding scales.

%% file: appendix/tuning/ndt2.tex
\subsubsection{NDT2}
\label{app:hyperparameters_ndt2}
A key source of difficulty in applying NDT2 to \brainwidebench~is the mismatch between the data regime for which it was originally developed and our evaluation setting. NDT2 was designed and validated on monkey recordings from Blackrock Utah arrays, which typically yield a large number of units with relatively dense, structured spiking activity. Our benchmark, by contrast, includes Neuropixel recordings from mouse, which differ substantially in array geometry, unit count, and spike statistics. This introduces a fundamental challenge around patch size: the number of neurons per patch directly determines the sequence length seen by the encoder, and hence GPU memory requirements. When units are filtered with QC, the reduced neuron count allows smaller, more tractable patches. However, our evaluation protocol requires using \textit{unfiltered} units, which inflates the neuron count and forces larger patches to remain within memory budgets. This introduces a systematic gap between the pretraining regime, where filtered units and smaller patches can be used, and evaluation, where larger patches are imposed by the unfiltered setting, potentially degrading the quality of learned representations.

Despite these challenges, pretraining appeared to proceed successfully: reconstruction quality on held-out masked regions, measured via the $D^2$ metric from TS2, was positive and indicative of meaningful learned structure. However, this did not translate to downstream decoding performance. We attempted both the original paper's calibration protocol (SSL finetuning on the target session followed by supervised decoding) and direct full finetuning, sweeping learning rates across both stages, but the best results obtained are those reported in Table~\ref{tab:neds_ndt2_results}, which remain at or below the linear baseline.

We omit NDT2 from the main comparison table and analysis, as our evaluation revealed the model to be in a failure mode: performance consistently remains at or below the single-session linear baseline across tasks, indicating that the pretrained representations do not provide meaningful nonlinear structure beyond what a simple linear decoder already captures. We note that this outcome also highlights a broader difficulty in iterating on methods of this class: because results depend jointly on pretraining scale, patch size, filtering choices, and downstream finetuning, the exploration space is large and each candidate configuration requires rerunning a full pretraining pipeline, severely limiting the number of configurations that can be practically evaluated. 

Full details of the hyperparameter search, including learning rate schedules, calibration duration, stitcher initialization strategies, patch size configurations, and finetuning depth (frozen encoder vs.\ full finetuning), are reported in Appendix~\ref{app:hyperparameters_ndt2}.

%% file: appendix/tuning/neds.tex
\subsubsection{NEDS}
\label{app:hyperparameters_neds}

Notably, NEDS shares important similarities with our benchmark setting: the original model was also pretrained on data from the International Brain Laboratory~\cite{ibl_decision_task}, providing a degree of overlap with our pretraining distribution. However, key differences remain. The original NEDS pretraining operates at a smaller scale, spanning up to 83 animals, and uses a longer context window of approximately 2 seconds centered around stimulus onset. Furthermore, the original pretraining covers a narrower set of task types 2 sequence-level and 2 frame-level targets — which may limit transfer to the broader range of decoding tasks evaluated in our benchmark. An additional source of discrepancy is our evaluation protocol, which uses causal temporal splits (TS1) that differ from the original training and evaluation setup.

We were unable to reproduce the original pretrained model results. We re-implemented it in the benchmark codebase and pretrained it from scratch. Pretraining proceeded successfully, even if we observed that some tasks can be overfit more easily than others (sequence-level tasks in particular). However, finetuning on downstream evaluation sessions did not yield strong decoding performance: we observed a consistent generalization gap between validation and test sets, mirroring the failure mode encountered with NDT2. We swept learning rates across finetuning stages and additionally explored a calibration strategy analogous to NDT2's neural calibration, using the masked pretraining objective to adapt the model to each new evaluation session before supervised finetuning. This calibration did not improve downstream performance, and the best results obtained across all configurations are those reported in Table~\ref{tab:neds_ndt2_results}.

%% file: appendix/computing.tex
\section{Computing Resources}
\label{app:computing_resources}

\begin{table}[h]
\centering
\small
\setlength{\tabcolsep}{4pt}
\caption{Computational resources used across the three task suites. Hyperparameter tuning sweeps are run on A10 GPUs (AMD EPYC 7502); all other stages are run on B200 GPUs (Xeon Platinum 8570). Tuning runtimes are estimated by sampling 200 runs; all other figures are measured from W\&B run logs for the released projects, excluding crashed and failed runs. Training/finetuning runtimes are reported as the mean over all runs in the block, where a block spans all tasks $\times$ sessions $\times$ seeds.}
\label{tab:bwb-compute}
\begin{tabular}{llccccr}
\toprule
\textbf{Suite} & \textbf{Model} & \textbf{Stage} & \textbf{GPU} & \textbf{Trials/Seeds} & \textbf{Runtime} & \textbf{Total (GPU-hr)} \\
\midrule
\multirow{22}{*}{TS1}
 & Linear     & Tuning      & A10  & 100  & 1.2 hr (sweep) & 339.6 \\
 & MLP        & Tuning      & A10  & 100  & 1.3 hr (sweep) & 378.5 \\
 & GRU        & Tuning      & A10  & 100  & 1.5 hr (sweep) & 429.9 \\
 & TCN        & Tuning      & A10  & 100  & 1.3 hr (sweep) & 386.1 \\
 & CEBRA      & Tuning      & A10  & 100  & 9.1 hr (sweep) & 2638.7 \\
 & POYO       & Tuning      & A10  & 50   & 1.2 hr (sweep) & 352.6 \\
 & NDT        & Tuning      & A10  & 50   & 0.9 hr (sweep) & 256.7 \\
\cmidrule(lr){2-7}
& Linear     & Training    & B200 & 1160 & 2.1 min / run  & 41.0 \\
& MLP        & Training    & B200 & 1160 & 1.8 min / run  & 35.7 \\
& GRU        & Training    & B200 & 1160 & 2.1 min / run  & 40.1 \\
& CNN        & Training    & B200 & 1160 & 2.1 min / run  & 39.7 \\
& CEBRA      & Training    & B200 & 1160 & 5.2 min / run  & 101.2 \\
& POYO       & Training    & B200 & 1160 & 2.1 min / run  & 41.5 \\
& NDT        & Training    & B200 & 1160 & 1.9 min / run  & 36.2 \\
\cmidrule(lr){2-7}
& POYO+      & Pretraining & B200 & 1    & 6.0 hr / seed  & 6.0 \\
& POSSM      & Pretraining & B200 & 1    & 7.8 hr / seed  & 7.8 \\
& NDT-stitch & Pretraining & B200 & 1    & 43.4 hr / seed & 43.4 \\
& MtM        & Pretraining & B200 & 1    & 45.8 hr / seed & 45.8 \\
\cmidrule(lr){2-7}
& POYO+      & Finetuning  & B200 & 1160 & 3.8 min / run  & 73.2 \\
& POSSM      & Finetuning  & B200 & 1160 & 1.9 min / run  & 36.9 \\
& NDT-stitch & Finetuning  & B200 & 1160 & 1.4 min / run  & 26.2 \\
& MtM        & Finetuning  & B200 & 1160 & 1.6 min / run  & 30.6 \\
\midrule
\multirow{8}{*}{TS2}
 & Autoencoder & Tuning     & A10  & 100  & 6.2 hr (sweep) & 357.6 \\
 & NDT         & Tuning     & A10  & 50   & 3.2 hr (sweep) & 188.1 \\
\cmidrule(lr){2-7}
& Autoencoder & Training   & B200 & 290  & 19.8 min / run & 95.8 \\
& LFADS       & Training   & B200 & 290  & 29.7 min / run & 143.3 \\
& NDT         & Training   & B200 & 290  & 12.6 min / run & 61.1 \\
& Stat.\ baselines & Training & B200 & 290 & 1.1 min / run & 5.3 \\
\cmidrule(lr){2-7}
& NDT-stitch  & Finetuning & B200 & 290  & 10.4 min / run & 50.4 \\
& MtM         & Finetuning & B200 & 290  & 9.9 min / run  & 47.7 \\
\midrule
\multirow{7}{*}{TS3}
& NuCLR & Pretraining & B200 & 5   & 5.8 hr / seed  & 28.9 \\
& NEMO  & Pretraining & B200 & 5   & 0.2 hr / seed  & 1.0 \\
\cmidrule(lr){2-7}
& POYO+      & Calibration & B200 & 145 & 21.7 min / run & 52.5 \\
& POSSM      & Calibration & B200 & 145 & 10.5 min / run & 25.4 \\
& MtM        & Calibration & B200 & 145 & 17.9 min / run & 43.1 \\
& NDT-stitch & Calibration & B200 & 145 & 13.2 min / run & 32.0 \\
\bottomrule
\end{tabular}
\end{table}

%% file: appendix/additional_results.tex
\section{Additional Results}
\label{app:additional_results}

In this section, we present a wider array of results across the 3 task suites, including per-session results, additional metrics, and additional analyses.

\subsection{TS1 Additional Results: Behavior Prediction}

\subsubsection{Per-session breakdown}

\begin{figure}[ht]
    \centering
    \includegraphics[width=\linewidth]{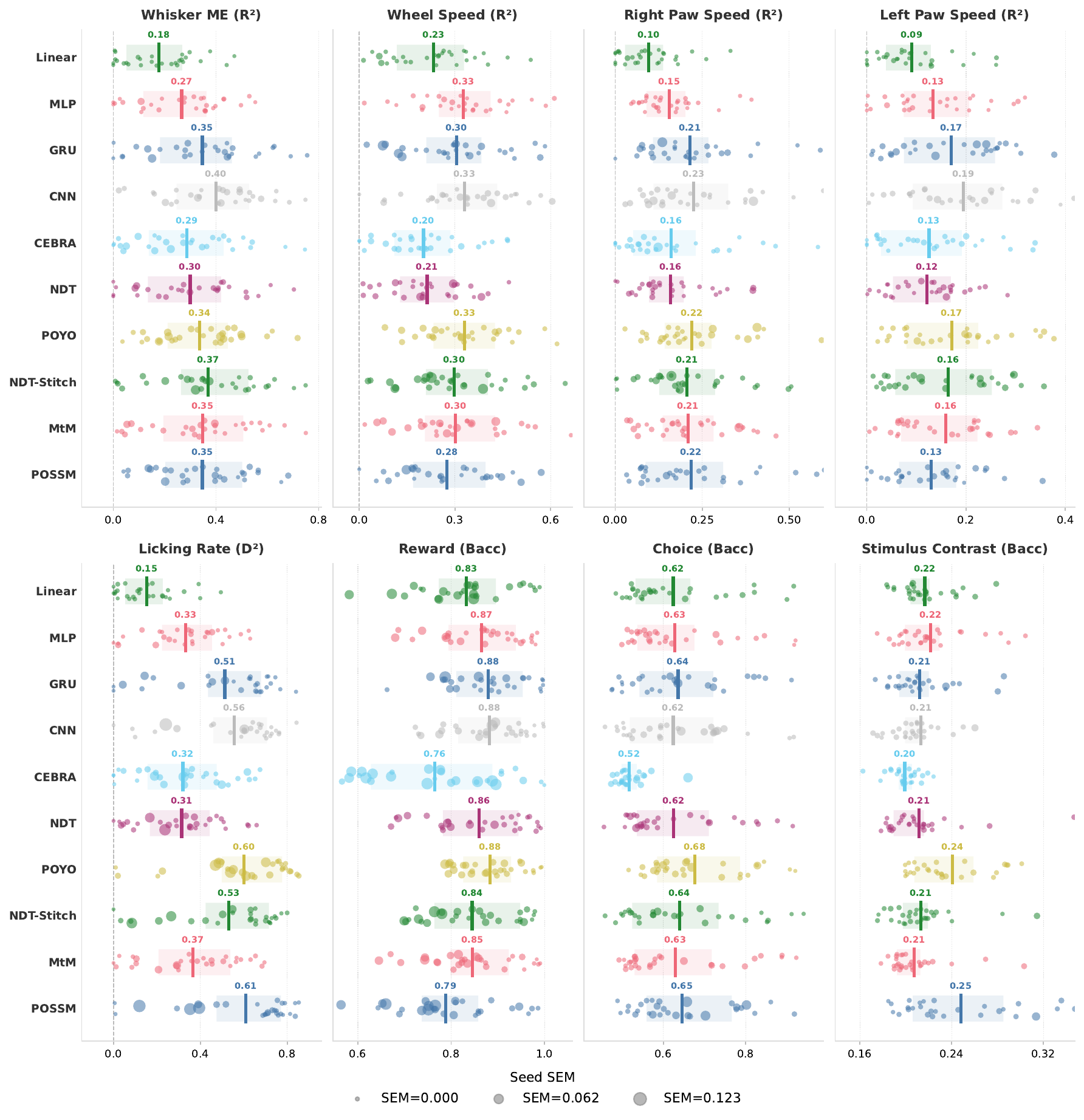}
    \caption{\textbf{Per-Session Behavior Decoding Performance Across Baselines (TS1).} Each panel shows performance for a single decoding target, Whisker Motion Energy, Wheel Speed, Right Paw Speed, and Left Paw Speed ($R^2$, top row), and Licking Rate ($D^2$), Reward, Choice, and Stimulus Contrast (balanced accuracy, bottom row), broken down by model and evaluation session. Each dot represents a single session, with dot size proportional to the standard error of the mean (SEM) across 5 random seeds; larger dots indicate higher variability across seeds. The vertical bar indicates the median performance across sessions, and the shaded region spans the interquartile range. Dots are jittered vertically for visibility.}
    \label{fig:ts1_main_metric_distribution}
\end{figure}

\begin{figure}[ht]
    \centering
    \includegraphics[width=\linewidth]{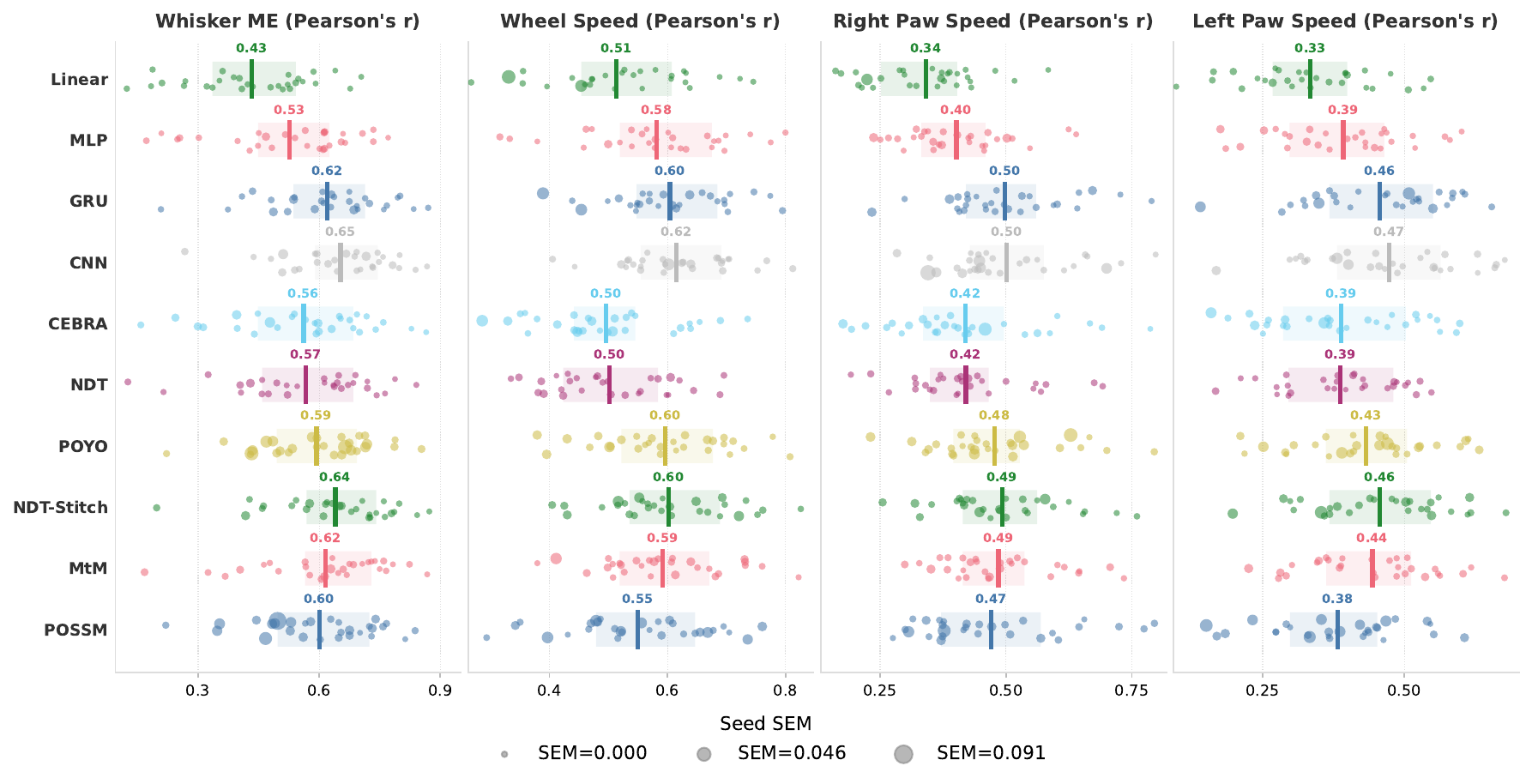}
    \caption{\textbf{Alternative Per-Session Behavior Decoding Performance Across Baselines for Regression Tasks (TS1).} Each panel shows performance for a single regression target, Whisker Motion Energy, Wheel Speed, Right Paw Speed, and Left Paw Speed, using Pearson's $r$, broken down by model and evaluation session.
    Plotting conventions as in Figure~\ref{fig:ts1_main_metric_distribution}.}
\label{fig:ts1_pearson_distribution}
\end{figure}

\begin{figure}[ht]
    \centering
    \includegraphics[width=\linewidth]{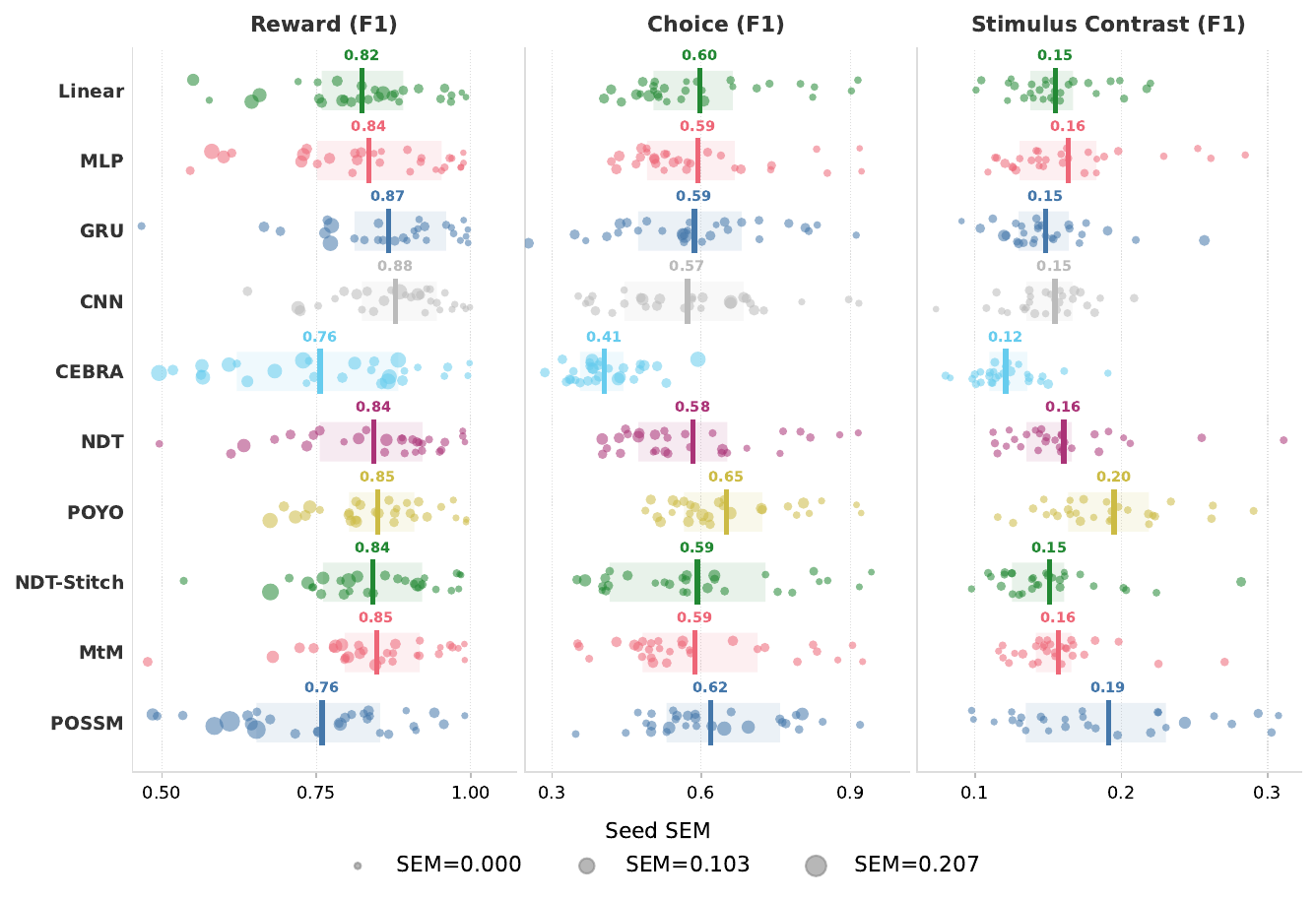}
    \caption{\textbf{Alternative Per-Session Behavior Decoding Performance Across Baselines for Classification Tasks (TS1).} Each panel shows performance for a single classification target, Reward, Choice, and Stimulus Contrast, using F1, broken down by model and evaluation session.
    Plotting conventions as in Figure~\ref{fig:ts1_main_metric_distribution}.}
    \label{fig:ts1_f1_distribution}
\end{figure}

The main text reports TS1 results (Table~\ref{tab:ts1_main}) averaged across all 29 evaluation sessions. Here we provide per-session breakdowns for both the primary metrics ($R^2$/$D^2$ for regression tasks and balanced accuracy for classification tasks; Figure~\ref{fig:ts1_main_metric_distribution}) and supplementary metrics: Pearson's $r$ for regression (Figure~\ref{fig:ts1_pearson_distribution}) and F1-score for classification (Figure~\ref{fig:ts1_f1_distribution}). Average per-task rankings across all 29 sessions are additionally reported in Table~\ref{tab:ts1_avg_ranks_app}.

\clearpage
\newpage

\subsubsection{Per-task ranks}

We report per-task average ranks in Table~\ref{tab:ts1_avg_ranks_app}, following the procedure detailed in Appendix~\ref{app:rank_procedure}.

\begin{table}[ht]
\centering
\caption{\footnotesize{\textbf{Average per-task rankings in TS1.} Comparison of average ranks across all tasks for single-session and pretrained models. Ranks are determined by pairwise one-sided Welch's t-tests ($\alpha=0.05$) using standard competition (1224) ranking. Lower ranks indicate better performance.}}
\label{tab:ts1_avg_ranks_app}
\footnotesize
\begin{tabular}{llcccccccc>{\cellcolor{blue!10}}c}
\toprule
& \textbf{Method} & \textbf{Licks} & \textbf{Whisker} & \textbf{Wheel} & \textbf{RPaw} & \textbf{LPaw} & \textbf{Reward} & \textbf{Choice} & \textbf{Stim} & \textbf{Avg. Rank $\downarrow$} \\
\midrule
\multirow{7}{*}{\rotatebox[origin=c]{90}{Single-Session}}
& Linear     & 10.10 & 9.66 & 8.52 & 9.10 & 9.38 & 6.31 & 5.07 & 3.69 & 7.73 \\
& MLP        & 7.76 & 7.34 & 4.21 & 6.34 & 6.03 & 5.17 & 3.72 & 3.10 & 5.46 \\
& GRU        & 4.55 & 3.83 & 4.41 & 4.03 & 3.52 & 3.69 & 4.03 & 3.62 & 3.96 \\
& CNN        & 3.48 & 2.86 & 3.48 & 3.59 & 2.48 & 3.90 & 5.24 & 4.86 & 3.74 \\
& CEBRA      & 8.28 & 7.21 & 8.24 & 6.55 & 7.03 & 6.86 & 8.86 & 6.48 & 7.44 \\
& NDT        & 8.24 & 7.03 & 8.72 & 7.83 & 7.38 & 4.45 & 5.34 & 5.41 & 6.80 \\
& POYO       & 3.52 & 6.45 & 5.00 & 5.31 & 6.03 & 3.66 & 4.07 & 4.34 & 4.80 \\
\midrule
\multirow{4}{*}{\rotatebox[origin=c]{90}{Pretrain}}
& POYO+      & 1.34 & 2.97 & 2.86 & 2.90 & 2.66 & 3.83 & 2.03 & 1.66 & 2.53 \\
& POSSM      & 1.97 & 4.10 & 2.83 & 3.03 & 3.45 & 3.83 & 2.93 & 3.31 & 3.18 \\
& NDT-Stitch & 4.28 & 2.72 & 4.31 & 3.62 & 3.55 & 4.72 & 2.52 & 2.93 & 3.58 \\
& MtM        & 6.10 & 3.59 & 4.14 & 3.86 & 3.31 & 4.17 & 3.17 & 4.07 & 4.05 \\
\bottomrule
\end{tabular}

\end{table}

\subsubsection{Secondary metrics}

Table~\ref{tab:ts1_main} reports a single primary metric per task. Here we report the
secondary metrics recorded by the evaluation protocol, averaged over the 29 evaluation
sessions: Pearson's $r$ for the four continuous regression targets
(Table~\ref{tab:ts1_pearson_app}), and F1-score together with average precision (AP) for
the three classification targets (Table~\ref{tab:ts1_classif_secondary_app}). These
session-averaged summaries complement the per-session distributions shown in
Figures~\ref{fig:ts1_pearson_distribution} and~\ref{fig:ts1_f1_distribution}; AP is
reported here only. Licking rate admits no secondary metric under our protocol: Pearson's
$r$ is not meaningful for a count target, so Poisson $D^2$ is the sole metric for that
task. Note that both Pearson's $r$ and AP are insensitive to the scale of a model's outputs: $r$ is invariant to affine rescaling of the predictions, and AP depends only on the ranking of the predicted scores. They therefore isolate how well each method tracks the target, separately from how well its outputs are calibrated. In particular, $r$ measures co-variation with the behavioral time series independently of scale and offset, and AP measures class discriminability independently of the decision threshold.

\begin{table}[ht]
\centering
\caption{\footnotesize{\textbf{Pearson's $r$ on the TS1 regression tasks.} Session-averaged
Pearson correlation between predicted and observed behavior for the four continuous
targets, reported as mean $\pm$ SEM over 5 seeds. Licking rate is excluded, as Pearson's $r$ is not appropriate for a
count target. Higher is better.}}
\label{tab:ts1_pearson_app}
\footnotesize
\begin{tabular}{llcccc}
\toprule
& \textbf{Method} & \textbf{Whisker} & \textbf{Wheel} & \textbf{RPaw} & \textbf{LPaw} \\
\midrule
\multirow{7}{*}{\rotatebox[origin=c]{90}{Single-Session}}
& Linear      & $0.431\tpms{0.000}$ & $0.504\tpms{0.001}$ & $0.325\tpms{0.001}$ & $0.325\tpms{0.001}$ \\
& MLP         & $0.529\tpms{0.001}$ & $0.583\tpms{0.000}$ & $0.406\tpms{0.001}$ & $0.393\tpms{0.001}$ \\
& GRU         & $0.625\tpms{0.001}$ & $0.597\tpms{0.002}$ & $0.499\tpms{0.001}$ & $0.460\tpms{0.001}$ \\
& CNN         & $0.652\tpms{0.001}$ & $0.616\tpms{0.001}$ & $0.502\tpms{0.003}$ & $0.472\tpms{0.001}$ \\
& CEBRA       & $0.559\tpms{0.002}$ & $0.497\tpms{0.000}$ & $0.420\tpms{0.003}$ & $0.384\tpms{0.002}$ \\
& NDT         & $0.569\tpms{0.001}$ & $0.507\tpms{0.001}$ & $0.424\tpms{0.001}$ & $0.392\tpms{0.001}$ \\
& POYO        & $0.558\tpms{0.003}$ & $0.572\tpms{0.001}$ & $0.454\tpms{0.002}$ & $0.402\tpms{0.002}$ \\
\midrule
\multirow{4}{*}{\rotatebox[origin=c]{90}{Pretrain}}
& POYO+       & $0.631\tpms{0.001}$ & $0.616\tpms{0.000}$ & $0.505\tpms{0.002}$ & $0.468\tpms{0.000}$ \\
& POSSM       & $0.626\tpms{0.002}$ & $0.613\tpms{0.001}$ & $0.493\tpms{0.002}$ & $0.444\tpms{0.001}$ \\
& NDT-Stitch  & $0.648\tpms{0.002}$ & $0.611\tpms{0.001}$ & $0.505\tpms{0.003}$ & $0.462\tpms{0.002}$ \\
& MtM         & $0.629\tpms{0.002}$ & $0.603\tpms{0.001}$ & $0.495\tpms{0.001}$ & $0.452\tpms{0.000}$ \\
\bottomrule
\end{tabular}
\end{table}

\begin{table}[ht]
\centering
\caption{\footnotesize{\textbf{Secondary metrics on the TS1 classification tasks.}
Session-averaged F1-score and average precision (AP) for the three classification targets,
reported as mean $\pm$ SEM over 5 seeds. Both are macro-averaged over classes. AP is not
comparable across tasks, as each has a different number of classes and class prior;
chance-level macro AP is approximately the mean class prevalence ($0.5$ for the binary
tasks under a balanced prior, $0.2$ for the five-way Stimulus Contrast task). Higher is
better.}}
\label{tab:ts1_classif_secondary_app}
\footnotesize
\begin{tabular}{ll cc cc cc}
\toprule
& & \multicolumn{2}{c}{\textbf{Reward}} & \multicolumn{2}{c}{\textbf{Choice}} & \multicolumn{2}{c}{\textbf{Stim}} \\
\cmidrule(lr){3-4} \cmidrule(lr){5-6} \cmidrule(lr){7-8}
& \textbf{Method} & F1 & AP & F1 & AP & F1 & AP \\
\midrule
\multirow{7}{*}{\rotatebox[origin=c]{90}{Single-Session}}
& Linear      & $0.808\tpms{0.005}$ & $0.934\tpms{0.002}$ & $0.582\tpms{0.005}$ & $0.675\tpms{0.003}$ & $0.160\tpms{0.002}$ & $0.242\tpms{0.002}$ \\
& MLP         & $0.847\tpms{0.006}$ & $0.967\tpms{0.001}$ & $0.600\tpms{0.005}$ & $0.692\tpms{0.002}$ & $0.170\tpms{0.001}$ & $0.248\tpms{0.002}$ \\
& GRU         & $0.874\tpms{0.005}$ & $0.961\tpms{0.003}$ & $0.600\tpms{0.008}$ & $0.686\tpms{0.006}$ & $0.151\tpms{0.002}$ & $0.242\tpms{0.002}$ \\
& CNN         & $0.883\tpms{0.005}$ & $0.969\tpms{0.002}$ & $0.575\tpms{0.002}$ & $0.652\tpms{0.002}$ & $0.145\tpms{0.001}$ & $0.228\tpms{0.001}$ \\
& CEBRA       & $0.775\tpms{0.010}$ & $0.898\tpms{0.007}$ & $0.414\tpms{0.007}$ & $0.549\tpms{0.001}$ & $0.118\tpms{0.002}$ & $0.223\tpms{0.002}$ \\
& NDT         & $0.857\tpms{0.005}$ & $0.963\tpms{0.001}$ & $0.577\tpms{0.004}$ & $0.651\tpms{0.002}$ & $0.154\tpms{0.002}$ & $0.233\tpms{0.001}$ \\
& POYO        & $0.865\tpms{0.003}$ & $0.957\tpms{0.006}$ & $0.591\tpms{0.008}$ & $0.665\tpms{0.006}$ & $0.173\tpms{0.002}$ & $0.248\tpms{0.003}$ \\
\midrule
\multirow{4}{*}{\rotatebox[origin=c]{90}{Pretrain}}
& POYO+       & $0.850\tpms{0.002}$ & $0.957\tpms{0.001}$ & $0.669\tpms{0.004}$ & $0.750\tpms{0.003}$ & $0.218\tpms{0.001}$ & $0.297\tpms{0.001}$ \\
& POSSM       & $0.821\tpms{0.005}$ & $0.957\tpms{0.001}$ & $0.643\tpms{0.006}$ & $0.707\tpms{0.004}$ & $0.155\tpms{0.002}$ & $0.249\tpms{0.001}$ \\
& NDT-Stitch  & $0.858\tpms{0.007}$ & $0.965\tpms{0.001}$ & $0.643\tpms{0.002}$ & $0.737\tpms{0.003}$ & $0.176\tpms{0.001}$ & $0.259\tpms{0.001}$ \\
& MtM         & $0.872\tpms{0.002}$ & $0.967\tpms{0.001}$ & $0.624\tpms{0.003}$ & $0.692\tpms{0.004}$ & $0.172\tpms{0.003}$ & $0.236\tpms{0.001}$ \\
\bottomrule
\end{tabular}
\end{table}


\subsection{TS2 Additional Results: Neural Activity Prediction} \label{app:additional_results_ts2}

\subsubsection{Statistical Baselines for TS2}
\label{app:ts2_statistical}

In Table~\ref{tab:app_ts2_full} we report results on statistical baselines, described in Appendix~\ref{app:ts2_statistical}, along with the full results already in Table~\ref{tab:table2}.

\begin{table}[t!]
\centering
\small
\caption{\footnotesize{{\bf Task Suite 2: Co-smoothing and forecasting performance.} Model performance for both tasks is reported in terms of $D^2$ and bps. All metrics are reported as the average over evaluation sessions ($\tpms{\text{SEM}}$ over 5 finetuning seeds). 
Rankings incorporate statistical significance, see Appendix~\ref{app:rank_procedure} for more details.}}
\vspace{2mm}
\resizebox{0.85\columnwidth}{!}{
\begin{tabular}{+c|^l|^c^c|^c^c|^R}
\toprule
& \textbf{Method} 
& \multicolumn{2}{c|}{\bf Co-smoothing}
& \multicolumn{2}{c|}{\bf Forecasting}
& \multicolumn{1}{R}{Average} \\
& 
& $D^2$ & bps 
& $D^2$ & bps
& Rank \\\midrule

\multirow{9}{*}{\rotatebox[origin=c]{90}{\textbf{Single-Session}}}
& Population coupling 
& $0.003$
& $0.010$
& -- 
& -- 
& -- \\

& RRR 
& $0.112$
& $0.304$
& -- 
& -- 
& -- \\

& RRR (w/ ISI features) 
& $0.128$
& $0.328$
& -- 
& -- 
& -- \\

& Trailing mean 
& -- 
& -- 
& $0.002$
& $0.025$
& -- \\

& Shrinkage 
& -- 
& -- 
& $0.077$
& $0.161$
& -- \\

& Ridge autoregression 
& -- 
& -- 
& $0.108$
& $0.253$
& -- \\

& Autoencoder 
& $0.088 \tpms{0.000}$ 
& $0.200 \tpms{0.001}$ 
& $0.037 \tpms{0.001}$ 
& $0.065 \tpms{0.001}$ 
& 4.81 \\

& LFADS 
& $0.176 \tpms{0.000}$ 
& $0.421 \tpms{0.001}$ 
& $0.144 \tpms{0.000}$ 
& $0.325 \tpms{0.001}$ 
& 2.34 \\

& NDT 
& $0.132 \tpms{0.001}$ 
& $0.304 \tpms{0.003}$ 
& $0.158 \tpms{0.000}$ 
& $0.343 \tpms{0.001}$ 
& 2.79 \\

\midrule
\multirow{2}{*}{\rotatebox[origin=c]{90}{\textbf{Pre}}}

& NDT-Stitch
& $0.157 \tpms{0.001}$ 
& $0.369 \tpms{0.002}$ 
& $0.177 \tpms{0.000}$ 
& $0.384 \tpms{0.000}$ 
& 1.64 \\

& MtM
& $0.191 \tpms{0.001}$ 
& $0.459 \tpms{0.002}$ 
& $0.131 \tpms{0.001}$ 
& $0.288 \tpms{0.001}$ 
& 2.57 \\

\bottomrule

\end{tabular}
}
\vspace{-2mm}
\label{tab:app_ts2_full}
\end{table}

\subsubsection{Per-session breakdown}

\begin{figure}[ht]
    \centering
    \begin{minipage}{0.48\linewidth}
        \centering
        \includegraphics[width=\linewidth]{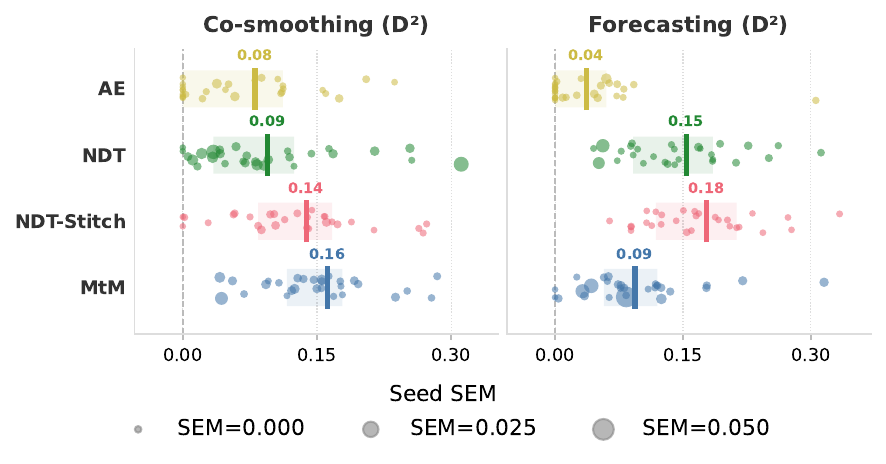}
    \end{minipage}
    \hfill
    \begin{minipage}{0.48\linewidth}
        \centering
        \includegraphics[width=\linewidth]{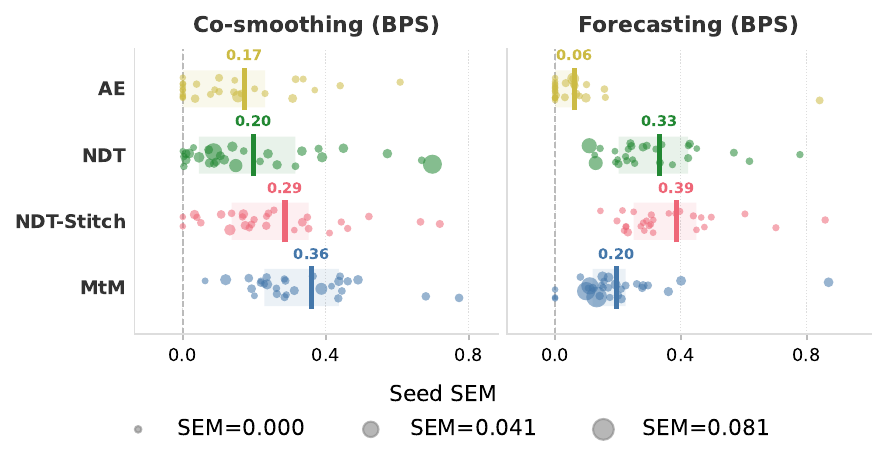}
    \end{minipage}
    \caption{\textbf{Per-Session Neural Activity Prediction Performance Across Baselines (TS2).} Left panels: $D^2$ metric; Right panels: BPS metric. Each panel shows performance for a single task (co-smoothing left subplot or forecasting right subplot) broken down by model and evaluation session.
    Plotting conventions as in Figure~\ref{fig:ts1_main_metric_distribution}.
    }
    \label{fig:ts2_main_metric_distribution}
\end{figure}

The main text reports TS2 results (Table~\ref{tab:table2}) averaged across all 29 evaluation sessions; here we provide the corresponding per-session breakdowns (Figure~\ref{fig:ts2_main_metric_distribution}).

\subsubsection{Per-task ranks}

In Tables~\ref{tab:ts2_avg_ranks_app_nostat} and~\ref{tab:ts2_avg_ranks_app_stat} we provide the average per-task rankings for TS2, following the procedure detailed in Appendix~\ref{app:rank_procedure}. Note that the stats baselines are excluded from the main rankings since they are task-specific, hence they would interfere with the rankings of the other methods that can be ranked on both tasks. Below, we include two tables with per-task rankings: one excluding the stats baselines (matching Table~\ref{tab:table2}), and one including them.

\begin{table}[ht]
\centering
\caption{\footnotesize{\textbf{Average per-task rankings in TS2, excluding stats baselines.} Comparison of average ranks for the co-smoothing and forecasting tasks. Ranks are determined by pairwise one-sided Welch's t-tests ($\alpha=0.05$) using standard competition (1224) ranking. Lower ranks indicate better performance.}}
\label{tab:ts2_avg_ranks_app_nostat}
\small

\begin{tabular}{llcc>{\cellcolor{blue!10}}c}
\toprule
& \textbf{Method} & \textbf{Co-smoothing} & \textbf{Forecasting} & \textbf{Avg. Rank $\downarrow$} \\
\midrule
\multirow{3}{*}{\rotatebox[origin=c]{90}{SS}}
& AE             & 4.69 & 4.93 & 4.81 \\
& NDT            & 3.55 & 2.03 & 2.79 \\
& LFADS          & 1.76 & 2.93 & 2.34 \\
\midrule
\multirow{2}{*}{\rotatebox[origin=c]{90}{Pre.}}
& NDT-Stitch     & 2.28 & 1.00 & 1.64 \\
& MtM            & 1.34 & 3.79 & 2.57 \\
\bottomrule
\end{tabular}
  
\end{table}

\begin{table}[ht]
\centering
\caption{\footnotesize{\textbf{Average per-task rankings in TS2, including stats baselines.} Comparis\ref{app:computing_resources}on of average ranks for the co-smoothing and forecasting tasks, including stats baselines. Ranks are determined by pairwise one-sided Welch's t-tests ($\alpha=0.05$) using standard competition (1224) ranking. Lower ranks indicate better performance.}}
\label{tab:ts2_avg_ranks_app_stat}
\small
\begin{tabular}{llcc>{\cellcolor{blue!10}}c}
\toprule
& \textbf{Method} & \textbf{Co-smoothing} & \textbf{Forecasting} & \textbf{Avg. Rank $\downarrow$} \\
\midrule
\multirow{9}{*}{\rotatebox[origin=c]{90}{Single-Session}}
& Population coupling    & 7.76 & --   & 7.76 \\
& RRR                    & 5.14 & --   & 5.14 \\
& RRR (w/ ISI features)  & 4.62 & --   & 4.62 \\
& Trailing mean          & --   & 7.59 & 7.59 \\
& Shrinkage              & --   & 5.72 & 5.72 \\
& Ridge autoregression   & --   & 5.03 & 5.03 \\
& Autoencoder            & 6.34 & 6.86 & 6.60 \\
& NDT                    & 4.55 & 2.10 & 3.33 \\
& LFADS                  & 2.00 & 3.03 & 2.52 \\
\midrule
\multirow{2}{*}{\rotatebox[origin=c]{90}{Pre.}}
& NDT-Stitch             & 2.76 & 1.03 & 1.90 \\
& MtM                    & 1.34 & 3.93 & 2.64 \\
\bottomrule
\end{tabular}
\end{table}

\newpage

\subsection{TS3 Additional Results: Neuron Identity Prediction}

Table~\ref{tab:brain_region_full} expands on the TS3 results of Table~\ref{tab:ts3_main} by reporting per-region F1 scores across held-out sessions, offering a finer-grained view of where methods succeed and struggle across the 10 brain regions. In addition, we report per-region recall and precision in Table~\ref{tab:brain_region_full_recall} and Table~\ref{tab:brain_region_full_precision}, respectively.

We note that in Table~\ref{tab:ts3_main} seeds generally parameterize embedding generation, with the exception of the ISI baseline that has constant embeddings. For this we run multiple seeds on the MLP probe, though the linear probe is deterministic hence we omit SEM.

\input{tables/app/table_task3_brain_region_level}

\begin{table}[ht]
\centering
\caption{\footnotesize{\textbf{Per-region recall on TS3 neuron identity prediction.}
Recall for each of the 10 Cosmos-level brain regions, reported per method, unit level
(SU: single-unit; MU: multi-unit) and probe type. Values are averaged over seeds. The
rightmost column is the macro-average over regions, which for recall is exactly balanced
accuracy. Higher is better.}}
\label{tab:brain_region_full_recall}
\footnotesize
\resizebox{\textwidth}{!}{%
\begin{tabular}{l l l  cccccccccc  >{\cellcolor{blue!10}}c}
\toprule
\textbf{Method} & \textbf{Level} & \textbf{Probe} &
\textbf{CB} & \textbf{CNU} & \textbf{CTXsp} & \textbf{HB} &
\textbf{HPF} & \textbf{HY} & \textbf{Iso.} &
\textbf{MB} & \textbf{OLF} & \textbf{TH} & \textbf{Avg} \\
\midrule

\multicolumn{13}{l}{\textit{Transductive}} \\
\midrule
\multirow{4}{*}{POYO+}
    & \multirow{2}{*}{SU} & Linear & 0.215 & 0.066 & 0.233 & 0.059 & 0.013 & 0.089 & 0.032 & 0.102 & 0.075 & 0.413 & 0.130 \\
&                         & MLP    & 0.178 & 0.077 & 0.079 & 0.109 & 0.076 & 0.030 & 0.080 & 0.158 & 0.217 & 0.303 & 0.131 \\
    & \multirow{2}{*}{MU} & Linear & 0.249 & 0.042 & 0.283 & 0.013 & 0.001 & 0.037 & 0.017 & 0.067 & 0.046 & 0.704 & 0.146 \\
&                         & MLP    & 0.165 & 0.061 & 0.050 & 0.076 & 0.051 & 0.007 & 0.087 & 0.203 & 0.227 & 0.526 & 0.145 \\
\cmidrule(lr){1-14}
\multirow{4}{*}{POSSM}
    & \multirow{2}{*}{SU} & Linear & 0.036 & 0.057 & 0.033 & 0.197 & 0.194 & 0.130 & 0.056 & 0.146 & 0.065 & 0.251 & 0.117 \\
&                         & MLP    & 0.094 & 0.112 & 0.075 & 0.230 & 0.234 & 0.089 & 0.099 & 0.133 & 0.191 & 0.145 & 0.140 \\
    & \multirow{2}{*}{MU} & Linear & 0.002 & 0.017 & 0.013 & 0.164 & 0.290 & 0.052 & 0.045 & 0.218 & 0.014 & 0.416 & 0.123 \\
&                         & MLP    & 0.120 & 0.089 & 0.071 & 0.313 & 0.329 & 0.107 & 0.076 & 0.107 & 0.232 & 0.156 & 0.160 \\
\cmidrule(lr){1-14}
\multirow{4}{*}{NDT-Stitch}
    & \multirow{2}{*}{SU} & Linear & 0.202 & 0.032 & 0.204 & 0.271 & 0.104 & 0.133 & 0.111 & 0.129 & 0.106 & 0.330 & 0.162 \\
&                         & MLP    & 0.132 & 0.077 & 0.054 & 0.309 & 0.163 & 0.033 & 0.166 & 0.337 & 0.086 & 0.372 & 0.173 \\
    & \multirow{2}{*}{MU} & Linear & 0.263 & 0.027 & 0.229 & 0.372 & 0.098 & 0.144 & 0.118 & 0.139 & 0.113 & 0.410 & 0.191 \\
&                         & MLP    & 0.153 & 0.053 & 0.013 & 0.426 & 0.178 & 0.019 & 0.175 & 0.464 & 0.069 & 0.460 & 0.201 \\
\cmidrule(lr){1-14}
\multirow{4}{*}{MtM}
    & \multirow{2}{*}{SU} & Linear & 0.156 & 0.039 & 0.079 & 0.130 & 0.212 & 0.037 & 0.211 & 0.153 & 0.101 & 0.290 & 0.141 \\
&                         & MLP    & 0.026 & 0.056 & 0.008 & 0.091 & 0.225 & 0.004 & 0.311 & 0.434 & 0.011 & 0.319 & 0.148 \\
    & \multirow{2}{*}{MU} & Linear & 0.191 & 0.006 & 0.083 & 0.172 & 0.257 & 0.015 & 0.267 & 0.176 & 0.097 & 0.373 & 0.164 \\
&                         & MLP    & 0.006 & 0.041 & 0.004 & 0.056 & 0.265 & 0.000 & 0.362 & 0.612 & 0.002 & 0.387 & 0.173 \\

\midrule

\multicolumn{13}{l}{\textit{Inductive}} \\
\midrule
\multirow{4}{*}{ISI Baseline}
    & \multirow{2}{*}{SU} & Linear & $0.549$ & $0.059$ & $0.333$ & $0.358$ & $0.254$ & $0.130$ & $0.358$ & $0.245$ & $0.275$ & $0.416$ & $0.298$ \\
&                         & MLP    & $0.652$ & $0.227$ & $0.208$ & $0.502$ & $0.527$ & $0.178$ & $0.467$ & $0.254$ & $0.367$ & $0.764$ & $0.415$ \\
    & \multirow{2}{*}{MU} & Linear & $0.747$ & $0.032$ & $0.438$ & $0.505$ & $0.375$ & $0.074$ & $0.548$ & $0.211$ & $0.400$ & $0.693$ & $0.402$ \\
&                         & MLP    & $0.835$ & $0.286$ & $0.271$ & $0.730$ & $0.685$ & $0.200$ & $0.680$ & $0.359$ & $0.553$ & $0.946$ & $0.554$ \\
\cmidrule(lr){1-14}
\multirow{2}{*}{LOLCAT}
    & SU & Linear & 0.520 & 0.350 & 0.000 & 0.470 & 0.459 & 0.033 & 0.455 & 0.381 & 0.246 & 0.782 & 0.370 \\
&  MU & Linear & 0.712 & 0.508 & 0.000 & 0.591 & 0.623 & 0.000 & 0.661 & 0.505 & 0.325 & 0.948 & 0.487 \\
\cmidrule(lr){1-14}
\multirow{4}{*}{NEMO}
    & \multirow{2}{*}{SU} & Linear & $0.706$ & $0.395$ & $0.225$ & $0.517$ & $0.446$ & $0.304$ & $0.597$ & $0.365$ & $0.346$ & $0.881$ & $0.478$ \\
&                         & MLP    & $0.728$ & $0.495$ & $0.179$ & $0.572$ & $0.495$ & $0.256$ & $0.624$ & $0.393$ & $0.380$ & $0.886$ & $0.501$ \\
    & \multirow{2}{*}{MU} & Linear & $0.884$ & $0.557$ & $0.279$ & $0.671$ & $0.604$ & $0.430$ & $0.802$ & $0.484$ & $0.456$ & $0.980$ & $0.615$ \\
&                         & MLP    & $0.891$ & $0.675$ & $0.204$ & $0.672$ & $0.624$ & $0.389$ & $0.812$ & $0.554$ & $0.520$ & $0.981$ & $0.632$ \\
\cmidrule(lr){1-14}
\multirow{4}{*}{NuCLR}
    & \multirow{2}{*}{SU} & Linear & $0.965$ & $0.141$ & $0.362$ & $0.804$ & $0.678$ & $0.241$ & $0.702$ & $0.844$ & $0.662$ & $0.960$ & $0.636$ \\
&                         & MLP    & $0.953$ & $0.127$ & $0.479$ & $0.876$ & $0.680$ & $0.248$ & $0.654$ & $0.858$ & $0.748$ & $0.951$ & $0.658$ \\
    & \multirow{2}{*}{MU} & Linear & $0.965$ & $0.135$ & $0.379$ & $0.866$ & $0.726$ & $0.244$ & $0.776$ & $0.882$ & $0.741$ & $0.983$ & $0.670$ \\
&                         & MLP    & $0.951$ & $0.118$ & $0.479$ & $0.886$ & $0.711$ & $0.170$ & $0.737$ & $0.889$ & $0.764$ & $0.982$ & $0.669$ \\

\bottomrule
\end{tabular}%
}
\end{table}

\begin{table}[ht]
\centering
\caption{\footnotesize{\textbf{Per-region precision on TS3 neuron identity prediction.}
Precision for each of the 10 Cosmos-level brain regions, reported per method, unit level
(SU: single-unit; MU: multi-unit) and probe type. Values are averaged over seeds. The
rightmost column is the macro-average over regions. Higher is better.}}
\label{tab:brain_region_full_precision}
\footnotesize
\resizebox{\textwidth}{!}{%
\begin{tabular}{l l l  cccccccccc  >{\cellcolor{blue!10}}c}
\toprule
\textbf{Method} & \textbf{Level} & \textbf{Probe} &
\textbf{CB} & \textbf{CNU} & \textbf{CTXsp} & \textbf{HB} &
\textbf{HPF} & \textbf{HY} & \textbf{Iso.} &
\textbf{MB} & \textbf{OLF} & \textbf{TH} & \textbf{Avg} \\
\midrule

\multicolumn{13}{l}{\textit{Transductive}} \\
\midrule
\multirow{4}{*}{POYO+}
    & \multirow{2}{*}{SU} & Linear & 0.069 & 0.043 & 0.026 & 0.136 & 0.151 & 0.013 & 0.350 & 0.228 & 0.123 & 0.292 & 0.143 \\
&                         & MLP    & 0.055 & 0.049 & 0.027 & 0.114 & 0.156 & 0.006 & 0.305 & 0.299 & 0.154 & 0.286 & 0.145 \\
    & \multirow{2}{*}{MU} & Linear & 0.086 & 0.031 & 0.039 & 0.096 & 0.033 & 0.007 & 0.468 & 0.244 & 0.175 & 0.307 & 0.149 \\
&                         & MLP    & 0.058 & 0.051 & 0.019 & 0.103 & 0.170 & 0.002 & 0.338 & 0.319 & 0.225 & 0.317 & 0.160 \\
\cmidrule(lr){1-14}
\multirow{4}{*}{POSSM}
    & \multirow{2}{*}{SU} & Linear & 0.059 & 0.076 & 0.018 & 0.083 & 0.132 & 0.016 & 0.257 & 0.277 & 0.111 & 0.213 & 0.124 \\
&                         & MLP    & 0.086 & 0.070 & 0.038 & 0.092 & 0.142 & 0.019 & 0.223 & 0.295 & 0.150 & 0.244 & 0.136 \\
    & \multirow{2}{*}{MU} & Linear & 0.054 & 0.072 & 0.043 & 0.086 & 0.171 & 0.014 & 0.367 & 0.326 & 0.110 & 0.203 & 0.145 \\
&                         & MLP    & 0.090 & 0.087 & 0.023 & 0.110 & 0.184 & 0.015 & 0.258 & 0.345 & 0.190 & 0.272 & 0.157 \\
\cmidrule(lr){1-14}
\multirow{4}{*}{NDT-Stitch}
    & \multirow{2}{*}{SU} & Linear & 0.105 & 0.039 & 0.043 & 0.126 & 0.169 & 0.016 & 0.262 & 0.286 & 0.159 & 0.343 & 0.155 \\
&                         & MLP    & 0.114 & 0.073 & 0.074 & 0.134 & 0.200 & 0.017 & 0.282 & 0.316 & 0.265 & 0.357 & 0.183 \\
    & \multirow{2}{*}{MU} & Linear & 0.137 & 0.046 & 0.046 & 0.139 & 0.203 & 0.019 & 0.296 & 0.361 & 0.218 & 0.383 & 0.185 \\
&                         & MLP    & 0.178 & 0.080 & 0.037 & 0.168 & 0.270 & 0.031 & 0.291 & 0.379 & 0.351 & 0.385 & 0.217 \\
\cmidrule(lr){1-14}
\multirow{4}{*}{MtM}
    & \multirow{2}{*}{SU} & Linear & 0.102 & 0.048 & 0.019 & 0.061 & 0.222 & 0.011 & 0.257 & 0.273 & 0.096 & 0.405 & 0.149 \\
&                         & MLP    & 0.176 & 0.066 & 0.020 & 0.077 & 0.227 & 0.013 & 0.244 & 0.274 & 0.107 & 0.381 & 0.159 \\
    & \multirow{2}{*}{MU} & Linear & 0.142 & 0.013 & 0.020 & 0.072 & 0.268 & 0.006 & 0.280 & 0.339 & 0.094 & 0.506 & 0.174 \\
&                         & MLP    & 0.300 & 0.124 & 0.200 & 0.075 & 0.333 & 0.000 & 0.270 & 0.307 & 0.050 & 0.484 & 0.214 \\

\midrule

\multicolumn{13}{l}{\textit{Inductive}} \\
\midrule
\multirow{4}{*}{ISI Baseline}
    & \multirow{2}{*}{SU} & Linear & $0.283$ & $0.108$ & $0.055$ & $0.274$ & $0.318$ & $0.023$ & $0.422$ & $0.390$ & $0.228$ & $0.677$ & $0.278$ \\
&                         & MLP    & $0.416$ & $0.239$ & $0.076$ & $0.352$ & $0.463$ & $0.049$ & $0.565$ & $0.565$ & $0.353$ & $0.755$ & $0.383$ \\
    & \multirow{2}{*}{MU} & Linear & $0.492$ & $0.171$ & $0.076$ & $0.379$ & $0.417$ & $0.016$ & $0.529$ & $0.527$ & $0.290$ & $0.806$ & $0.370$ \\
&                         & MLP    & $0.700$ & $0.403$ & $0.167$ & $0.473$ & $0.556$ & $0.111$ & $0.693$ & $0.822$ & $0.505$ & $0.814$ & $0.524$ \\
\cmidrule(lr){1-14}
\multirow{2}{*}{LOLCAT}
    & SU & Linear & 0.488 & 0.212 & 0.000 & 0.491 & 0.357 & 0.026 & 0.471 & 0.509 & 0.321 & 0.723 & 0.360 \\
&  MU & Linear & 0.722 & 0.324 & 0.000 & 0.659 & 0.438 & 0.000 & 0.630 & 0.700 & 0.541 & 0.810 & 0.482 \\
\cmidrule(lr){1-14}
\multirow{4}{*}{NEMO}
    & \multirow{2}{*}{SU} & Linear & $0.568$ & $0.354$ & $0.057$ & $0.385$ & $0.465$ & $0.068$ & $0.713$ & $0.712$ & $0.366$ & $0.888$ & $0.458$ \\
&                         & MLP    & $0.623$ & $0.379$ & $0.068$ & $0.370$ & $0.463$ & $0.075$ & $0.719$ & $0.725$ & $0.390$ & $0.924$ & $0.474$ \\
    & \multirow{2}{*}{MU} & Linear & $0.798$ & $0.553$ & $0.107$ & $0.512$ & $0.614$ & $0.120$ & $0.779$ & $0.838$ & $0.550$ & $0.948$ & $0.582$ \\
&                         & MLP    & $0.836$ & $0.570$ & $0.155$ & $0.496$ & $0.578$ & $0.178$ & $0.775$ & $0.845$ & $0.565$ & $0.968$ & $0.597$ \\
\cmidrule(lr){1-14}
\multirow{4}{*}{NuCLR}
    & \multirow{2}{*}{SU} & Linear & $0.814$ & $0.358$ & $0.186$ & $0.883$ & $0.598$ & $0.209$ & $0.781$ & $0.891$ & $0.540$ & $0.910$ & $0.617$ \\
&                         & MLP    & $0.838$ & $0.343$ & $0.220$ & $0.855$ & $0.624$ & $0.170$ & $0.839$ & $0.891$ & $0.537$ & $0.920$ & $0.624$ \\
    & \multirow{2}{*}{MU} & Linear & $0.846$ & $0.376$ & $0.196$ & $0.905$ & $0.638$ & $0.416$ & $0.865$ & $0.918$ & $0.588$ & $0.915$ & $0.666$ \\
&                         & MLP    & $0.859$ & $0.319$ & $0.217$ & $0.895$ & $0.658$ & $0.262$ & $0.901$ & $0.905$ & $0.550$ & $0.917$ & $0.648$ \\

\bottomrule
\end{tabular}%
}
\end{table}

\clearpage
\newpage

\subsection{POYO Additional Results}

Table~\ref{tab:poyo_add_results} compares \textbf{POYO+}, a multi-task
supervised pre-trained model, against \textbf{POYO-ST}, a single-task
variant. POYO-ST trains a separate model for each
behavioral variable independently, allowing it to achieve stronger absolute
performance on several tasks (e.g., Licks $D^2$: $0.680$ vs.\ $0.664$,
Reward Acc: $0.568$ vs.\ $0.905$). However, this comes at the cost of
generality: a distinct model must be trained and maintained for every task,
and no shared representation is learned across behavioral variables.

POYO+, by contrast, learns a single model across all tasks simultaneously.
This introduces the non-trivial challenge of \emph{loss reweighting}: tasks
with different scales, difficulties, and label frequencies must be balanced
within a shared training objective, and suboptimal weighting directly hurts
performance on individual tasks. The modest gap in absolute metrics therefore
reflects not a fundamental limitation of the multi-task approach, but rather
the inherent difficulty of task balancing—a known open problem in multi-task
learning. The benefit is a single, general model that captures shared
structure across behavioral variables.

Interestingly, POYO-ST struggles with classification tasks, whereas POYO+ seems to generalize well. One possible explanation is that during multi-task training, additional supervision from dense targets (i.e.,~all the frame-level tasks) can assist in tasks with relatively low supervisory signal, e.g.,~classification tasks.

\begin{table}[ht]
\centering
\small
\caption{\textbf{POYO Models Performance Summary.} Results are reported as in
Table~\ref{tab:ts1_main} except avg.\ rank, which is not calculated.
POYO+ is a multi-task supervised pre-trained model;
POYO-ST is a single-task supervised pre-trained model;
POYO is a single-session baseline.}
\label{tab:poyo_add_results}
\vspace{1mm}
\resizebox{\columnwidth}{!}{%
\begin{tabular}{l cccccccc}
\toprule
\textbf{Model} & Licks & Whisker & Wheel & RPaw & LPaw & Reward & Choice & Contrast \\
 & ($D^2$) & ($R^2$) & ($R^2$) & ($R^2$) & ($R^2$) & (Acc) & (Acc) & (Acc) \\
\midrule
\textbf{POYO+}
  & $0.664\tpms{0.005}$
  & $0.390\tpms{0.001}$
  & $0.351\tpms{0.001}$
  & $0.247\tpms{0.001}$
  & $0.197\tpms{0.001}$
  & $0.905\tpms{0.003}$
  & $0.719\tpms{0.002}$
  & $0.259\tpms{0.003}$ \\
\textbf{POYO-ST}
  & $0.680\tpms{0.006}$
  & $0.319\tpms{0.003}$
  & $0.344\tpms{0.002}$
  & $0.236\tpms{0.001}$
  & $0.132\tpms{0.003}$
  & $0.568\tpms{0.003}$
  & $0.497\tpms{0.002}$
  & $0.200\tpms{0.002}$ \\

\textbf{POYO}
  & $0.536\tpms{0.008}$
  & $0.282\tpms{0.005}$
  & $0.309\tpms{0.005}$
  & $0.192\tpms{0.003}$
  & $0.145\tpms{0.002}$
  & $0.861\tpms{0.002}$
  & $0.633\tpms{0.004}$
  & $0.222\tpms{0.001}$ \\
\bottomrule
\end{tabular}%
}
\end{table}

\subsection{NDT2 and NEDS Additional Results}

Table~\ref{tab:neds_ndt2_results} reports detailed results for NDT2 and NEDS alongside a single-session linear baseline. Both models perform at or below the linear baseline across most tasks: NDT2 falls short of linear on Licks and Wheel, while NEDS underperforms linear on Wheel, Reward, and Choice. Rather than capturing the nonlinear structure of neural population
activity, these models largely recover what a simple linear decoder can already extract—suggesting a failure to leverage the representational capacity of their architectures under the fine-tuning regime evaluated here.

\begin{table}[ht]
\centering
\small
\caption{\textbf{NEDS and NDT2 Models Performance Summary.} Results are reported as in
Table~\ref{tab:ts1_main} except avg.\ rank, which is not calculated.}
\label{tab:neds_ndt2_results}
\vspace{1mm}
\resizebox{\columnwidth}{!}{%
\begin{tabular}{l cccccccc}
\toprule
\textbf{Model} & Licks & Whisker & Wheel & RPaw & LPaw & Reward & Choice & Contrast \\
 & ($D^2$) & ($R^2$) & ($R^2$) & ($R^2$) & ($R^2$) & (Acc) & (Acc) & (Acc) \\
\midrule

\textbf{Linear} & $0.152 \tpms{0.001}$ & $0.177 \tpms{0.001}$ & $0.234 \tpms{0.001}$ & $0.096 \tpms{0.001}$ & $0.090 \tpms{0.001}$ & $0.833 \tpms{0.010}$ & $0.624 \tpms{0.003}$ & $0.217 \tpms{0.001}$ \\

\textbf{NDT2}
  & $0.127\tpms{0.003}$
  & $0.229\tpms{0.002}$
  & $0.211\tpms{0.003}$
  & $0.103\tpms{0.004}$
  & $0.100\tpms{0.001}$
  & $0.847\tpms{0.003}$
  & $0.618\tpms{0.002}$
  & $0.214\tpms{0.001}$ \\
  
\textbf{NEDS}
  & $0.220\tpms{0.012}$
  & $0.243\tpms{0.012}$
  & $0.068\tpms{0.010}$
  & $0.136\tpms{0.003}$
  & $0.075\tpms{0.003}$
  & $0.659\tpms{0.009}$
  & $0.547\tpms{0.004}$
  & $0.206\tpms{0.001}$ \\

\bottomrule
\end{tabular}%
}
\end{table}




%% file: tables/app/table_task3_brain_region_level.tex
\begin{table}[h!]
\centering
\small
\caption{{\bf Brain region prediction performance across held-out sessions.}
F1-scores are reported per region and as macro-average.
Methods are evaluated on Single Unit (SU) and Multi-Unit (MU) data
with both Linear and MLP probes.
CB = Cerebellum, CNU = Caudate-Putamen, CTXsp = Cortical Subplate,
HB = Hindbrain, HPF = Hippocampal Formation, HY = Hypothalamus, Iso. = Isocortex
MB = Midbrain, OLF = Olfactory Areas, TH = Thalamus.}
\resizebox{\columnwidth}{!}{
\begin{tabular}{l l l  cccccccccc  >{\cellcolor{blue!10}}c}
\toprule
\textbf{Method} & \textbf{Level} & \textbf{Probe} &
\textbf{CB} & \textbf{CNU} & \textbf{CTXsp} & \textbf{HB} &
\textbf{HPF} & \textbf{HY} & \textbf{Iso.} &
\textbf{MB} & \textbf{OLF} & \textbf{TH} & \textbf{Avg} \\
\midrule

\multicolumn{13}{l}{\textit{Transductive}} \\
\midrule
\multirow{4}{*}{POYO+}
    & \multirow{2}{*}{SU} & Linear & 0.105 & 0.052 & 0.046 & 0.082 & 0.024 & 0.023 & 0.059 & 0.141 & 0.093 & 0.342 & 0.097 \\
&                         & MLP    & 0.083 & 0.057 & 0.039 & 0.110 & 0.102 & 0.009 & 0.125 & 0.206 & 0.178 & 0.291 & 0.120 \\
    & \multirow{2}{*}{MU} & Linear & 0.127 & 0.035 & 0.069 & 0.023 & 0.001 & 0.011 & 0.032 & 0.104 & 0.072 & 0.427 & 0.090 \\
&                         & MLP    & 0.086 & 0.051 & 0.028 & 0.085 & 0.079 & 0.003 & 0.134 & 0.246 & 0.216 & 0.389 & 0.132 \\
\cmidrule(lr){1-14}
\multirow{4}{*}{POSSM}
    & \multirow{2}{*}{SU} & Linear & 0.044 & 0.065 & 0.023 & 0.116 & 0.157 & 0.028 & 0.092 & 0.191 & 0.082 & 0.230 & 0.103 \\
&                         & MLP    & 0.089 & 0.086 & 0.050 & 0.131 & 0.177 & 0.031 & 0.137 & 0.183 & 0.167 & 0.182 & 0.123 \\
    & \multirow{2}{*}{MU} & Linear & 0.005 & 0.027 & 0.019 & 0.113 & 0.215 & 0.021 & 0.079 & 0.261 & 0.024 & 0.273 & 0.104 \\
&                         & MLP    & 0.102 & 0.086 & 0.035 & 0.163 & 0.236 & 0.026 & 0.118 & 0.163 & 0.206 & 0.197 & 0.133 \\
\cmidrule(lr){1-14}
\multirow{4}{*}{NDT-Stitch}
    & \multirow{2}{*}{SU} & Linear & 0.138 & 0.035 & 0.070 & 0.171 & 0.126 & 0.029 & 0.155 & 0.177 & 0.127 & 0.335 & 0.136 \\
&                         & MLP    & 0.119 & 0.074 & 0.062 & 0.187 & 0.178 & 0.022 & 0.209 & 0.326 & 0.130 & 0.363 & 0.167 \\
    & \multirow{2}{*}{MU} & Linear & 0.179 & 0.033 & 0.075 & 0.201 & 0.123 & 0.033 & 0.167 & 0.199 & 0.144 & 0.391 & 0.155 \\
&                         & MLP    & 0.153 & 0.063 & 0.019 & 0.240 & 0.209 & 0.023 & 0.218 & 0.416 & 0.113 & 0.417 & 0.187 \\
\cmidrule(lr){1-14}
\multirow{4}{*}{MtM}
    & \multirow{2}{*}{SU} & Linear & 0.123 & 0.042 & 0.030 & 0.083 & 0.217 & 0.016 & 0.232 & 0.195 & 0.097 & 0.337 & 0.137 \\
&                         & MLP    & 0.045 & 0.060 & 0.012 & 0.083 & 0.225 & 0.006 & 0.273 & 0.336 & 0.019 & 0.346 & 0.140 \\
    & \multirow{2}{*}{MU} & Linear & 0.162 & 0.008 & 0.033 & 0.101 & 0.262 & 0.008 & 0.271 & 0.229 & 0.092 & 0.427 & 0.159 \\
&                         & MLP    & 0.012 & 0.059 & 0.008 & 0.064 & 0.290 & 0.000 & 0.308 & 0.409 & 0.003 & 0.429 & 0.158 \\

\midrule

\multicolumn{13}{l}{\textit{Inductive}} \\
\midrule
\multirow{4}{*}{ISI Baseline}
    & \multirow{2}{*}{SU} & Linear & $0.374$ & $0.076$ & $0.095$ & $0.310$ & $0.282$ & $0.039$ & $0.388$ & $0.301$ & $0.250$ & $0.515$ & $0.263$ \\
&                         & MLP    & $0.504$ & $0.228$ & $0.110$ & $0.413$ & $0.493$ & $0.076$ & $0.511$ & $0.350$ & $0.359$ & $0.759$ & $0.380$ \\
    & \multirow{2}{*}{MU} & Linear & $0.593$ & $0.054$ & $0.130$ & $0.433$ & $0.395$ & $0.026$ & $0.539$ & $0.301$ & $0.337$ & $0.745$ & $0.355$ \\
&                         & MLP    & $0.757$ & $0.324$ & $0.199$ & $0.573$ & $0.614$ & $0.140$ & $0.686$ & $0.498$ & $0.527$ & $0.875$ & $0.519$ \\
\cmidrule(lr){1-14}
\multirow{2}{*}{LOLCAT}
    & SU & Linear & 0.500 & 0.262 & 0.000 & 0.480 & 0.401 & 0.028 & 0.462 & 0.435 & 0.272 & 0.751 & 0.359 \\
&  MU & Linear & 0.713 & 0.393 & 0.000 & 0.621 & 0.514 & 0.000 & 0.643 & 0.586 & 0.391 & 0.873 & 0.473 \\
\cmidrule(lr){1-14}
\multirow{4}{*}{NEMO}
    & \multirow{2}{*}{SU} & Linear & $0.629$ & $0.373$ & $0.091$ & $0.441$ & $0.455$ & $0.110$ & $0.650$ & $0.482$ & $0.356$ & $0.884$ & $0.447$ \\
&                         & MLP    & $0.671$ & $0.429$ & $0.099$ & $0.449$ & $0.478$ & $0.115$ & $0.668$ & $0.510$ & $0.385$ & $0.905$ & $0.471$ \\
    & \multirow{2}{*}{MU} & Linear & $0.838$ & $0.555$ & $0.155$ & $0.581$ & $0.609$ & $0.187$ & $0.790$ & $0.613$ & $0.499$ & $0.964$ & $0.579$ \\
&                         & MLP    & $0.863$ & $0.618$ & $0.175$ & $0.570$ & $0.600$ & $0.242$ & $0.793$ & $0.668$ & $0.541$ & $0.974$ & $0.605$ \\
\cmidrule(lr){1-14}
\multirow{4}{*}{NuCLR}
    & \multirow{2}{*}{SU} & Linear & $0.883$ & $0.200$ & $0.241$ & $0.841$ & $0.635$ & $0.222$ & $0.739$ & $0.867$ & $0.595$ & $0.934$ & $0.616$ \\
&                         & MLP    & $0.892$ & $0.185$ & $0.301$ & $0.865$ & $0.651$ & $0.202$ & $0.735$ & $0.874$ & $0.625$ & $0.935$ & $0.626$ \\
    & \multirow{2}{*}{MU} & Linear & $0.901$ & $0.197$ & $0.253$ & $0.885$ & $0.679$ & $0.306$ & $0.817$ & $0.899$ & $0.655$ & $0.948$ & $0.654$ \\
&                         & MLP    & $0.902$ & $0.171$ & $0.298$ & $0.890$ & $0.683$ & $0.206$ & $0.810$ & $0.897$ & $0.640$ & $0.949$ & $0.645$ \\

\bottomrule
\end{tabular}
}
\label{tab:brain_region_full}
\end{table}

%% file: appendix/related_work.tex
\section{Related Work}
\subsection{Neural Data Benchmarks}

A growing number of benchmarks have been proposed to evaluate models of neural activity, each targeting specific aspects of neural representation learning. The Neural Latents Benchmark (NLB)~\cite{pei2021neural} provides a standardized framework for evaluating latent variable models on neural population dynamics, focusing on within-session reconstruction (co-smoothing) and behavior prediction tasks. While NLB has been instrumental in advancing dynamical systems modeling, the evaluation is limited to single sessions and they do not evaluate generalization across animals or tasks.

More recent efforts have begun to incorporate cross-session and multi-task evaluation. The FALCON benchmark~\cite{karpowicz2024few} evaluates few-shot transfer for neural decoding across datasets and subjects, highlighting the importance of cross-animal generalization. However, its task structure remains relatively narrow, focusing primarily on decoding and not explicitly evaluating dynamical modeling or anatomical structure. Similarly, Sensorium~\cite{willeke2022sensorium} evaluates large-scale neural prediction in visual cortex, enabling comparison of models for stimulus-response mapping. While Sensorium provides a strong testbed for within-domain generalization, it is restricted to a single brain region and task setting.

Beyond these, several community-driven benchmarks and competitions have focused on specific subproblems in neuroscience, including spike inference~\cite{berens2018community}, spike sorting~\cite{magland2020spikeforest}, and brain-computer interface decoding~\cite{blankertz2004bci, wei20222021}. While these efforts have been critical for advancing individual components of the analysis pipeline, they do not provide a unified evaluation of representation learning across multiple dimensions of neural data.

In contrast, \brainwidebench~is designed to evaluate models across three complementary task families while simultaneously supporting multi-region, across-animal, and multi-task evaluation. As summarized in Table~\ref{tab:neural_benchmarks}, this combination of properties distinguishes it from prior benchmarks, which typically focus on only one or two of these dimensions.

\begin{table}[h]
\centering
\setlength{\tabcolsep}{2.8pt}
\renewcommand{\arraystretch}{1.12}
\small
\caption{
\textbf{Comparison of neural data benchmarks across scale and evaluation axes.}
We summarize dataset scale using reported recording hours, subjects, sessions/recordings, and recorded units.
For FALCON, ``Units'' denotes recording channels rather than spike-sorted units.
For \brainwidebench, we report both well-isolated units and total detected units.
``n/r'' denotes not reported as a unified benchmark-level quantity.
Prior benchmarks focus on isolated aspects of neural modeling, whereas \brainwidebench~jointly evaluates behavior, dynamics, and anatomy across brain-wide recordings and animals.
}
\resizebox{\textwidth}{!}{
\begin{tabular}{lcccccccccc}
\toprule
\textbf{Benchmark} 
& \textbf{Hours} 
& \textbf{Subjects} 
& \textbf{Sessions} 
& \textbf{Units}
& \textbf{Across-Regions} 
& \textbf{Across-Subject} 
& \textbf{\# Tasks} 
& \textbf{Behavior} 
& \textbf{Dynamics} 
& \textbf{Anatomy} \\
\midrule

NLB~\cite{pei2021neural}
& 1.75 h
& 4
& 7
& 887
& $\times$ 
& $\times$ 
& 4
& \checkmark 
& \checkmark 
& $\times$ \\

FALCON~\cite{karpowicz2024few}
& 50 h
& 6
& 68
& 789 ch.
& $\times$ 
& $\times$ 
& 5
& \checkmark 
& $\times$ 
& $\times$ \\
Sensorium~\cite{willeke2022sensorium}
& 49 h
& 7
& 7
& 28k+
& $\times$
& $\times$ 
& 1 
& \checkmark 
& $\times$ 
& $\times$ \\

Dynamic Sensorium~\cite{dynamic_sensorium_retrospective2024}
& 20 h
& 10
& 10
& 78.9k
& $\times$
& $\times$ 
& 1 
& \checkmark 
& \checkmark 
& $\times$ \\

\midrule
\brainwidebench~(Ours)
& 600+ h
& 139
& 452
& 611k
& \checkmark (brain-wide) 
& \checkmark 
& 3 suites, 12 tasks 
& \checkmark 
& \checkmark 
& \checkmark \\
\bottomrule
\end{tabular}
}
\vspace{2mm}
\label{tab:neural_benchmarks}
\end{table}






\subsection{Large-scale Pretraining for Neural Data}

Recent work has explored the development of large-scale pretraining for neural activity, motivated by its success in other domains~\cite{bommasani2021opportunities, radford2019language, achiam2023gpt}. A range of architectures have been proposed to learn representations from large neural datasets, including transformer-based approaches such as Neural Data Transformers (NDT and its variants, e.g., NDT2 and NDT-Stitch)~\cite{ye2024neural}, as well as models designed for multi-session and population-level learning such as POYO and POSSM~\cite{azabou2024unified, ryoo2025generalizable}. 

In parallel, several methods have focused on self-supervised and multimodal representation learning. Masked modeling approaches such as MtM extend ideas from language and vision to neural time series, while embedding-based methods such as NEDS~\cite{zhang2024towards} and contrastive approaches such as NEMO and NuCLR~\cite{yu2025in, arora2025know} aim to align neural activity across contexts, animals, or modalities. Together, these approaches demonstrate the promise of large-scale pretraining for capturing structure in neural data, but are typically evaluated on different datasets and tasks, making it difficult to assess their relative strengths in a unified setting.

These models have demonstrated promising results in tasks such as decoding, neural prediction, and cross-session transfer. However, evaluation of these models remains fragmented, with different works focusing on different datasets, tasks, and metrics. As a result, it is difficult to assess whether improvements reflect general advances in representation learning or task-specific gains. 
\brainwidebench~addresses this gap by providing a unified evaluation framework that enables systematic comparison of pretraining strategies across multiple tasks and data regimes. In doing so, it complements existing modeling work by providing a common standard for measuring progress.

\subsection{Scaling and Generalization in Neural Data}

A central question in the development of neural foundation models is how performance scales with data size and diversity. Inspired by scaling laws observed in language and vision models~\cite{kaplan2020scaling, hoffmann2022training, cherti2023reproducible}, recent work has begun to investigate scaling behavior in neural data~\cite{antonello2023scaling, sato2024scaling, banville2025scaling}. However, emerging evidence suggests that scaling in neural data is more complex, with heterogeneity across sessions and subjects playing a critical role~\cite{jiang2025data}.

At the same time, a growing body of work highlights the importance of generalization across animals and recording conditions~\cite{safaie2023preserved, azabou2025multisession}. Neural activity exhibits substantial variability across individuals, making it challenging to learn invariant representations that transfer without adaptation.

By enabling evaluation across animals, regions, and tasks, \brainwidebench~provides a platform for systematically studying scaling and generalization in representation learning from neural data. In particular, the inclusion of quality-controlled metadata allows future work to investigate how data quality, diversity, and structure impact representation learning.




%% file: appendix/contribution_statement.tex
\section{Contribution Statement}
\label{app:contribution_statement}

\input{tables/app/contribution_statement}

\newpage

\textbf{Definitions of contribution categories from Table~\ref{tab:author-contributions}.}
The categories follow the \href{https://credit.niso.org}{CRediT} (Contributor Roles Taxonomy) definitions.\\

\textbf{Conceptualization.} Ideas; formulation or evolution of overarching
research goals and aims.

\textbf{Methodology.} Development or design of methodology; creation of models.

\textbf{Data Collection.} Conducting a research and investigation process,
specifically performing the experiments, or data/evidence collection.

\textbf{Data Curation.} Management activities to annotate (produce metadata),
scrub data and maintain research data (including software code, where it is
necessary for interpreting the data itself) for initial use and later re-use.

\textbf{Resources.} Provision of study materials, reagents, materials, patients,
laboratory samples, animals, instrumentation, computing resources, or other
analysis tools.

\textbf{Software (pipeline development).} Programming, software development; designing computer programs; implementation of the computer code and supporting algorithms; testing of existing code components.

\textbf{Model Training and Analysis.} Application of statistical, mathematical,
computational, or other formal techniques to analyze, or synthesize study data.

\textbf{Project Administration.} Management and coordination responsibility for
the research activity planning and execution.

\textbf{Writing -- Original Draft Preparation.} Preparation, creation, and/or
presentation of the published work, specifically writing the initial draft
(including substantive translation).

\textbf{Writing -- Review \& Editing.} Preparation, creation, and/or presentation
of the published work by those from the original research group, specifically
critical review, commentary, or revision -- including pre- or post-publication
stages.

\textbf{Funding Acquisition.} Acquisition of financial support for the
project leading to this publication.

%% file: tables/app/contribution_statement.tex
\begingroup
\scriptsize
\setlength{\tabcolsep}{2.5pt}
\renewcommand{\arraystretch}{1.08}

\def\acMark{$\bullet$}        
\definecolor{acstripe}{gray}{0.93}   
\def\acAngle{35}              

\def\acY{\acMark}
\def\acN{}
\def\acRot#1{\makebox[0pt][l]{\rotatebox[origin=l]{\acAngle}{\textbf{#1}}}}
\def\acAff#1{\textsuperscript{#1}}
\captionsetup{type=table, singlelinecheck=false}
\caption{\textbf{Author contributions.} A filled dot indicates that the author contributed to that role. Superscript numbers after each name refer to the affiliation list; letters in the
Grants column refer to the funding list.
}
\label{tab:author-contributions}
\vspace{2pt}

\rowcolors{2}{acstripe}{white}
\noindent\begin{tabular}{%
  >{\raggedright\arraybackslash}p{3.75cm}@{\hspace{7pt}}     
  *{2}{>{\centering\arraybackslash}p{1.7em}}@{\hspace{6pt}}  
  *{3}{>{\centering\arraybackslash}p{1.7em}}@{\hspace{6pt}}  
  *{2}{>{\centering\arraybackslash}p{1.7em}}@{\hspace{6pt}}  
  *{4}{>{\centering\arraybackslash}p{1.7em}}@{\hspace{7pt}}  
  >{\raggedright\arraybackslash}p{1.65cm}                    
}
\toprule
\multicolumn{1}{l}{\bf{Author}} &
\multicolumn{1}{l}{\acRot{Conceptualization}} &
\multicolumn{1}{l}{\acRot{Methodology}} &
\multicolumn{1}{l}{\acRot{Data Collection}} &
\multicolumn{1}{l}{\acRot{Data Curation}} &
\multicolumn{1}{l}{\acRot{Resources}} &
\multicolumn{1}{l}{\acRot{Software (Pipeline Development)}} &
\multicolumn{1}{l}{\acRot{Model Training and Analysis}} &
\multicolumn{1}{l}{\acRot{Project Administration}} &
\multicolumn{1}{l}{\acRot{Writing -- Original Draft Preparation}} &
\multicolumn{1}{l}{\acRot{Writing -- Review \& Editing}} &
\multicolumn{1}{l}{\acRot{Funding Acquisition}} &
\multicolumn{1}{l}{\acRot{Grants}} \\
\midrule
Alexandre Andre\acAff{1,*} &
\acY & \acY & \acN & \acY & \acN & \acY & \acY & \acN & \acY & \acY & \acN &  \\
Shivashriganesh P. Mahato\acAff{1,*} &
\acY & \acY & \acN & \acY & \acN & \acY & \acY & \acN & \acY & \acY & \acN &  \\
Vinam Arora\acAff{1} &
\acY & \acY & \acN & \acY & \acN & \acY & \acY & \acN & \acN & \acY & \acN &  \\
Keshav Balaji\acAff{1} &
\acN & \acY & \acN & \acN & \acN & \acN & \acY & \acN & \acY & \acN & \acN &  \\
Divyansha Lachi\acAff{1} &
\acY & \acY & \acN & \acN & \acN & \acY & \acY & \acN & \acY & \acY & \acN &  \\
Nanda H. Krishna\acAff{2,3} &
\acN & \acN & \acN & \acN & \acN & \acN & \acY & \acN & \acY & \acN & \acN & a \\
Jingyun Xiao\acAff{1} &
\acN & \acN & \acN & \acN & \acN & \acN & \acY & \acN & \acY & \acN & \acN &  \\
Yizi Zhang\acAff{4} &
\acN & \acY & \acN & \acN & \acN & \acN & \acY & \acN & \acN & \acY & \acN & b,c \\
Ximeng Mao\acAff{2,3} &
\acN & \acN & \acN & \acN & \acN & \acN & \acY & \acN & \acN & \acN & \acN &  \\
Wenrui Ma\acAff{1} &
\acN & \acN & \acN & \acN & \acN & \acN & \acY & \acN & \acN & \acN & \acN &  \\
Han Yu\acAff{5} &
\acN & \acY & \acN & \acN & \acN & \acN & \acN & \acN & \acN & \acN & \acN & b,c \\
\addlinespace[3pt]
International Brain Laboratory &
\acN & \acN & \acN & \acN & \acN & \acN & \acN & \acN & \acN & \acN & \acN & d,e,f \\
Daniel Birman\acAff{6} &
\acN & \acN & \acN & \acN & \acY & \acN & \acN & \acN & \acN & \acN & \acN & d,e \\
Niccol\`{o} Bonacchi\acAff{7,8} &
\acN & \acY & \acN & \acY & \acY & \acY & \acN & \acN & \acN & \acY & \acN & d,e \\
Gaelle A. Chapuis\acAff{9} &
\acN & \acN & \acN & \acY & \acY & \acN & \acN & \acY & \acN & \acN & \acN & d,e \\
Joana A. Catarino\acAff{10} &
\acN & \acN & \acY & \acY & \acN & \acN & \acN & \acN & \acN & \acN & \acN & d,e \\
Felicia Davatolhagh\acAff{11} &
\acN & \acN & \acY & \acY & \acN & \acN & \acN & \acN & \acN & \acN & \acN & d,e \\
Mayo Faulkner\acAff{12} &
\acN & \acN & \acN & \acY & \acN & \acY & \acN & \acN & \acN & \acN & \acN & d,e \\
Laura Freitas-Silva\acAff{13} &
\acN & \acN & \acY & \acY & \acN & \acN & \acN & \acN & \acN & \acN & \acN & d,e \\
Fei Hu\acAff{14} &
\acN & \acN & \acY & \acY & \acN & \acN & \acN & \acN & \acN & \acN & \acN & d,e \\
Julia M. Huntenburg\acAff{13} &
\acN & \acN & \acN & \acY & \acY & \acY & \acN & \acN & \acN & \acN & \acN & d,e \\
Anup Khanal\acAff{11} &
\acN & \acN & \acY & \acY & \acN & \acN & \acN & \acN & \acN & \acN & \acN & d,e \\
In\^{e}s Laranjeira\acAff{13} &
\acN & \acN & \acY & \acY & \acN & \acN & \acN & \acN & \acN & \acN & \acN & d,e \\
Petrina Lau\acAff{15} &
\acN & \acN & \acY & \acY & \acN & \acN & \acN & \acN & \acN & \acN & \acN & d,e \\
Guido T. Meijer\acAff{16} &
\acN & \acN & \acY & \acY & \acN & \acN & \acN & \acN & \acN & \acN & \acN & d,e \\
Nathaniel J. Miska\acAff{12} &
\acN & \acN & \acY & \acY & \acN & \acN & \acN & \acN & \acN & \acN & \acN & d,e \\
Jean-Paul Noel\acAff{17} &
\acN & \acN & \acY & \acY & \acN & \acN & \acN & \acN & \acN & \acN & \acN & d,e,g,h \\
Alejandro Pan-Vazquez\acAff{18} &
\acN & \acN & \acY & \acY & \acN & \acN & \acN & \acN & \acN & \acN & \acN & d,e \\
Georg Raiser\acAff{13} &
\acN & \acN & \acN & \acN & \acY & \acY & \acN & \acN & \acN & \acN & \acN & d,e \\
Cyrille Rossant\acAff{12} &
\acN & \acN & \acN & \acN & \acY & \acY & \acN & \acN & \acN & \acN & \acN & d,e \\
Karolina Z. Socha\acAff{11} &
\acN & \acN & \acY & \acN & \acN & \acN & \acN & \acN & \acN & \acN & \acN & d,e \\
Anne E. Urai\acAff{19} &
\acN & \acN & \acY & \acN & \acN & \acN & \acN & \acN & \acN & \acN & \acN & d,e,i \\
Miles J. Wells\acAff{12} &
\acN & \acN & \acN & \acN & \acN & \acY & \acN & \acN & \acN & \acN & \acN & d,e \\
Steven J. West\acAff{12} &
\acN & \acN & \acY & \acN & \acY & \acN & \acN & \acN & \acN & \acN & \acN & d,e \\
Olivier Winter\acAff{13} &
\acN & \acN & \acN & \acY & \acY & \acY & \acN & \acY & \acN & \acN & \acN & d,e \\
\addlinespace[3pt]
Blake Richards\acAff{20,2} &
\acN & \acY & \acN & \acN & \acN & \acN & \acN & \acN & \acN & \acY & \acY & j,k \\
Guillaume Lajoie\acAff{3,2} &
\acN & \acY & \acN & \acN & \acN & \acN & \acN & \acN & \acN & \acY & \acY & l,m \\
Cole Hurwitz\acAff{21} &
\acY & \acY & \acN & \acY & \acN & \acN & \acN & \acY & \acN & \acN & \acN & b,c \\
Mehdi Azabou\acAff{5} &
\acY & \acY & \acN & \acY & \acN & \acY & \acY & \acN & \acN & \acN & \acN & b,j \\
Matthew R. Whiteway\acAff{5,\dag} &
\acY & \acY & \acN & \acY & \acN & \acY & \acN & \acY & \acY & \acY & \acN & b,c,d,f,n,o \\
Liam Paninski\acAff{5,\dag} &
\acY & \acY & \acN & \acN & \acN & \acN & \acN & \acY & \acN & \acY & \acY & b,c,d,f,j \\
Eva L. Dyer\acAff{1,\dag} &
\acY & \acY & \acN & \acN & \acN & \acN & \acN & \acY & \acY & \acY & \acY & j,p,q\\
\bottomrule
\end{tabular}
 \par\vspace{4pt}
\textbf{Affiliations.} \textsuperscript{1}University of Pennsylvania. \textsuperscript{2}Mila. \textsuperscript{3}Universit\'{e} de Montr\'{e}al. \textsuperscript{4}Stanford University. \textsuperscript{5}Columbia University. \textsuperscript{6}Allen Institute. \textsuperscript{7}William James Center for Research. \textsuperscript{8}ISPA - Instituto Universit\'{a}rio. \textsuperscript{9}University of Geneva. \textsuperscript{10}Karolinska Institutet. \textsuperscript{11}UCLA. \textsuperscript{12}University College London. \textsuperscript{13}Champalimaud Foundation. \textsuperscript{14}Lingang Laboratory. \textsuperscript{15}The Chinese University of Hong Kong. \textsuperscript{16}Donders Institute. \textsuperscript{17}University of Minnesota. \textsuperscript{18}Princeton University. \textsuperscript{19}Leiden University. \textsuperscript{20}McGill University. \textsuperscript{21}IBM.
\par\vspace{2pt}
\textbf{Funding.} \textsuperscript{a}FRQNT 2009130. \textsuperscript{b}NSF 1707398. \textsuperscript{c}Gatsby Charitable Foundation GAT3708. \textsuperscript{d}Simons Foundation 543023. \textsuperscript{e}Wellcome Trust 216324. \textsuperscript{f}NIH U19NS123716. \textsuperscript{g}NIH R00NS128075. \textsuperscript{h}Sloan Research Fellowship. \textsuperscript{i}Leopoldina fellowship. \textsuperscript{j}NSF/DoD OUSD (R\&E) DBI-2229929 (ARNI). \textsuperscript{k}NSERC DG RGPIN-2020-05105. \textsuperscript{l}Canada CIFAR AI Research Chair program. \textsuperscript{m}Canada Research Chair in Neural Computations and Interfacing. \textsuperscript{n}NIH 1R50NS145433. \textsuperscript{o}ZI Team Science. \textsuperscript{p}NSF CAREER Award RI:2146072. \textsuperscript{q}CIFAR Learning in Machines and Brains Program.

\par\vspace{2pt}
\textsuperscript{*}These authors contributed equally.
\textsuperscript{\dag}These senior authors contributed equally.
\endgroup